\documentclass[11pt]{article}
\usepackage[sfdefault]{carlito}

\usepackage[margin=1in]{geometry}  
\usepackage[singlespacing]{setspace}
\usepackage{epsfig}
\usepackage{graphics,color}
\usepackage{amssymb,amsfonts,amsthm,amsmath}
\usepackage{authblk}
\usepackage{multirow}
\usepackage{mathrsfs}
\usepackage{enumerate}
\usepackage{enumitem}
\usepackage{natbib}
\usepackage{float}
\usepackage{algorithm}
\usepackage{algpseudocode}
\usepackage{subcaption} 
\usepackage{url}
\usepackage[colorlinks=true, linkcolor=blue, citecolor=blue]{hyperref}
\usepackage{bm}
\usepackage{verbatim}
\usepackage{booktabs}
\usepackage{makecell}
\usepackage{needspace}
\usepackage[normalem]{ulem}

\usepackage{longtable,booktabs,array}
\usepackage{calc} 
\usepackage{etoolbox}
\makeatletter
\patchcmd\longtable{\par}{\if@noskipsec\mbox{}\fi\par}{}{}
\makeatother
\IfFileExists{footnotehyper.sty}{\usepackage{footnotehyper}}{\usepackage{footnote}}
\makesavenoteenv{longtable}
\usepackage{graphicx}
\makeatletter
\def\maxwidth{\ifdim\Gin@nat@width>\linewidth\linewidth\else\Gin@nat@width\fi}
\def\maxheight{\ifdim\Gin@nat@height>\textheight\textheight\else\Gin@nat@height\fi}
\makeatother
\setkeys{Gin}{width=\maxwidth,height=\maxheight,keepaspectratio}
\makeatletter
\def\fps@figure{htbp}
\makeatother

\newtheorem{theorem}{Theorem}
\newtheorem{lemma}{Lemma}
\newtheorem{definition}{Definition}
\newtheorem{proposition}{Proposition}
\newtheorem{corollary}{Corollary}

\newcommand{\dd}{\mbox{d}}

\newcommand{\var}{\mbox{var}}

\DeclareMathOperator*{\argmin}{argmin}

\DeclareMathOperator*{\E}{E}

\graphicspath{ {./Figures/} }

\begin{document}

\def\spacingset#1{\renewcommand{\baselinestretch}%
{#1}\small\normalsize} \spacingset{1}

  \title{\bf Optimal Kernel Learning for Gaussian Process Models with High-Dimensional Input}
  \author{Lulu Kang\thanks{
    The authors gratefully acknowledge \textit{National Science Foundation award DMS \# 2429324.}}\hspace{.2cm}
    and Minshen Xu\\
    Department of Mathematics and Statistics, \\
    University of Massachusetts Amherst}
  \maketitle

\begin{abstract}
Gaussian process (GP) regression is a popular surrogate modeling tool for computer simulations in engineering and scientific domains. 
However, it often struggles with high computational costs and low prediction accuracy when the simulation involves too many input variables. 
For some simulation models, the outputs may only be significantly influenced by a small subset of the input variables, referred to as the ``active variables''. 
We propose an optimal kernel learning approach to identify these active variables, thereby overcoming GP model limitations and enhancing system understanding.
Our method approximates the original GP model's covariance function through a convex combination of kernel functions, each utilizing low-dimensional subsets of input variables.
Inspired by the Fedorov-Wynn algorithm from optimal design literature, we develop an optimal kernel learning algorithm to determine this approximation. 
We incorporate the effect heredity principle, a concept borrowed from the field of ``design and analysis of experiments'', to facilitate efficient computation.
Through several examples, we demonstrate that the proposed method provides competitive prediction and variable selection, with favorable computational performance in the examples considered.
It is an effective solution for interpreting the surrogate GP regression and simplifying the complex underlying system.
\end{abstract}

\noindent%
{\it Keywords:} Optimal Design, Additive Gaussian Process Model, Functional ANOVA, Kernel Function.
\vfill

\spacingset{1}

\section{Introduction}\label{sec:intro}

Gaussian process (GP) regression is a powerful surrogate modeling approach for both computer and physical experiments characterized by highly nonlinear input-output relationships. 
However, like many supervised learning methods, GP models face the ``curse of dimensionality'' when dealing with high-dimensional input data and relatively small sample sizes--the so-called ``large $p$ small $n$'' problem. 
This challenge is particularly acute when estimating unknown parameters of an anisotropic covariance function in the GP model \citep{joseph2011regression,liu2020gaussian}.
Fortunately, many computer simulation models and physical systems are inherently influenced by only a small subset of input variables, known as ``active variables'' or ``active dimensions''. 
Identifying these variables can alleviate computational bottlenecks and improve the prediction accuracy of GP models. 

\subsection{Related Literature}\label{subset:literature}

Researchers have proposed various methods to address the issue of high input dimensionality in GP models. 
Each approach has unique strengths and limitations, with performance varying depending on the application and data. 
These approaches can be broadly categorized into three groups.

The first group of methods uses automatic relevance determination and sensitivity analysis \citep{moon2012two,borgonovo2014transformations, Iooss2017introduction}. 
The most influential variables are kept by ranking the input variables based on their influence on the response variable, and the rest are removed. 
However, they typically require an estimated surrogate model before the dimension reduction step. 

The second class of methods seeks to build the GP model on a linear transformation of the input space \citep{constantine2014active}, i.e., $Y(\bm x)=Z(\bm B^\top \bm x)+\epsilon$, where $Z(\cdot)$ is a GP with unknown parameters and $\bm B$ is full-rank and of dimension $p\times d$ such that $\bm B^\top $ projects the original $p$-dimensional vector $\bm x$ onto a lower $d$-dimensional space with $d<p$.
The number $d$ is usually user-specified. 
The column space of $\bm B$ is referred to as the ``active subspace'', which corresponds to the projected directions in which $\bm x$ has the most influence on $\bm y$.
It is preferred that the columns of $\bm B$ are orthonormal to avoid identifiability issues, and thus $\bm B$ is constrained to an appropriate manifold \citep{tripathy2016gaussian, seshadri2018dimension}. 
GP single-index models \citep{choi2011gaussian, gramacy2012gaussian} are a special case of the active subspace approach with $d=1$.
While effective, these methods often require significant computation and sometimes gradient information of the underlying unknown function \citep{constantine2014active, fukumizu2014gradient}.

The third group of methods is based on functional ANOVA decomposition, such as two recent works by \cite{borgonovo2018functional} and \cite{sung2019multiresolution}. 
As a popular statistical tool, the functional ANOVA decomposition provides an approximation of the underlying unknown function $Y(\bm x)$ as a linear combination of basis functions, each involving only one or two input variables. 
While this approach can handle high-dimensional input variables, including more complex basis functions inevitably increases computational demands.

Other approaches to high-dimensional regression include Bayesian tree-based methods, such as Dirichlet additive regression trees (DART) \citep{linero2018bayesian}, which encourage sparse variable selection. 
Recent developments in GP modeling also include sparse variational approximations \citep{hensman2013gaussian}, which can
be combined with automatic relevance determination, and deep Gaussian processes \citep{salimbeni2017doubly}, which model
complex relationships through compositions of GP layers.
These methods provide alternative strategies for modeling high-dimensional inputs, although they do not explicitly
construct sparse combinations of low-dimensional kernels as considered in this paper.
More recent approaches integrate variable selection directly into GP estimation, including penalized likelihood with
Vecchia approximations \citep{cao2022scalable}, spike-and-slab priors with scalable variational inference \citep{dance2022fast}, and GP-based feature selection using automatic relevance determination \citep{deng2024gaussian}.


\subsection{Our Contribution}
In this paper, we introduce a novel approach to building GP regression models. 
As reviewed in Section \ref{sec:review}, the computational bottleneck for estimating the GP model lies in the covariance kernel function.
The commonly used anisotropic covariance kernel function, also known as the product covariance function, contains the same number of unknown parameters as the input dimension. 
Therefore, compared to a low input dimension problem, with too many input variables, the optimization procedure to obtain the maximum likelihood estimates can require more computational effort, considering the matrix inversion and determinant are re-calculated in each optimization iteration. 
More importantly, the estimated parameters are likely to be locally optimal solutions due to the high dimensionality. 

Recognizing this bottleneck, we propose to approximate the covariance kernel function of the original high-dimensional input variables via a convex combination of kernel functions of lower-dimensional input variables. 
This is equivalent to approximating the original GP of the high-dimensional input space by functional ANOVA decomposition, which leads to an additive GP model with each added component being a GP of low-dimensional input space. 

The low-dimensional kernel functions are identified by minimizing the regularized loss function. 
As explained in Section \ref{sec:learning}, given a compact space of kernel functions, 
 \cite{micchelli2005learning} have proved the existence of the optimal convex combination of kernel functions, and there are only a finite number of kernel functions with non-zero weights. 
Inspired by a similar optimization problem and the Fedorov–Wynn-type of algorithm in \emph{optimal design} \citep{dean2015handbook, li2022maximin}, we propose an optimal kernel learning algorithm that sequentially selects kernels from the candidate set and simultaneously updates the weights of all selected kernels.
For high-dimensional input data, the candidate set of low-dimensional kernels can be large, so the computation needed to prepare the candidate kernels also increases drastically. 
To overcome this issue, we follow the \emph{effect heredity principle} from the design and analysis of experiments (DAE) literature \citep{hamada1992analysis, wu2011experiments} and construct the candidate set of kernel functions stage-wise based on the previously selected input variables. 
This work showcases the potential of applying traditional concepts from DAE literature to modern statistical learning problems, bridging established methodologies with contemporary challenges in data analysis and modeling.

The paper is organized as follows. 
In Section \ref{sec:review}, we review the GP regression from the parametric stochastic modeling perspective and nonparametric kernel method perspective. 
Section \ref{sec:learning} explains the discrete nature of the kernel learning problem. 
In Sections \ref{sec:opt} and \ref{sec:low}, we develop the forward stepwise optimal kernel learning approach with theoretical convergence and then use it to learn the optimal kernel from the set of low-dimensional basic kernel functions. 
In Section \ref{sec:num}, we demonstrate that the proposed method can accurately identify the active input dimensions and improve prediction accuracy. 
We conclude the paper in Section \ref{sec:end} with a discussion of future work. 

\section{Review of the GP Model}\label{sec:review}

GP regression is a nonparametric regression model with a parameterized covariance family.
Its simplest form is equivalent to the regularized reproducing kernel Hilbert space regression. 
In this section, we define the notation and review GP regression from both perspectives.  

\subsection{Stochastic GP model}\label{subsec:statGP}

A Gaussian process is a stochastic process or a random function such that every finite collection of function values evaluated at different input design points has a multivariate normal distribution.
Any GP can be completely characterized by a mean and a covariance function.
Given the input domain $\Omega \subset \mathbb{R}^p$, the simplest stationary GP model assumes the response $Y(\bm x)$ for any $\bm x\in \Omega$ as follows,
\begin{equation}\label{eq:GP}
Y(\bm x)=Z(\bm x)+\epsilon
\end{equation}
where $Z(\bm x)\sim GP(0,\tau^2K_{\bm \theta}(\cdot,\cdot))$ and $\epsilon\sim N(0,\sigma^2)$. 
The two are independent of each other.
The correlation function of $Z(\bm x)$ is $K_{\bm\theta}(\cdot,\cdot): \Omega \times \Omega \rightarrow \mathbb{R}$, a positive definite kernel function with unknown parameters $\bm \theta$. 
A commonly used kernel function is the separable Gaussian kernel function $K_{\bm\theta}(\bm x_1,\bm x_2)=\exp(-\sum_{i=1}^p\theta_i(x_{1,i}-x_{2,i})^2)$. 
These separable kernel functions, or product covariance functions, are \emph{anisotropic}. 
An \emph{isotropic} Gaussian kernel has the same $\theta_i$'s, i.e., $K_{\theta}(\bm x_1,\bm x_2)=\exp(-\theta\sum_{i=1}^p(x_{1,i}-x_{2,i})^2)$.
The methodology developed in this paper can be used for other kernel functions, such as the Mat\'{e}rn kernel function, a popular choice especially for spatial statistics.

Suppose we observe data $\{\bm x_i,y_i\}_{i=1}^n$. 
Denote $\mathcal{X}=\{\bm x_i\}_{i=1}^n\subset \Omega$ as $n$ design points, $\bm y=[y_1,\cdots,y_n]^\top$ is the vector of $n$ output observations, and $\bm K$ as the $n\times n$ kernel matrix with $K_{i,j}=K_{j,i}=K(\bm x_i,\bm x_j)$. 
Based on the assumption in \eqref{eq:GP}, the conditional distribution of $Y(\bm x)$ at any query point is $Y(\bm x)|\bm y, \bm \theta, \tau^2, \sigma^2 \sim \mathcal{N}(\hat{Y}(\bm x), \sigma^2_{\bm x|n}(\bm x))$. 
The conditional mean is 
\begin{equation}\label{GP-prediction}
\hat{Y}(\bm x)=\E(Y(\bm x)|\bm y)=\bm k(\bm x)^\top(\bm K+\eta\bm I_n)^{-1}\bm y,
\end{equation}
where $\bm k(x)=[K(x,x_1),\ldots,K(x,x_n)]^\top$ contains the correlations between $Y(x)$ and $Y(x_i)$, $\eta=\sigma^2/\tau^2$ is the noise-to-process variance ratio, and $\mathbf{I}_n$ is the $n\times n$ identity matrix. 
For deterministic computer experiments, $\sigma^2=0$ in the stochastic model and hence $\eta=0$. 
In practice, however, a small positive value of $\eta$ is often introduced as a nugget parameter to improve numerical stability.
The conditional variance is
\[
\sigma^2_{\bm x|n}(\bm x)=\var(Y(\bm x)|\bm y)=\tau^2\left\{1-\bm k(\bm x)^\top(\bm K+\eta \bm I_n)^{-1}\bm k(\bm x)\right\}+\sigma^2. 
\]
Based on the conditional distribution, the prediction and inference at any query point can be obtained. 

The unknown parameters $\tau^2$, $\sigma^2$ and $\bm\theta$ can be estimated using maximum likelihood estimation (MLE). 
For physical experiments or stochastic computer experiments, $\sigma^2$ is estimated from replicated output observations at the same design point.
For deterministic computer experiments, since no measurement noise is involved, $\sigma^2$ should be specified as $0$, and thus $\eta=0$. 
In this case, the predictor in \eqref{GP-prediction} interpolates all the training data.
However, it is usually advantageous to specify a small positive value for $\eta$ known as the \emph{nugget} value \citep{peng2014choice} to overcome the possible ill-conditioning of the kernel matrix $\bm K$. 
Given $\bm \theta$ and $\eta$, the MLE for $\tau^2$ is $\hat\tau^2=\frac{1}{n}\bm y^T(\bm K+\eta\bm I)^{-1}\bm y$. 
The MLEs of $\bm\theta$ and $\eta$ solve
\[\min_{\bm\theta, \eta} n\log\hat{\tau}^2+\log\det(\bm K+\eta\bm I).\]
The minimization is computationally demanding as it needs to compute $(\bm K+\eta\bm I)^{-1}$, $\log \det(\bm K+\eta\bm I)$, and the gradient of the objective function in each iteration. 
The complexity of evaluating the gradient of the objective ranges from $O((p+1)n^3)$ in naive implementation to $O(n^3+n^2p)$ in more sophisticated implementation, which increases with respect to the input dimension $p$.

\subsection{GP model as a regularized RKHS regression}\label{subsec:RKHS}

For every positive definite kernel function $K(\cdot,\cdot)$ defined on the domain $\Omega\subseteq\mathbb{R}^p$, let $\mathcal{F}$ be the set of functions $f:\Omega\rightarrow\mathbb{R}$ such that $f(\cdot)=\sum_{i=1}^n c_iK(\bm x_i,\cdot)$, for any positive integer $n$, any $\bm c\in\mathbb{R}^n$, and any $\mathcal{X}=\{\bm x_1,\cdots,\bm x_n\}\subseteq\Omega$.
Therefore, $\mathcal{F}$ is a linear space.
In addition, for any $f,g\in\mathcal{F}$ with $f=\sum_{i=1}^n c_iK(\bm x_i,\cdot)$ and $g=\sum_{j=1}^m d_j K(\bm y_j,\cdot)$, define the bilinear operator
\begin{equation} \label{inner-product}
\langle f,g \rangle=\sum_{i=1}^n\sum_{j=1}^m c_id_j K(\bm x_i,\bm y_j).
\end{equation}
It can be easily shown that \eqref{inner-product} defines an inner product on $\mathcal{F}$ \citep{wahba1990spline, wendland2004scattered}. 
The completion of $\mathcal{F}$ with the inner product forms a Hilbert space, which is called the reproducing kernel Hilbert space (RKHS) induced by the kernel $K(\cdot,\cdot)$. 
Conventionally, it is denoted by $\mathcal{H}_K$. 
It satisfies the following two conditions:
\begin{enumerate}
\item For any $\bm x\in\Omega$, we have $K(\bm x,\cdot)\in\mathcal{H}_K$;
\item For any $f\in\mathcal{H}_K$ and any $\bm x\in\Omega$, we have $\langle f,K(\bm x,\cdot)\rangle_{\mathcal{H}_K}=f(\bm x)$, also known as reproducing property. 
Here $\langle\cdot,\cdot\rangle_{\mathcal{H}_K}$ stands for the inner product of $\mathcal{H}_K$ that is deduced from the bilinear operator $\langle\cdot,\cdot\rangle$ defined by \eqref{inner-product}.
\end{enumerate}

Given data $\{\bm x_i,y_i\}_{i=1}^n$, for any kernel function $K(\cdot,\cdot)$ and $f\in\mathcal{H}_K$, define the regularized loss function 
\begin{equation*}
Q_{\eta}(f,K,\mathcal{X},\bm y) = Q(f,K,\mathcal{X},\bm y) + \eta \|f\|^2_{\mathcal{H}_K},
\end{equation*}
where $Q(f,K,\bm X,\bm y)$ is a user-specified loss function measuring the goodness-of-fit and $\eta>0$ is the regularization parameter. 
The penalized regression problem is to solve the following minimization problem
\begin{equation}\label{eq:minQ}
\min_{f\in\mathcal{H}_K}Q_{\eta}(f,K,\mathcal{X},\bm y). 
\end{equation}
By the Representer Theorem, a minimizer of \eqref{eq:minQ} can be written as $f(\bm x)=\sum_{i=1}^n c_i K(\bm x, \bm x_i)=\bm c^\top \bm k(\bm x)$.
Then \eqref{eq:minQ} is equivalent to
\begin{equation}\label{eq:minQ2} 
\min_{\bm c\in\mathbb{R}^n}Q_{\eta}(\bm c,K)=Q(\bm c,K)+\eta\bm c^\top \bm K\bm c,
\end{equation}
where $Q(\bm c,K)= Q(\bm c^\top \bm k(\bm x) ,K,\mathcal{X},\bm y)$ ($\mathcal{X}$ and $\bm y$ are dropped for simplicity).  
Define the vector $\bm f=[f(\bm x_1),\ldots, f(\bm x_n)]^\top$. 
It is a well-known result that when $Q(f,K)$ is the squared-error loss, i.e., $Q(f,K)=\|\bm y-\bm f\|_2^2$, the solution to \eqref{eq:minQ2} is equivalent to the predictor $\hat{Y}(\bm x)$ in \eqref{GP-prediction} obtained under the GP assumption, i.e., the optimal $\bm c^*=(\bm K+\eta \bm I_n)^{-1}\bm y$ and thus $\hat{f}(\bm x)=\hat{Y}(\bm x)$ \citep{wahba1990spline, girosi1995regularization}. 
Under the equivalence between GP prediction and regularized RKHS regression, the parameter $\eta$ corresponds to the
regularization parameter; in the stochastic GP formulation of Section~2.1, it is the noise-to-process variance ratio, or a user-specified nugget in the deterministic-computer-experiment setting.

\section{Discrete Nature of the Optimal Kernel}\label{sec:learning}

Suppose we can define a space of kernel functions, denoted by $\mathcal{K}$.
Given data, how should we find the optimal kernel function $K^*\in\mathcal{K}$ for a specific kernel learning method, such as the GP regression model or the support vector machine?
This general question is sometimes referred to as a multiple kernel learning (MKL) problem, a topic thoroughly discussed in a review by \cite{gonen2011multiple}.
In this paper, we narrow our focus on the special case of MKL for the GP model.

To determine the optimal $K^* \in\mathcal{K}$, we aim to minimize the regularized loss in \eqref{eq:minQ} or \eqref{eq:minQ2} w.r.t. $K$, expressed as:
\begin{equation}\label{eq:min-minQK}
Q_{\eta}(\mathcal{K})=\min_{K\in\mathcal{K}}Q_{\eta}(K),
\end{equation}
where $Q_{\eta}(K)=\min_{f\in\mathcal{H}_K}Q_{\eta}(f,K)$ represents the optimal solution of \eqref{eq:minQ2}.
The minimal solution in \eqref{eq:min-minQK} is attainable due to the property of $\mathcal{K}$, which is elaborated later. 

Consider the squared-error loss function $Q(f,K)=\|\bm y-\bm f\|^2_2$. 
Then, $Q_{\eta}(K)=\min_{f\in\mathcal{H}_K}Q_{\eta}(f,K)$ takes the form:
\begin{equation}\label{eq:min-minQL2}
 Q_{\eta}(K)=\min_{\bm c\in\mathbb{R}^n} \left\{ Q(\bm c,K)+\eta {\bm c}^\top \bm K\bm c \right\} =(\bm y-\bm K\bm c^*)^\top(\bm y-\bm K\bm c^*)+\eta {\bm c^*}^\top \bm K \bm c^*,
\end{equation}
where $\bm c^*=(\bm K+\eta \bm I_n)^{-1}\bm y$.
Substituting \eqref{eq:min-minQL2} into \eqref{eq:min-minQK}, the minimization problem to find optimal kernel is 
\begin{equation*}
Q_{\eta}(\mathcal{K})=\min_{K\in\mathcal{K}}\left\{ (\bm y-\bm K\bm c^*)^\top(\bm y-\bm K\bm c^*)+\eta {\bm c^*}^\top \bm K \bm c^*\right\}. 
\end{equation*}
According to \cite{argyriou2005learning} and \cite{micchelli2005learning}, the solution to the general problem \eqref{eq:min-minQK} has a discrete nature, even if $\mathcal{K}$ is a space of continuously parameterized kernel functions. 
To explain this point, we introduce some new notation and concepts.

Let $\mathcal{A}_+(\Omega)$ denote the set of kernel functions such that for any set of design points in $\Omega$ the resulting kernel matrix $\bm K$ is positive definite. 
If $\mathcal{K}$ is a compact and convex subset of $\mathcal{A}_+(\Omega)$ and $Q:\mathbb{R}^n\rightarrow \mathbb{R}$ is continuous and nonnegative, then the solution of \eqref{eq:min-minQK} exists. 
See Lemma 2 of \cite{micchelli2005learning}. 
Furthermore, let $\mathcal{G}\subset \mathcal{A}_{+}(\Omega)$ be a compact set of basic kernels and assume the loss function $Q:\mathbb{R}^n\rightarrow \mathbb{R}$ is continuous and $\eta>0$. 
If $\mathcal{K}$ is the closure of the convex hull of $\mathcal{G}$, denoted by $\overline{\text{conv}(\mathcal{G})}$, then there exists a subset $\mathcal{S}\subset \mathcal{G}$ that contains at most $n+2$ basic kernels from $\mathcal{G}$ such that $Q_{\eta}(\mathcal{K})$ admits a minimizer $K^*\in \text{conv}(\mathcal{S})$ and $Q_{\eta}(\text{conv}(\mathcal{S}))=Q_{\eta}(\mathcal{K})$, recalling the definition of $Q_{\eta}$ in \eqref{eq:min-minQK}.

In simpler terms, it implies that the optimal kernel $K^*$ solving $Q_{\eta}(\mathcal{K})$ is a convex combination of at most $n+2$ basic kernels from $\mathcal{G}$ when $\mathcal{K}$ is a closed convex hull of $\mathcal{G}$. 
An optimal kernel exists under the stated assumptions, but the optimal kernel and its mixture representation need not be unique.
This result reveals two key steps in learning the optimal kernel. 
\begin{enumerate}
\item[(1)] specify the compact set of basic kernel functions $\mathcal{G}$;
\item[(2)] given $\mathcal{G}$, identify the optimal convex combination of at most $n+2$ kernel functions from $\mathcal{G}$ by minimizing $Q_{\eta}(K)$ w.r.t. $K$, where $K\in \overline{\text{conv}(\mathcal{G})}$. 
\end{enumerate}
Each step presents unique challenges. 
As reviewed in Section \ref{subsec:RKHS}, one assumption of the RKHS regression is that the underlying function $f$ belongs to $\mathcal{H}_K$. 
Hence, the basic kernels of $\mathcal{G}$ should be chosen judiciously and $\mathcal{G}$ must be sufficiently large to ensure that the RKHS $\mathcal{H}_{K^*}$ induced by the optimal kernel $K^*$ includes the unknown underlying function $f$. 
Furthermore, given $\mathcal{G}$, a computationally feasible algorithm is required to solve \eqref{eq:min-minQK} to find the optimal convex combination. 
To tackle these challenges, we draw inspiration from the literature on optimal design, a sub-area within the field of ``design and analysis of experiments''.

\section{Learning the Optimal Kernel Function}\label{sec:opt}

This section focuses on learning the optimal kernel from a \emph{pre-defined} set of basic kernels $\mathcal{G}$, i.e., the step (2) described above.
The task of optimal kernel learning, given $\mathcal{G}$, closely resembles the optimal design problem.
We introduce the conventional notation from the optimal design literature to establish this connection.

\subsection{Optimal Kernel and Optimal Design}\label{subsec:optkernel}

In approximate design terms \citep{kiefer1974general, atkinson2014optimal}, a design $\xi$ belongs to a class $\Xi$ of probability measures on a compact design space $\mathcal{X}\subset \mathbb{R}^d$, and we require that $\Xi$ includes all discrete measures. 
Under the linear or generalized linear regression model with $\bm x$ as the input variables, let $M(\bm x)$ denote the information matrix at a design point $\bm x$. 
The information matrix of a design, $M(\xi)$, is defined as $M(\xi)=\int_{\mathcal{X}} M(\bm x)\xi(\dd \bm x)$.  
Common design criteria, such as $D$- and $I$-optimal criteria, which are convex in the information matrix $M$, are also convex in $\xi$ \citep{kiefer1974general}. 
The optimal design $\xi^*$ minimizing such a design criterion consists of $m$ support points $\{\bm x_1,\ldots, \bm x_m\}\subset \mathcal{X}$ and the optimal weights $\bm \lambda^*$, where $0<\lambda_i^*\leq 1$ and $\sum_{i=1}^m \lambda_i^* =1$.
Thus, $\lambda_i^*$ is the optimal probability mass allocated to each support point $\bm x_i^*$. 

Returning to the context of the optimal kernel, we introduce the concept of \emph{design}, which is a probability measure $\xi\in \Xi$, where $\Xi$ is a class of probability measures on the compact set of basic kernels $\mathcal{G}\subset \mathcal{A}_{+}(\Omega)$ including all discrete measures. 
Let $\mathcal{K}= \overline{\text{conv}(\mathcal{G})}$. 
It is evident that for any $K\in \mathcal{K}$, there exists a design $\xi\in \Xi$, such that $K=\int G \xi(\dd G)$, where $G$ is the notation for any kernel in $\mathcal{G}$, and vice versa. 
Note that such a relationship between kernel function and design is not necessarily one-to-one, but it does not affect the technical derivation. 
If $\mathcal{G}$ is a countable and compact set, i.e., $\mathcal{G}=\{G_1,G_2,\ldots \}$, then $K=\sum_{i=1}\xi_i G_i $, where $0\leq \xi_i\leq 1$ is the probability mass for $G_i$ and $\sum\xi_i=1$. 
We can denote a kernel function by $K(\xi)$ as a function of $\xi$ when emphasizing the connection between the kernel function and the design. 
Therefore, finding the optimal kernel is equivalent to finding the optimal design $\xi^*$ with $m$ \emph{support kernels} $\mathcal{S}=\{K_1,\ldots, K_m\}$ selected from $\mathcal{G}$, borrowing the term ``support points'' and the optimal weights $\bm \lambda^*$ corresponding to the support kernels. 
Here $0<\lambda_i^*\leq 1$ for $i=1,\ldots, m$, and $\sum_{i=1}^m \lambda_i^*=1$. 
The optimal kernel then can be expressed by $K(\xi^*)=\sum_{i=1}^m \lambda_i^*K_i$. 
Acknowledging the result reviewed in Section \ref{sec:learning}, $m\leq \min\{n+2,|\mathcal{G}|\}$, where $|\mathcal{G}|$ is the cardinality of $\mathcal{G}$ and it can be $\infty$. 

In the rest of the paper, we only consider the squared-error loss function and the corresponding $Q_{\eta}(K)$, since it is the most commonly used loss function for the GP regression model. 
The most direct formula of $Q_{\eta}(K)$ is given in \eqref{eq:min-minQL2}. 
Since any $K\in \overline{\text{conv}(\mathcal{G})}$ can be represented by a design $\xi \in \Xi$, $Q_{\eta}(K(\xi))$ can be seen as a functional of $\xi$, which we denote by $Q_{\eta}(\xi)$ for short. 
We are going to show that $Q_{\eta}(K)$ is convex in $K$ and thus also convex in $\xi$. 
Comparing optimal kernel and optimal design, we can conclude that they are essentially the same problem. 
The effort invested in connecting the two is worthwhile because the theories and algorithms for solving optimal design can also be adapted for optimal kernel learning. 

\subsection{General Equivalence Theorem}\label{subsec:gen_equ}

The General Equivalence Theorem \citep{kiefer1974general} is fundamental for optimal design. 
It provides crucial guidelines for constructing optimal designs for convex design criteria in $\xi$. 
The derivation of the General Equivalence Theorem varies based on the definition of the design criterion, as shown in works like \cite{atkinson2015designs}, \cite{yang2013optimal}, and \cite{li2022maximin}.
Inspired by the optimal design literature, we also derive the General Equivalence Theorem for optimal kernel learning. 
First, we define two versions of the directional derivative. 
One version is the directional derivative of loss function $Q_{\eta}(K)$ w.r.t. the kernel function $K$. 
Due to the relationship between any $K$ and design $\xi$, we can also define another version of the directional derivative of $Q_{\eta}(\xi)$ w.r.t. the design $\xi$. 
Proposition \ref{prop:dir-diff} gives the formula of the directional derivative of $Q_{\eta}(\xi)$. 
All results obtained here follow the notation and assumptions earlier. 
The proof and derivation are included in the Supplement. 

\begin{definition}[Directional Derivative w.r.t. Kernel]
Let $K$ and $K^\prime$ be two kernel functions in the kernel space $\mathcal{K}$, where $\mathcal{K}$ is a compact and convex subset of $\mathcal{A}_{+}(\Omega)$. 
The directional derivative of $Q_{\eta}(K)$ in the direction of $K^{\prime}$ is
\begin{equation*}
  \phi(K^{\prime},K):= \bigtriangledown_{K^{\prime}}Q_{\eta}(K)=\lim_{\alpha \to 0^+} \frac{Q_{\eta}((1-\alpha)K+\alpha K^{\prime})-Q_{\eta}(K)}{\alpha}.
\end{equation*}
\end{definition}

\begin{definition}[Directional Derivative w.r.t. Design]\label{def:dir-diff}
Given a compact set of kernel functions $\mathcal{G}\subset \mathcal{A}_{+}(\Omega)$, let $\xi$ and $\xi^\prime$ be two probability measures in $\Xi$ on $\mathcal{G}$, including all discrete measures.
As a function of $\xi$, the directional derivative of $Q_{\eta}(\xi)$ in the direction of $\xi^{\prime}$ is
\begin{equation*}
  \phi(\xi^{\prime},\xi):= \bigtriangledown_{\xi^{\prime}}Q_{\eta}(\xi)=\lim_{\alpha \to 0^+} \frac{Q_{\eta}((1-\alpha)\xi+\alpha \xi^{\prime})-Q_{\eta}(\xi)}{\alpha}.
\end{equation*}
\end{definition}

\begin{proposition}\label{prop:dir-diff}
The directional derivative of $Q_{\eta}(\xi)$ in the direction of $\xi^{\prime}$ is given as,
\begin{equation}\label{eq:dir-diff}
\phi (\xi^{\prime},\xi)=\frac{\partial Q_{\eta}(\xi)}{\partial \alpha}\bigg|_{\alpha =0}=-\eta \bm y^\top((\bm K_{\xi}+\eta \bm I_n)^{-1}(\bm K_{\xi^{\prime}}-\bm K_{\xi})(\bm K_{\xi}+\eta\bm I_n)^{-1})\bm y,
\end{equation}
where $\bm K_{\xi}$ and $\bm K_{\xi'}$ are the $n\times n$ kernel matrices computed by evaluating $K(\xi)$ and $K(\xi')$ on $\mathcal{X}=\{\bm x_i\}_{i=1}^n$. 
\end{proposition}

Based on Proposition 13 of \cite{micchelli2005learning}, if $Q:\mathbb{R}^n \rightarrow \mathbb{R}_+$ is convex, which is the case of the squared-error loss function and $\eta >0$, the functional $Q_{\eta}(K):\mathcal{A}_+(\Omega)\rightarrow \mathbb{R}_+$ is convex.
Lemma \ref{lemma:convex} here shows $Q_{\eta}(\xi)$ is convex in $\xi\in \Xi$ as well. 
With these results, we can obtain the General Equivalence Theorem for the optimal $\xi^*\in \Xi$ that minimizes $Q_{\eta}(\xi)$.

\begin{lemma}\label{lemma:convex}
Assume $Q_{\eta}(K):\mathcal{A}_+(\Omega)\rightarrow \mathbb{R}_+$ is convex in $\mathcal{K}$ where $\eta>0$ and $\mathcal{K}$ and $\mathcal{G}$ are defined in Section \ref{subsec:optkernel}. 
Then $Q_{\eta}(\xi)$ is also convex in $\Xi$. 
\end{lemma}

\begin{theorem}[General Equivalence Theorem]\label{thm:GE}
Assume the same definition of $\Xi$, $\mathcal{G}$, $\mathcal{K}$, and $Q_{\eta}(\cdot)$ in Section \ref{subsec:optkernel}. 
The following conditions of a design $\xi^* \in \Xi$ are equivalent:
\begin{enumerate}
\item[(1)] The design $\xi^*\in \Xi$ minimizes $Q_{\eta}(\xi)$;
\item[(2)] $\phi (\xi^{\prime},\xi^*)\geq 0$ holds for any $\xi^{\prime}\in \Xi$; 
\item[(3)] $\phi(G, \xi^*)\geq 0$ holds for any $G\in \mathcal{G}$ and the inequality becomes equality if $G$ is a support kernel of $\xi^*$. Here, the derivative $\phi(G,\xi)$ is a simplified notation for $\phi(\xi_G,\xi)$. The design $\xi_G$ is a probability measure assigning unit probability to the single kernel $G$ in $\mathcal{G}$.
\end{enumerate}
\end{theorem}

\subsection{Forward Stepwise Optimal Kernel Learning}\label{subsec:weights}

The General Equivalence Theorem \ref{thm:GE} provides some insight on how to select the support kernels sequentially. 
Based on it, we propose an optimization algorithm that iterates between selecting a kernel from $\mathcal{G}$ as a new support kernel and updating the weights, which can be considered as a Fedorov-Wynn type of algorithm \citep{dean2015handbook}.
In each iteration, we check the sign of $\phi(G,\xi^r)$ for any kernel function $G$ that has not been selected into the current design $\xi^r$.
If $\phi(G,\xi^r)$ is non-negative for all $G$, then $\xi^{r}$ is optimal. 
But if $\phi(G,\xi^r)<0$ for some $G$, it indicates that $G$ is a potential support kernel and should be added to the design. 
To achieve the maximum reduction of the loss function $Q_{\eta}(\xi^r)$ in each iteration, we add the kernel $G^{*}=\argmin_{G} \phi(G,\xi^r)$ with $\phi(G,\xi^r)<0$ into the current set of support kernels for $\xi^r$.
This procedure is outlined in Algorithm \ref{alg:fed-wynn}. 
Because it sequentially selects one support kernel from the set of the basic kernels, we call it the \emph{Forward Stepwise Optimal Kernel Learning Algorithm}.
We also show the convergence of Algorithm \ref{alg:fed-wynn} in Theorem \ref{thm:convergence} and it requires Lemma \ref{lem:inequality}.

\begin{lemma}\label{lem:inequality}
For any design $\xi\in \Xi$ and the optimal design $\xi^*$ that minimizes $Q_{\eta}(\xi)$, the following inequality holds: 
\begin{equation}\label{eq:ineq}
\min_{G\in \mathcal{G}} \phi(G, \xi)\leq \phi(\xi^*, \xi)\leq Q_{\eta}(\xi^*)-Q_{\eta}(\xi)\leq 0,
\end{equation}
where $\phi(G, \xi)$ and $\phi(\xi^*,\xi)$ are the directional derivative of $Q_{\eta}(\xi)$ defined in Definition \ref{def:dir-diff} and computed by Proposition \ref{prop:dir-diff}.
\end{lemma}


\begin{theorem}[Convergence of Algorithm
\ref{alg:fed-wynn}]\label{thm:convergence}
Let $\mathcal G\subset\mathcal A_+(\Omega)$ be compact and $\mathcal K=\overline{\operatorname{conv}(\mathcal G)}$.
For fixed $\eta>0$, consider Algorithm \ref{alg:fed-wynn} with exact weight optimization, without deletion or finite stopping rules. 
Then its objective values converge to the global minimum:
\[
\lim_{r\to\infty}Q_\eta(\xi^r)
=Q_\eta(\xi^*),
\]
where $\xi^*$ minimizes $Q_\eta$ over $\Xi$.
\end{theorem}

To achieve sparsity, we add an optional step at the end of Algorithm \ref{alg:fed-wynn}. 
We trim excessively small weights and the associated support kernels from the optimal design via hard thresholding.
The hard-thresholding step provides a simple way to obtain a sparse kernel representation. 
Since the kernel weights are nonnegative and sum to one, an $\ell_1$ penalty on the weights would be constant and thus would not induce sparsity.
More explicit sparsity constraints, such as cardinality constraints considered by \citet{bertsimas2025sparse}, offer an alternative direction for future research.

After a new support kernel is added to the current design, we need to update the weights of all the current support kernels. 
Due to the limited space, we move the introduction of finding the optimal weight procedure for given support kernels to Section S1 of the Supplementary Material. 
In Section S1, we first obtain the conditions of optimal weights presented in the following Corollary \ref{thm:weight} and then develop Algorithm \ref{alg:multiplicative} to update the weights given the support kernels. 

\begin{algorithm}[H]
\caption{Forward Stepwise Optimal Kernel Learning Algorithm\label{alg:fed-wynn}}
\begin{algorithmic}[1]
\State {\bf Setup:} Given data $\{\bm x_i, y\}_{i=1}^n$, construct the set of basic kernels $\mathcal{G}$, and set the following parameters.
\begin{itemize}
\item $\eta$: the regularization/nugget parameter;
\item \texttt{DEL}: a small positive value as the backward elimination threshold;
\item \texttt{Tol}: a small positive value to check convergence;
\item \texttt{MaxIter}: a large integer as the maximum iterations allowed.
\end{itemize} 
\State {\bf Initiate:} Randomly choose one kernel function from $\mathcal{G}$. Without loss of generality, denote it by $K_0$. At the step $r=0$, the design $\xi^{0}$ contains the support kernel $\mathcal{S}^{0}=\{K_0\}$ with weight $\bm \lambda^{(0)}=\{1\}$. Compute the objective function value $Q_{\eta}(\xi^{0})$ from \eqref{eq:min-minQL2}. 
\State Set counter $r=0$, \texttt{change}$=1$, and $m=|\mathcal{S}^{r}|=1$. 
\While {\texttt{change} $> $ \texttt{Tol} and $r < $ \texttt{MaxIter}}
\If{$\phi(G, \xi^{r})\geq 0$ for any $G\in \mathcal{G}\setminus \mathcal{S}^{r}$} Break;
\Else
\State Add the kernel $K_{r+1}=\argmin_{G\in \mathcal{G}\setminus\mathcal{S}^{r}} \phi(G,\xi^{r})$ and $\phi(K_{r+1},\xi^{r})<0$ to the set of the support kernels of the current design, i.e., $\mathcal{S}^{r+1}=\mathcal{S}^{r}\cup \{K_{r+1}\}$, where $\phi(\cdot,\xi^{r})$ is calculated by \eqref{eq:dir-diff} in Proposition \ref{prop:dir-diff}. 
\State $m \leftarrow m+1$
\State $r \leftarrow r+1$
\State Obtain optimal weights $\bm \lambda^{(r)}$ for the support kernels in $\mathcal{S}^{r}$ using Algorithm \ref{alg:multiplicative}. The current design $\xi^{r}$ consists of support kernels $\mathcal{S}^{r}$ and weights $\bm \lambda^{(r)}$. 
\State Compute $Q_{\eta}(\xi^{r})$ and 
\texttt{change}$=\bigg|\frac{Q_\eta(\xi^{r})-Q_\eta(\xi^{r-1})}{Q_\eta(\xi^{r-1})}\bigg|$
\EndIf
\EndWhile
\State {\bf Delete (Optional):} Given $\xi^{r}$ with support kernels $\mathcal{S}^{r}$ and weights $\bm \lambda^{(r)}$, remove the support kernels whose weights satisfy $\lambda_{i}^{(r)}<\texttt{DEL}$. Update the design $\xi^r$ by re-scaling the remaining weights such that their sum is equal to 1. 
\end{algorithmic}
\end{algorithm}

\section{Set of Basic Kernels: Low-Dimensional Kernel Functions}\label{sec:low}
 
In this section, we answer the question of how to construct the set of basic kernels $\mathcal{G}$ to identify the active input dimensions, i.e., the step (1) described at the end of Section \ref{sec:learning}.

\subsection{ANOVA Model}\label{subsec:anova}
Functional Analysis of Variance model \citep{huang1998functional, huang1998projection, hooker2007generalized} takes the following general model assumption
\begin{equation}\label{ANOVA}
y(\bm x)=\sum_{j=1}^p Z_j(x_{j}) + \sum_{j=1}^{p-1}\sum_{k=j+1}^p Z_{jk}(x_{j}, x_{k}) +\ldots+ \epsilon,
\end{equation}
where each $Z_j(x_j)$ is the 1-dim sub-model for the $j$-th dimension of $\bm x$, and $Z_{jk}(x_j,x_k)$ is the 2-dim sub-model for the $j$-th and $k$-th dimensions of $\bm x$, and so on.
It is common to restrict the model \eqref{ANOVA} to contain sub-models up to certain lower input dimensions.
We focus on the special case where each $Z(\cdot)$ in \eqref{ANOVA} is a low-dimensional Gaussian process.
Such a model is a special case of \emph{additive GP} and has been studied in, for example, \cite{kaufman2010bayesian}, \cite{duvenaud2011additive}, \cite{durrande2012additive}, \cite{durrande2013anova}, etc.
Many believe such a model works only when the underlying response function has a coordinate-wise additive structure (see e.g.,\cite{tripathy2016gaussian}).
However, as is shown in \cite{durrande2012additive} and \cite{duvenaud2011additive}, functional ANOVA GP models can recover arbitrarily flexible underlying data relationships, regardless of whether an additive structure is present.
Under most circumstances, it is sufficient to include only the low-dimensional sub-models such as 1-dim and 2-dim ones in \eqref{ANOVA}.
The rationale is analogous to the traditional ANOVA employed in the analysis of experiments, namely the \emph{effect sparsity principle} \citep{wu2011experiments}.
It states that lower-order terms are the more likely to be statistically significant than higher-order terms in the analysis of experimental data.

One major limitation of \eqref{ANOVA} is that only one GP sub-model is included for each subset of input variables, such as $\{x_j\}$, $\{x_j,x_k\}$, etc. 
In \cite{ba2012composite}, the authors observed that it is advantageous to model the covariance function as a summation of two covariance functions that are used to capture the local and global variations of the underlying function.
Motivated by this idea, instead of specifying a single GP model for each subset of input variables, we propose supplying multiple ones to allow for more flexibility, and generalize model \eqref{ANOVA} to the following form:
\begin{equation}\label{ANOVA-generalized}
y(\bm x)=\sum_{j=1}^p\sum_{m=1}^{M_j} \beta_{j,m} Z_{j,m}(x_j) + \sum_{j=1}^{p-1}\sum_{k=j+1}^{p}\sum_{m=1}^{M_{jk}} \beta_{j,k,m} Z_{j,k,m}(x_j,x_k) +\cdots +\epsilon,
\end{equation}
Here each $Z_{j,m}(x_j)$ is a GP with input $x_j$ and $Z_{j,m} \sim GP(0, \tau_{j,m}^2 K_{j,m})$ where $K_{j,m}$ is a kernel function of input variable $x_j$. 
Similarly, $Z_{j,k,m}(\bm x_j,\bm x_k)$ is a GP with input $(x_j,x_k)$ and  $Z_{j,k,m} \sim GP(0, \tau_{j,k,m}^2 K_{j,k,m})$ where $K_{j,k,m}$ is a kernel function of input variables $(x_j,x_k)$. 
All the $Z_{j,m}$ and $Z_{j,k,m}$ are mutually independent. 
Other GP models in the omitted part in \eqref{ANOVA-generalized} are defined in the same way. 
Throughout, we assume that the training design has pairwise distinct values in each input coordinate,
which can be easily achieved by designs such as Latin hypercube design.
Consequently, the Gaussian basic kernels on any nonempty subset of input variables yield strictly positive definite kernel matrices on the training design.

From the additive GP model \cite{ba2012composite}, we know that if $Z(\bm x)\sim GP(0, \tau^2 K(\cdot,\cdot))$ can be considered as the sum of two different GPs, i.e., $Z(\bm x)=Z_1(\bm x)+Z_2(\bm x)$, where $Z_i(\bm x) \sim GP(0,\tau_i^2 K_i(\cdot,\cdot))$ for $i=1,2$ and are mutually independent, then equivalently, 
\[\tau^2 K(\cdot,\cdot)=\tau_1^2 K_1(\cdot,\cdot)+\tau_2^2 K_2(\cdot,\cdot).\]
In other words, $K=\lambda_1 K_1+\lambda_2 K_2$ where $\lambda_1+\lambda_2=1$ and $\lambda_i=\tau_i^2/\tau^2$ for $i=1,2$. 
This result applies to the more general case of \eqref{ANOVA-generalized} and we have the following equivalence. 
\begin{theorem}
Suppose the component processes in \eqref{ANOVA-generalized} are mutually
independent, with $Z_*(\cdot)\sim GP(0,\tau_*^2K_*)$.
Define
\[
\tau^2=\sum_*\beta_*^2\tau_*^2>0,
\qquad
\lambda_*=\frac{\beta_*^2\tau_*^2}{\tau^2},
\]
where $*$ represents all component indices. 
Then the latent process $Z(\boldsymbol{x})$ in \eqref{ANOVA-generalized} is a Gaussian process $GP(0,\tau^2K)$, where
\[
K= \sum_{j=1}^{p}\sum_{m=1}^{M_j} \lambda_{j,m}K_{j,m} + \sum_{j=1}^{p 1}\sum_{k=j+1}^{p}\sum_{m=1}^{M_{jk}} \lambda_{j,k,m}K_{j,k,m} +\cdots, \qquad \sum_*\lambda_*=1.
\]
Equivalently,
\[
Z(\boldsymbol{x})\overset{d}{=} \sum_{j=1}^{p}\sum_{m=1}^{M_j} \sqrt{\lambda_{j,m}}\,\widetilde Z_{j,m}(x_j) + \sum_{j=1}^{p 1}\sum_{k=j+1}^{p}\sum_{m=1}^{M_{jk}} \sqrt{\lambda_{j,k,m}}\, \widetilde Z_{j,k,m}(x_j,x_k)
+\cdots,
\]
where the processes $\widetilde Z_*(\cdot)\sim GP(0,\tau^2K_*)$ are mutually independent, and $\overset{d}{=}$ denotes equality in distribution as stochastic processes.
\end{theorem}

We aim to build a generalized functional ANOVA model in \eqref{ANOVA-generalized}.
The set of basic kernels for Algorithm \ref{alg:fed-wynn} is 
$\mathcal{G}=(\cup_{j=1}^p\mathcal{G}_j)\cup(\cup_{j\neq k}^p\mathcal{G}_{jk})\cup\cdots$, where $\mathcal{G}_j$ consists of $M_j$ kernel functions defined on the $j$th dimension $x_j$, $\mathcal{G}_{jk}$ consists of $M_{jk}$ kernel functions defined on $(x_j,x_k)$, and so on.
So $\mathcal{G}$ has a finite number of basic kernels, and thus is a compact set in $\mathcal{A}_{+}(\Omega)$. 
In this paper, we choose the isotropic Gaussian kernel as follows.
\begin{align*}
\mathcal{G}_j&=\left\{K_{j,m}(x_{1,j},x_{2,j})=\exp(-\theta_{j,m}(x_{1,j}-x_{2,j})^2), \text{where } \theta_{j,m} \in \Theta, m=1,\cdots, M_j\right\}\\
\mathcal{G}_{jk}&=\left\{K_{j,k,m}((x_{1,j},x_{1,k}),(x_{2,j},x_{2,k}))=\exp(-\theta_{j,k,m}[(x_{1,j}- x_{2,j})^2+(x_{1,k}-x_{2,k})^2]),\right. \\
&\left. \text{where } \theta_{j,k,m} \in \Theta, m=1,\cdots,M_{jk}\right\}.
\end{align*}
Here $\Theta$ is a pre-defined set of values for the parameter $\theta$. 
In our examples, we set
\[
\Theta= \left\{a\times 10^b: a\in\{1,3,5,7,9\},\ b\in\{-2,-1,0,1,2,3\} \right\} \cup\{10000,30000\},
\]
for all the input data that have been standardized before modeling and use the same $M=32$.
Users can set $M$ and $\Theta$ based on their data and available computing resources. 
Other radial basis kernel functions can be used if the Gaussian kernel is too smooth for the underlying function. 
We prefer the radial basis function because it significantly reduces the number of basic kernels. 
In Section~\ref{Ssec:kernelpara}, we have given a more detailed discussion on isotropic vs. anisotropic kernels and compared their performances via a small example. 
What is more, we have also studied how the setting of $\Theta$ and its size can affect \texttt{optK}.

\subsection{Forward Stage-wise Optimal Kernel Learning}\label{subsec:heredity}

If we only include 1-dim and 2-dim basic kernel functions in \eqref{ANOVA-generalized} and set all $M_j$ and $M_{jk}$ the same at $M=32$, then $\mathcal{G}$ has size $p\times M+\binom{p}{2}*M$. 
If the 3-dim basic kernel functions are considered too, the size of $\mathcal{G}$ is even larger. 
For large $p$ and $M$, this number can be prohibitive. 
To overcome the issue of large number of basic kernels, we propose selecting support kernels by running the Algorithm \ref{alg:fed-wynn} in sequential stages. 
Our inspiration comes from the \emph{effect heredity principle} in \cite{wu2011experiments}. 
It’s widely used in the design and analysis of experiments and other areas like in \cite{haris2016convex}.
The \emph{strong} heredity principle states that if an interaction term of two variables is statistically significant, both main effects of the two variables must be included in the model.
The \emph{weak} heredity principle only requires one of the two main effects to be included. 

For stage-wise optimal kernel learning, we first consider only 1-dim kernels $\mathcal{G}=(\cup_{j=1}^p\mathcal{G}_j)$.  
After running Algorithm \ref{alg:fed-wynn}, we identify the active variables that are the input dimensions of the selected basic kernels.
In the second stage, we only add 2-dim basic kernels of the active variables and proceed with Algorithm \ref{alg:fed-wynn}. 
The strong heredity requires both input variables of a 2-dim kernel to be active in the first stage, while the weak heredity only requires one. 
So the strong heredity reduces the number of basic kernels in the second stage. 
We can continue to the future stages to include basic kernels of higher dimensional variables if needed.

The effect sparsity and effect heredity principles play complementary roles in our approach. 
The effect sparsity principle motivates representing the response using a relatively small number of influential input variables and low-dimensional kernel components. 
To further encourage sparsity, we remove kernels with sufficiently small weights through the thresholding procedure in Algorithm \ref{alg:fed-wynn}. 
The effect heredity principle, on the other hand, provides a computationally efficient strategy for constructing candidate kernels involving interactions among input variables. 
By restricting higher-order candidate kernels according to the variables selected at previous stages, we substantially reduce the candidate search space. 
Thus, heredity is adopted primarily to improve computational efficiency, while sparsity remains a central objective of the
proposed method.

The procedure is summarized in Algorithm \ref{alg:stage-opt}. 
For short, we denote the proposed additive GP regression model \eqref{ANOVA-generalized} estimated via Algorithm \ref{alg:stage-opt} by \texttt{optK}. 
All the proposed algorithms are implemented in \texttt{R} and are available at \url{https://github.com/tonyxms/ConvexKernel}. 
Our implementation uses the \texttt{R} packages \texttt{foreach} \citep{foreach} and \texttt{doParallel} \citep{doParallel} to boost computing speed.

\begin{algorithm}[H]
\caption{Forward Stage-wise Optimal Kernel Learning \label{alg:stage-opt}}
\begin{algorithmic}[1]
\State {\bf Setup:}  Given data $\{\bm x_i,y_i\}_{i=1}^n$ and set the following parameters for the algorithm.
\begin{itemize}
\item \texttt{MaxDim}: maximum dimensions of input variables allowed for the basic kernel functions; 
\item $\eta$, \texttt{DEL}, \texttt{Tol}, and \texttt{MaxIter}: parameters of Algorithm \ref{alg:fed-wynn};
\item $\delta$, \texttt{Tol}, \texttt{MaxIter}$_0$: parameters of Algorithm \ref{alg:multiplicative}. 
\end{itemize} 
\State {\bf Initiate:} construct the set $\Theta$ and the set $\mathcal{G}=(\cup_{j=1}^p\mathcal{G}_j)$ of basic kernels of $x_j$ for $j=1,\ldots,p$. The current dimension of kernels is \texttt{Dim}=1. 
\State Run Algorithm \ref{alg:fed-wynn} and return $Q_{\eta}(\xi)$, where $\xi$ is the current optimal design. The set $\mathcal{S}$ is the set of selected support kernels and $\bm \lambda$ as the vector of non-zero weights. 
\State Identify the active variables based on the selected kernels in $\mathcal{S}$. For the same $\Theta$, prepare the $\mathcal{G}$ to include the kernels involving (\texttt{Dim}+1) variables that satisfy the specified heredity rule. If the weak heredity principle is used, inactive variables are included too. Set \texttt{Dim}=\texttt{Dim}+1. 
\State Stop if $\texttt{Dim}>\texttt{MaxDim}$. Otherwise, retain previously selected kernels and resume Algorithm \ref{alg:fed-wynn} with the enlarged dictionary.
\State Stop if the relative change in $Q_\eta$ between stages is less than \texttt{Tol}; otherwise, return to Step 4.
\end{algorithmic}
\end{algorithm}

\subsection{Computational Complexity}\label{subsec:complexity}

The main computational cost of Algorithm \ref{alg:fed-wynn} comes from evaluating the directional derivatives over candidate
kernels and repeatedly updating the support-kernel weights.
Importantly, evaluating the directional derivative in \eqref{eq:dir-diff} does not require a separate matrix inversion for
each candidate. 
At iteration $r$, let $\bm{\alpha}_r=(\mathbf K_{\xi^r}+\eta\mathbf I_n)^{-1} \mathbf y$. 
This vector is computed once using a Cholesky factorization, at a cost of $O(n^3)$ for dense matrices, and is reused for all candidates.
For a candidate kernel $G$ with matrix $\mathbf K_G$, the directional derivative becomes
\[
\phi(G,\xi^r) =-\eta\left(\bm{\alpha}_r^\top\mathbf K_G\bm{\alpha}_r -\bm{\alpha}_r^\top\mathbf K_{\xi^r}\bm{\alpha}_r \right).
\]
Once $\bm{\alpha}_r$ is available, evaluating this quantity requires $O(n^2)$ operations per dense candidate matrix. 
Thus, screening $N_r$ candidates at iteration $r$ requires $O(n^3+N_r n^2)$ operations, excluding kernel construction and the subsequent optimization of support-kernel weights.

The dependence on the input dimension $d$ enters primarily through the number of eligible candidate kernels. 
With a fixed parameter grid $\Theta$ and basic kernels involving at most $q$ input variables, unrestricted enumeration can produce $O(|\Theta|d^q)$ candidates for fixed $q$.
The forward stage-wise procedure and the heredity principle restrict the candidate sets at later stages, reducing this computational burden.
Nevertheless, dense matrix operations and repeated weight updates remain limiting factors as $n$ and the candidate dictionary grow. 
Empirical computation times and search iteration counts are reported in Section~\ref{sec:num}.

\section{Examples}\label{sec:num}
We examine prediction, variable selection and computational cost through two synthetic examples based on well-known test functions from the literature and a satellite-drag simulation case study. 
All numerical experiments are conducted in \texttt{R} \citep{R2026} and \texttt{Python} \citep{python2026}, with the same training and test data for each replication across all methods.
Implementation details and additional results are given in Section \ref{Ssec:extrasimulation} of the Supplementary Material.
The study comprises $21$ configurations, each with $B=100$ planned replications. 
Training and test data are regenerated for each replication in the two synthetic examples. For Satellite, training observations are sampled without replacement from a fixed pool, and the supplied physical test inputs and responses are reused, with irrelevant inputs generated anew for each replication. 
We compare three versions of the proposed method \texttt{optK} with the following seven alternative methods. 
\begin{itemize}[leftmargin=*]
\item \texttt{mlegp}: maximum-likelihood GP regression with a separable Gaussian correlation function \citep{santner2003design}, implemented in the R package \texttt{mlegp} \citep{dancik2008mlegp}.
\item \texttt{laGP}: local GP approximation \citep{gramacy2015local} implemented in the R package \texttt{laGP} \citep{gramacy2016lagp}
\item \texttt{MRFA}: multi-resolution functional ANOVA \citep{sung2019multiresolution}, implemented by \texttt{MRFA} Package in R, with predictions taken at the minimum-penalty endpoint of the fitted regularization path.
\item \texttt{DART}: Dirichlet additive regression trees \citep{linero2018bayesian}, implemented with \texttt{BART::wbart} in R using \texttt{sparse=TRUE} \citep{sparapani2021nonparametric}.
\item \texttt{SVGP-ARD}: a sparse variational GP with an automatic relevance determination Gaussian kernel \citep{hensman2013gaussian}, implemented in Python with GPyTorch \citep{gardner2018gpytorch} and PyTorch \citep{paszke2019pytorch}.
\item \texttt{DGP-2}: a two-layer variational GP \citep{salimbeni2017doubly}, implemented in GPyTorch with a five-output hidden GP layer, a scalar output layer and ARD Gaussian kernels.
\item \texttt{AS-GP}: a projection GP in the dimension-reduction framework of \citep{tripathy2016gaussian}, implemented in Python with PyTorch and Pymanopt \citep{townsend2016pymanopt} to jointly optimize the projection and GP parameters. 
The projection dimension is selected by cross-validation using the training data.
\end{itemize}
Detailed implementation settings, tuning procedures, and variable-selection rules of these alternative methods are provided in Section \ref{ssec:s3-implementation} of the Supplementary Material.

{\bf Algorithm Settings of \texttt{optK}:} The three proposed \texttt{optK} versions use isotropic Gaussian basic kernels and the same 32-value dictionary
\[
\Theta =\left\{a\times 10^b: a\in\{1,3,5,7,9\},\ b\in\{-2,-1,0,1,2,3\}\right\} \cup\{10000,30000\}.
\]
Strong heredity is used to construct the basic kernel set in Algorithm \ref{alg:stage-opt}.
For the tuning parameter $\eta$, we use a leave-one-out cross-validation procedure to select the optimal value from $\{0.005, 0.01, 0.02, 0.05, 0.1, 0.5\}$.
Please see Section \ref{Ssec:tuningpara} for more details related to tuning parameters. 
In Algorithm \ref{alg:fed-wynn}, we set $\texttt{Tol}=0.005$ and $\texttt{MaxIter}=1000$.
In Algorithm \ref{alg:multiplicative}, we set $\delta=1$, $\texttt{Tol=0.005}$, and $\texttt{MaxIter}_0=1000$. 
Both algorithms share the same maximum allowed iterations. 
In Algorithm \ref{alg:stage-opt}, we set $\texttt{MaxDim}=4$. 

The version \texttt{optK}\textsubscript{0.005} uses deletion threshold $\texttt{DEL}=0.005$ in Algorithm \ref{alg:fed-wynn}.
This small threshold may retain kernels with small weights and introduce false-positive variable selections, motivating the additional pruning step at the end of the search.
The version \texttt{optK-pruned} uses this same $\texttt{DEL}=0.005$, but removes support kernels with $\lambda<0.02$ once Algorithm \ref{alg:stage-opt} finishes. 
The final model is returned after re-weighting using the optimal weight procedure (Algorithm \ref{alg:multiplicative}). 
The version \texttt{optK}\textsubscript{0.05} instead uses $\texttt{DEL}=0.05$ for more aggressive deletion during the search, without the additional pruning step. 
The subscripts thus describe search deletion thresholds, not regularization parameter $\eta$.

For all methods under comparison, we have always scaled the training and testing designs in the same $[0,1]^d$ domain before calling the estimation and prediction procedure. 

{\bf Performance Metrics:} To evaluate the performance of the different methods, we compute the standardized root mean squared error (SRMSE) for a test sample of size $m$, defined as:
\begin{equation}\label{eq:SRMSE}
\mathrm{SRMSE}=\sqrt{\frac{\sum_{i=1}^m(\hat y_i-y_i)^2}
{\sum_{i=1}^m(y_i-\bar y_{\rm test})^2}}.
\end{equation}
The response scale is the original scale. 
Test responses are noiseless values for all three examples.
To assess the variable-selection performance, we report the number of false negatives (FN) and false positives (FP) for each method.
FN counts missed inputs in the analytic function and FP counts selected irrelevant inputs. 
For Michalewicz example, the functional input variables are treated as reference active inputs. 
For the Borehole and Satellite examples, the physical inputs described in Table~\ref{table-4} and Section~\ref{sec:satellite}, respectively, are treated as the reference active inputs when computing FN and FP. 
Although some reference active variables may have only a weak influence on the response, we still use them as the reference set for computing FP and FN, since identifying truly influential variables is challenging in practice. 
Indeed, identifying such influential variables is one of the primary objectives of our proposed methods.
Some methods learn continuous kernel parameters or projection subspaces without directly identifying individual inputs as active variables, and thus the FN and FP are reported as NA in tables and were omitted in figures. 
Some methods need an additional decision rule to decide whether an input is an active variable, which is given in Section \ref{ssec:s3-implementation}. 
Tables show means, with sample standard deviations in parentheses, across the $B=100$ replications.
The column Iter reports the search iteration count of each replication's own selected-$\eta$ candidate at 1-core, separately for each optK version.
The weight updates for the additional pruning step in \texttt{optK-pruned} are excluded. 
Other methods have no entry in this column.

{\bf Computation and time:} The experiments used CPU-only servers with AMD EPYC 7763 processors, combining R and Python implementations. 
Each method task received a CPU budget with different cores $C=1,2,4$.
The same data and initialization plans were used across CPU budgets.
Prediction, variable selection and Iter use $C=1$ while $C=2,4$ provide timing comparisons on the same replications.
The main tables also report wall-clock training time in seconds for each method with the optimal tuning parameter already chosen. 
Using more CPU cores does not necessarily reduce elapsed time, because parallelization can introduce overhead and some implementations make limited or no use of additional cores.
Details of parallel execution are provided in Section \ref{ssec:s3-implementation} of the Supplementary Material.

Figures \ref{fig:main-michal2-n1000}--\ref{fig:main-satellite-o2} and \ref{fig:s-case01}--\ref{fig:s-case14} summarize prediction accuracy, computation time, and variable selection across 100 replications. 
Boxes indicate the inter-quartile range, horizontal lines the medians, white diamonds the arithmetic means, and whiskers extend to the most extreme observations within 1.5 inter-quartile ranges of the box; all outliers are shown. 
SRMSE and wall-clock time use logarithmic axes. 
Prediction and variable-selection results use the one-core runs, while timing results compare one-, two-, and four-core budgets. 
FP and FN are shown only for methods with prescribed binary selection rules; blank rows indicate inapplicability. 
FN panels are omitted when all evaluated FN counts are zero such as in Figures \ref{fig:main-michal2-n1000} and \ref{fig:main-michal6-n500}. 
In all the tables, prediction, FN/FP and Iter use 1-core. 
An em dash in Iter denotes inapplicability. 

\subsection{Michalewicz Function}\label{sec:michal}
We use the positive-sign Michalewicz family \citep{simulationlib}
\begin{equation}\label{eq:michal-new}
f_{\bm a}(\bm x)=\sum_{j=1}^{p}\sin(x_j)
\left[\sin\!\left(\frac{a_jx_j^2}{\pi}\right)\right]^{20},
\qquad \bm x\in[0,\pi]^p.
\end{equation}
Here, $p$ denotes the number of active inputs. 
We consider $\bm a=(1,2)$ for $p=2$ and $\bm a=(1,2,1,2,1,2)$ for $p=6$.
To assess prediction and variable selection in the presence of irrelevant inputs, we embed each function in a $d$-dimensional input space containing $p$ active and $d-p$ irrelevant inputs.
For each replication, we generate a maximin Latin hypercube training design of size $n$ and a random Latin hypercube test design of size $m=3481$, both on $[0,1]^d$.
The function coordinates in \eqref{eq:michal-new} are $x_j=\pi u_j$, where $u_j$ denotes the unit-cube coordinate. 
Independent noise is added to training responses only.

The two-variable function is evaluated at $d=6$ with $n=200,500,1000$, and the six-variable function at $d=10,20,60$ with $n=300,500$. Each configuration uses $B=100$ replications.
Figure~\ref{fig:michale2d} illustrates the true function surface for $p=2$ and its \texttt{optK} prediction surface based on $n=200$ training observations.\par
\begingroup
\begin{figure}[H]\centering
\includegraphics[width=0.49\linewidth]{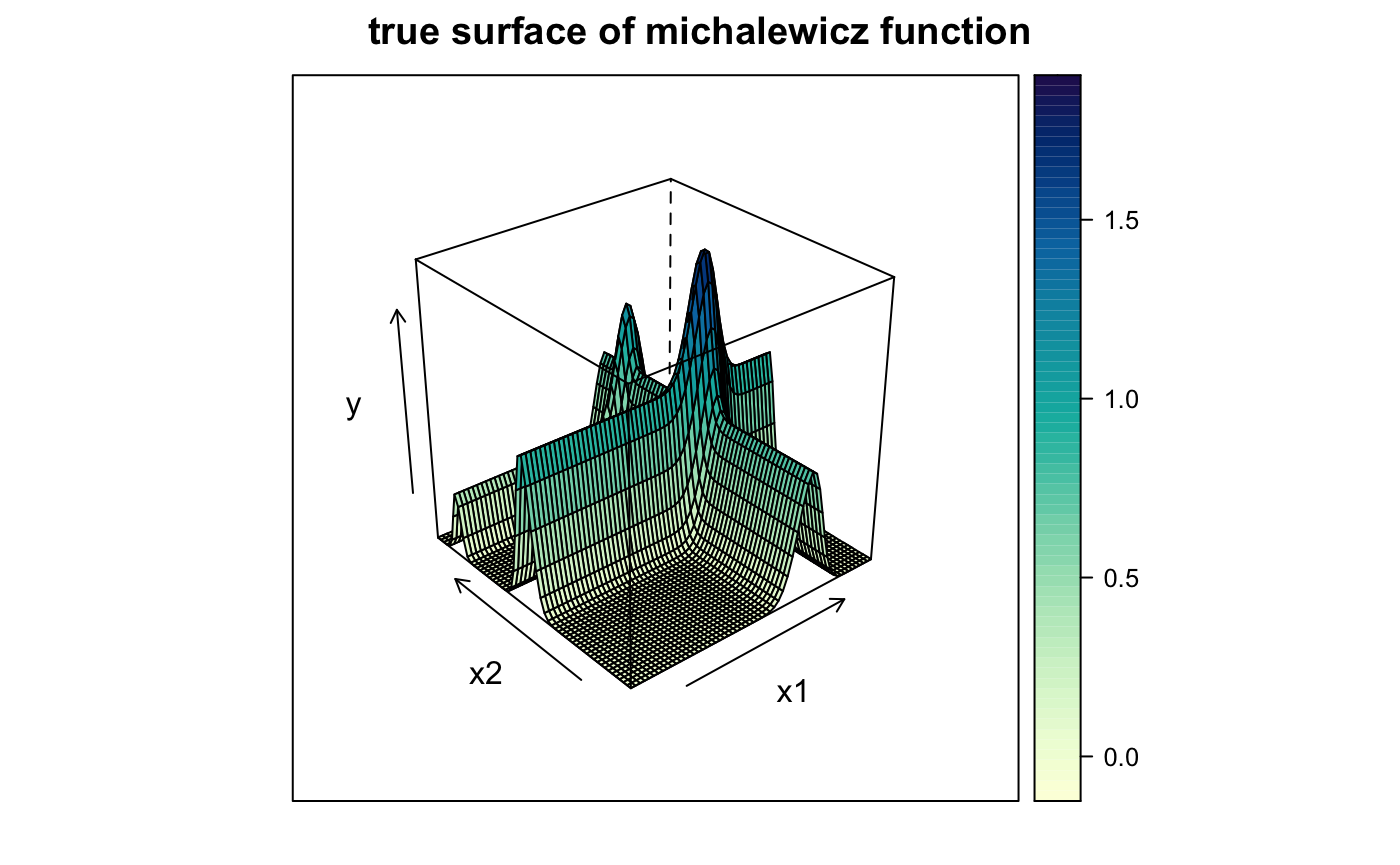}\hfill
\includegraphics[width=0.49\linewidth]{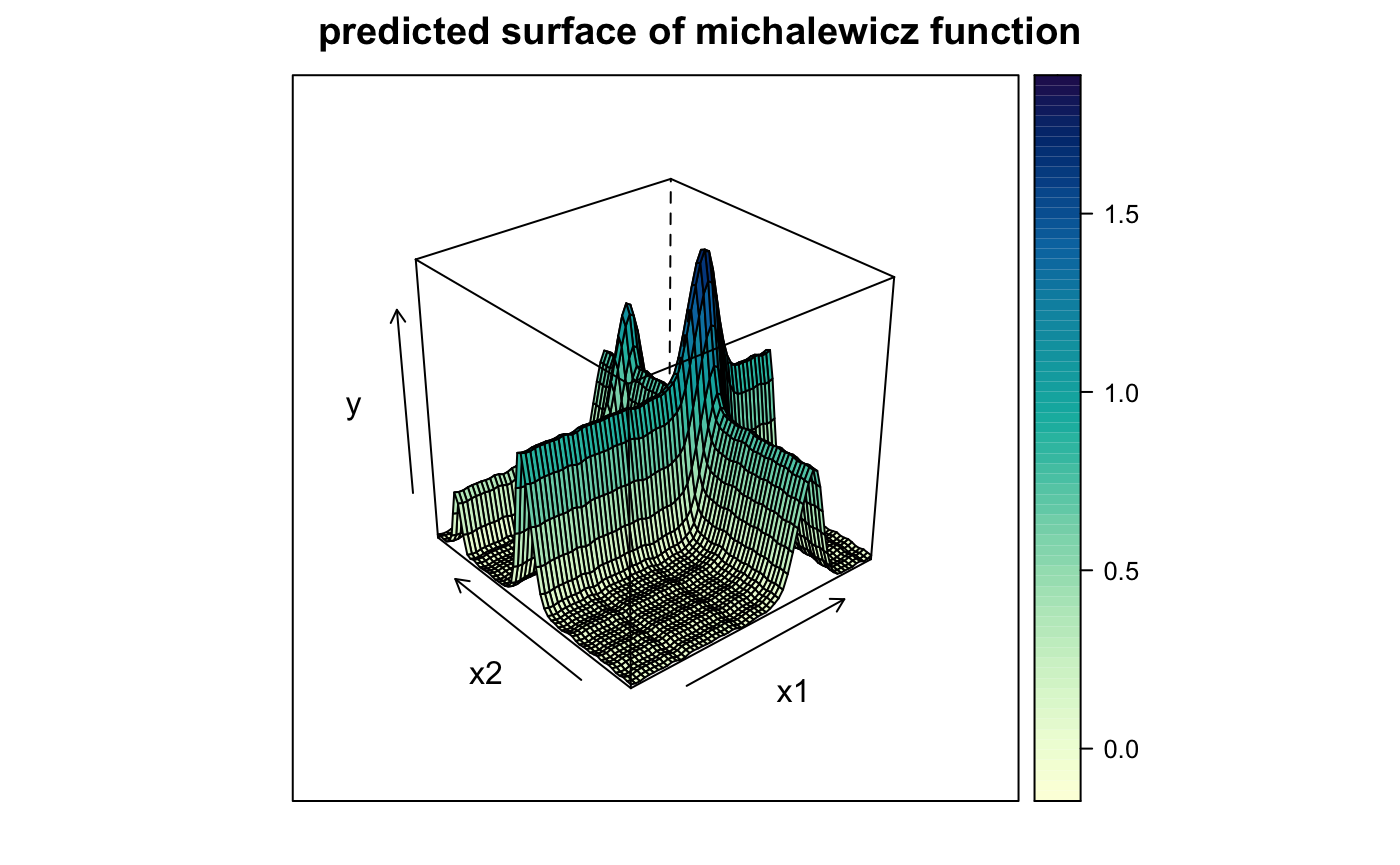}
\caption{The true (left) and predicted surface by \texttt{optK} (right) of the Michalewicz function.}\label{fig:michale2d}
\end{figure}
\endgroup

For $p=2$ and $d=6$, Table~\ref{tab:michal2} reports results for $n=200,500,1000$. 
The three optK versions achieve the three lowest mean SRMSEs at each sample size and retain both active inputs in every replication.
Their variable-selection performance differs mainly in false positives: \texttt{optK}\textsubscript{0.05} has mean FP counts close to zero, whereas \texttt{optK}\textsubscript{0.005} retains more irrelevant inputs.
Terminal pruning reduces false positives relative to \texttt{optK}\textsubscript{0.005}, but does not eliminate them.
DART also achieves FN$=$FP$=0$ at $n=200$, although its mean SRMSE is higher than those of the three optK versions.
Detailed distributions are in Figure~\ref{fig:main-michal2-n1000} for $n=1000$ and Figures~\ref{fig:s-case01} and~\ref{fig:s-case02} for $n=200,500$.
\par\begingroup\spacingset{1}\footnotesize
\setlength{\tabcolsep}{2pt}\renewcommand{\arraystretch}{1}
\def\ptmeasure#1#2{\settowidth{\dimen0}{#2}\ifdim\dimen0>#1\relax\edef#1{\the\dimen0}\fi}
\def\ptMethodWidth{0pt}
\ptmeasure\ptMethodWidth{Method}
\ptmeasure\ptMethodWidth{\texttt{AS-GP}}
\ptmeasure\ptMethodWidth{\texttt{DART}}
\ptmeasure\ptMethodWidth{\texttt{DGP-2}}
\ptmeasure\ptMethodWidth{\texttt{MRFA}}
\ptmeasure\ptMethodWidth{\texttt{SVGP-ARD}}
\ptmeasure\ptMethodWidth{\texttt{laGP}}
\ptmeasure\ptMethodWidth{\texttt{mlegp}}
\ptmeasure\ptMethodWidth{\texttt{optK-pruned}}
\ptmeasure\ptMethodWidth{\texttt{optK}\textsubscript{0.005}}
\ptmeasure\ptMethodWidth{\texttt{optK}\textsubscript{0.05}}
\def\ptSrmseWidth{0pt}
\ptmeasure\ptSrmseWidth{SRMSE}
\ptmeasure\ptSrmseWidth{\mbox{0.0141 (0.00115)}}
\ptmeasure\ptSrmseWidth{\mbox{0.0141 (0.00132)}}
\ptmeasure\ptSrmseWidth{\mbox{0.0146 (0.00141)}}
\ptmeasure\ptSrmseWidth{\mbox{0.0185 (0.00167)}}
\ptmeasure\ptSrmseWidth{\mbox{0.0189 (0.00160)}}
\ptmeasure\ptSrmseWidth{\mbox{0.0204 (0.00184)}}
\ptmeasure\ptSrmseWidth{\mbox{0.0318 (0.00387)}}
\ptmeasure\ptSrmseWidth{\mbox{0.0328 (0.00491)}}
\ptmeasure\ptSrmseWidth{\mbox{0.0340 (0.00376)}}
\ptmeasure\ptSrmseWidth{\mbox{0.0355 (0.0367)}}
\ptmeasure\ptSrmseWidth{\mbox{0.0376 (0.00262)}}
\ptmeasure\ptSrmseWidth{\mbox{0.0455 (0.00397)}}
\ptmeasure\ptSrmseWidth{\mbox{0.0461 (0.00385)}}
\ptmeasure\ptSrmseWidth{\mbox{0.0469 (0.00598)}}
\ptmeasure\ptSrmseWidth{\mbox{0.0644 (0.00150)}}
\ptmeasure\ptSrmseWidth{\mbox{0.0653 (0.0173)}}
\ptmeasure\ptSrmseWidth{\mbox{0.0658 (0.0530)}}
\ptmeasure\ptSrmseWidth{\mbox{0.0682 (0.00221)}}
\ptmeasure\ptSrmseWidth{\mbox{0.0810 (0.00409)}}
\ptmeasure\ptSrmseWidth{\mbox{0.0958 (0.0105)}}
\ptmeasure\ptSrmseWidth{\mbox{0.0978 (0.00996)}}
\ptmeasure\ptSrmseWidth{\mbox{0.135 (0.183)}}
\ptmeasure\ptSrmseWidth{\mbox{0.231 (0.0430)}}
\ptmeasure\ptSrmseWidth{\mbox{0.267 (0.0245)}}
\ptmeasure\ptSrmseWidth{\mbox{0.366 (0.0856)}}
\ptmeasure\ptSrmseWidth{\mbox{0.367 (0.198)}}
\ptmeasure\ptSrmseWidth{\mbox{0.460 (0.102)}}
\ptmeasure\ptSrmseWidth{\mbox{0.570 (0.123)}}
\ptmeasure\ptSrmseWidth{\mbox{0.585 (0.119)}}
\ptmeasure\ptSrmseWidth{\mbox{0.681 (0.145)}}
\def\ptFnWidth{0pt}
\ptmeasure\ptFnWidth{FN}
\ptmeasure\ptFnWidth{\mbox{0 (0)}}
\ptmeasure\ptFnWidth{\mbox{NA}}
\def\ptFpWidth{0pt}
\ptmeasure\ptFpWidth{FP}
\ptmeasure\ptFpWidth{\mbox{0 (0)}}
\ptmeasure\ptFpWidth{\mbox{0.01 (0.10)}}
\ptmeasure\ptFpWidth{\mbox{0.02 (0.14)}}
\ptmeasure\ptFpWidth{\mbox{0.03 (0.22)}}
\ptmeasure\ptFpWidth{\mbox{0.15 (0.39)}}
\ptmeasure\ptFpWidth{\mbox{0.30 (0.66)}}
\ptmeasure\ptFpWidth{\mbox{0.52 (0.59)}}
\ptmeasure\ptFpWidth{\mbox{0.77 (1.10)}}
\ptmeasure\ptFpWidth{\mbox{1.14 (1.03)}}
\ptmeasure\ptFpWidth{\mbox{1.26 (1.12)}}
\ptmeasure\ptFpWidth{\mbox{1.56 (1.24)}}
\ptmeasure\ptFpWidth{\mbox{1.67 (1.30)}}
\ptmeasure\ptFpWidth{\mbox{1.97 (1.60)}}
\ptmeasure\ptFpWidth{\mbox{2.17 (0.85)}}
\ptmeasure\ptFpWidth{\mbox{2.99 (0.75)}}
\ptmeasure\ptFpWidth{\mbox{3.50 (0.60)}}
\ptmeasure\ptFpWidth{\mbox{NA}}
\def\ptIterWidth{0pt}
\ptmeasure\ptIterWidth{Iter}
\ptmeasure\ptIterWidth{\mbox{---}}
\ptmeasure\ptIterWidth{\mbox{10.9 (3.4)}}
\ptmeasure\ptIterWidth{\mbox{11.5 (3.6)}}
\ptmeasure\ptIterWidth{\mbox{12.8 (6.3)}}
\ptmeasure\ptIterWidth{\mbox{14.8 (6.4)}}
\ptmeasure\ptIterWidth{\mbox{15.5 (6.6)}}
\ptmeasure\ptIterWidth{\mbox{16.2 (6.5)}}
\ptmeasure\ptIterWidth{\mbox{6.1 (1.0)}}
\ptmeasure\ptIterWidth{\mbox{6.3 (0.9)}}
\ptmeasure\ptIterWidth{\mbox{6.7 (1.0)}}
\def\ptTimeWidth{0pt}
\ptmeasure\ptTimeWidth{\makecell[r]{1-core\\(Sec)}}
\ptmeasure\ptTimeWidth{\makecell[r]{2-core\\(Sec)}}
\ptmeasure\ptTimeWidth{\makecell[r]{4-core\\(Sec)}}
\ptmeasure\ptTimeWidth{\mbox{10.9 (3.80)}}
\ptmeasure\ptTimeWidth{\mbox{1020 (861)}}
\ptmeasure\ptTimeWidth{\mbox{11.0 (3.48)}}
\ptmeasure\ptTimeWidth{\mbox{113 (10.5)}}
\ptmeasure\ptTimeWidth{\mbox{1170 (150)}}
\ptmeasure\ptTimeWidth{\mbox{12.0 (4.01)}}
\ptmeasure\ptTimeWidth{\mbox{12.9 (8.51)}}
\ptmeasure\ptTimeWidth{\mbox{12200 (2830)}}
\ptmeasure\ptTimeWidth{\mbox{127 (27.1)}}
\ptmeasure\ptTimeWidth{\mbox{128 (24.7)}}
\ptmeasure\ptTimeWidth{\mbox{1300 (313)}}
\ptmeasure\ptTimeWidth{\mbox{136 (18.4)}}
\ptmeasure\ptTimeWidth{\mbox{143 (14.6)}}
\ptmeasure\ptTimeWidth{\mbox{145 (22.6)}}
\ptmeasure\ptTimeWidth{\mbox{1480 (814)}}
\ptmeasure\ptTimeWidth{\mbox{1490 (689)}}
\ptmeasure\ptTimeWidth{\mbox{15.1 (9.01)}}
\ptmeasure\ptTimeWidth{\mbox{16.7 (3.93)}}
\ptmeasure\ptTimeWidth{\mbox{16.8 (3.13)}}
\ptmeasure\ptTimeWidth{\mbox{17.0 (4.24)}}
\ptmeasure\ptTimeWidth{\mbox{17.6 (3.15)}}
\ptmeasure\ptTimeWidth{\mbox{174 (14.9)}}
\ptmeasure\ptTimeWidth{\mbox{18.4 (11.5)}}
\ptmeasure\ptTimeWidth{\mbox{193 (26.5)}}
\ptmeasure\ptTimeWidth{\mbox{196 (46.5)}}
\ptmeasure\ptTimeWidth{\mbox{2.05 (0.988)}}
\ptmeasure\ptTimeWidth{\mbox{2.71 (1.13)}}
\ptmeasure\ptTimeWidth{\mbox{2.72 (1.02)}}
\ptmeasure\ptTimeWidth{\mbox{2.83 (1.16)}}
\ptmeasure\ptTimeWidth{\mbox{20.8 (18.5)}}
\ptmeasure\ptTimeWidth{\mbox{20.9 (12.1)}}
\ptmeasure\ptTimeWidth{\mbox{21.7 (12.1)}}
\ptmeasure\ptTimeWidth{\mbox{22200 (5290)}}
\ptmeasure\ptTimeWidth{\mbox{2320 (428)}}
\ptmeasure\ptTimeWidth{\mbox{239 (54.9)}}
\ptmeasure\ptTimeWidth{\mbox{242 (27.3)}}
\ptmeasure\ptTimeWidth{\mbox{287 (35.6)}}
\ptmeasure\ptTimeWidth{\mbox{290 (53.2)}}
\ptmeasure\ptTimeWidth{\mbox{2990 (785)}}
\ptmeasure\ptTimeWidth{\mbox{3.25 (1.29)}}
\ptmeasure\ptTimeWidth{\mbox{3.28 (1.20)}}
\ptmeasure\ptTimeWidth{\mbox{3.55 (2.11)}}
\ptmeasure\ptTimeWidth{\mbox{3.77 (2.64)}}
\ptmeasure\ptTimeWidth{\mbox{30.3 (4.19)}}
\ptmeasure\ptTimeWidth{\mbox{3080 (1310)}}
\ptmeasure\ptTimeWidth{\mbox{32.6 (2.75)}}
\ptmeasure\ptTimeWidth{\mbox{33.0 (6.45)}}
\ptmeasure\ptTimeWidth{\mbox{34.6 (4.52)}}
\ptmeasure\ptTimeWidth{\mbox{35.1 (3.12)}}
\ptmeasure\ptTimeWidth{\mbox{375 (48.2)}}
\ptmeasure\ptTimeWidth{\mbox{39.4 (9.10)}}
\ptmeasure\ptTimeWidth{\mbox{392 (890)}}
\ptmeasure\ptTimeWidth{\mbox{4.58 (0.926)}}
\ptmeasure\ptTimeWidth{\mbox{410 (556)}}
\ptmeasure\ptTimeWidth{\mbox{412 (411)}}
\ptmeasure\ptTimeWidth{\mbox{44300 (9430)}}
\ptmeasure\ptTimeWidth{\mbox{455 (861)}}
\ptmeasure\ptTimeWidth{\mbox{465 (933)}}
\ptmeasure\ptTimeWidth{\mbox{4810 (708)}}
\ptmeasure\ptTimeWidth{\mbox{4830 (947)}}
\ptmeasure\ptTimeWidth{\mbox{5220 (618)}}
\ptmeasure\ptTimeWidth{\mbox{523 (40.3)}}
\ptmeasure\ptTimeWidth{\mbox{534 (123)}}
\ptmeasure\ptTimeWidth{\mbox{5370 (1410)}}
\ptmeasure\ptTimeWidth{\mbox{552 (38.0)}}
\ptmeasure\ptTimeWidth{\mbox{6.32 (2.15)}}
\ptmeasure\ptTimeWidth{\mbox{6.35 (1.71)}}
\ptmeasure\ptTimeWidth{\mbox{60.9 (9.81)}}
\ptmeasure\ptTimeWidth{\mbox{653 (199)}}
\ptmeasure\ptTimeWidth{\mbox{654 (127)}}
\ptmeasure\ptTimeWidth{\mbox{676 (390)}}
\ptmeasure\ptTimeWidth{\mbox{69.9 (10.4)}}
\ptmeasure\ptTimeWidth{\mbox{72.3 (45.3)}}
\ptmeasure\ptTimeWidth{\mbox{73.5 (45.6)}}
\ptmeasure\ptTimeWidth{\mbox{73.6 (43.4)}}
\ptmeasure\ptTimeWidth{\mbox{76.7 (55.9)}}
\ptmeasure\ptTimeWidth{\mbox{76.8 (47.8)}}
\ptmeasure\ptTimeWidth{\mbox{798 (211)}}
\ptmeasure\ptTimeWidth{\mbox{8.33 (6.16)}}
\ptmeasure\ptTimeWidth{\mbox{8.84 (7.03)}}
\ptmeasure\ptTimeWidth{\mbox{8.96 (6.71)}}
\ptmeasure\ptTimeWidth{\mbox{81.9 (48.0)}}
\ptmeasure\ptTimeWidth{\mbox{82.0 (28.4)}}
\ptmeasure\ptTimeWidth{\mbox{83.3 (15.9)}}
\ptmeasure\ptTimeWidth{\mbox{86.6 (33.1)}}
\ptmeasure\ptTimeWidth{\mbox{864 (1840)}}
\ptmeasure\ptTimeWidth{\mbox{895 (482)}}
\ptmeasure\ptTimeWidth{\mbox{9.08 (6.71)}}
\ptmeasure\ptTimeWidth{\mbox{95.3 (34.2)}}
\ptmeasure\ptTimeWidth{\mbox{9920 (1940)}}
\def\ptUpdateWidth{\edef\ptTableWidth{\the\dimexpr\ptMethodWidth+\ptSrmseWidth+\ptFnWidth+\ptFpWidth+\ptIterWidth+\ptTimeWidth+\ptTimeWidth+\ptTimeWidth+14\tabcolsep\relax}}
\ptUpdateWidth\ifdim\ptTableWidth>\linewidth\setlength{\tabcolsep}{1pt}\ptUpdateWidth\fi
\typeout{PERFORMANCE-WIDTH main_michal2: table=\ptTableWidth; time=\ptTimeWidth; linewidth=\the\linewidth}
\ifdim\ptTableWidth>\linewidth\PackageError{performance-tables}{Table exceeds linewidth}{Check the table layout.}\fi
\setlength{\LTcapwidth}{\ptTableWidth}
\Needspace{570pt}
\begin{longtable}{@{}>{\raggedright\arraybackslash}p{\ptMethodWidth}>{\raggedleft\arraybackslash}p{\ptSrmseWidth}>{\raggedleft\arraybackslash}p{\ptFnWidth}>{\raggedleft\arraybackslash}p{\ptFpWidth}>{\raggedleft\arraybackslash}p{\ptIterWidth}>{\raggedleft\arraybackslash}p{\ptTimeWidth}>{\raggedleft\arraybackslash}p{\ptTimeWidth}>{\raggedleft\arraybackslash}p{\ptTimeWidth}@{}}
\caption{Two-variable Michalewicz function, $\bm a=(1,2)$; $d=6$, $p=2$, $m=3481$, $B=100$.}\label{tab:michal2}\\
\toprule
Method & SRMSE & FN & FP & Iter & \makecell[r]{1-core\\(Sec)} & \makecell[r]{2-core\\(Sec)} & \makecell[r]{4-core\\(Sec)} \\
\midrule\endfirsthead
\caption[]{Two-variable Michalewicz function, $\bm a=(1,2)$; $d=6$, $p=2$, $m=3481$, $B=100$. (continued).}\\\toprule
Method & SRMSE & FN & FP & Iter & \makecell[r]{1-core\\(Sec)} & \makecell[r]{2-core\\(Sec)} & \makecell[r]{4-core\\(Sec)} \\
\midrule\endhead
\midrule
\multicolumn{8}{@{}p{\ptTableWidth}@{}}{\footnotesize Values are reported as mean (sample SD). Prediction, FN/FP and Iter use 1-core. An em dash in Iter denotes inapplicability. NA denotes no approved binary selection rule.}\\
\multicolumn{8}{r@{}}{\footnotesize Continued on next page.}\\\endfoot
\bottomrule
\endlastfoot
\multicolumn{8}{@{}l}{\textbf{$n=200$}}\\*
\texttt{optK}\textsubscript{0.005} & \mbox{0.0340 (0.00376)} & \mbox{0 (0)} & \mbox{1.67 (1.30)} & \mbox{11.5 (3.6)} & \mbox{3.77 (2.64)} & \mbox{8.33 (6.16)} & \mbox{9.08 (6.71)} \\*
\texttt{optK-pruned} & \mbox{0.0328 (0.00491)} & \mbox{0 (0)} & \mbox{0.30 (0.66)} & \mbox{10.9 (3.4)} & \mbox{3.55 (2.11)} & \mbox{8.84 (7.03)} & \mbox{8.96 (6.71)} \\*
\texttt{optK}\textsubscript{0.05} & \mbox{0.0318 (0.00387)} & \mbox{0 (0)} & \mbox{0.01 (0.10)} & \mbox{6.1 (1.0)} & \mbox{2.05 (0.988)} & \mbox{3.28 (1.20)} & \mbox{3.25 (1.29)} \\*
\texttt{laGP} & \mbox{0.681 (0.145)} & \mbox{NA} & \mbox{NA} & \mbox{---} & \mbox{2.83 (1.16)} & \mbox{2.72 (1.02)} & \mbox{2.71 (1.13)} \\*
\texttt{mlegp} & \mbox{0.267 (0.0245)} & \mbox{0 (0)} & \mbox{0.52 (0.59)} & \mbox{---} & \mbox{35.1 (3.12)} & \mbox{32.6 (2.75)} & \mbox{33.0 (6.45)} \\*
\texttt{MRFA} & \mbox{0.0653 (0.0173)} & \mbox{0 (0)} & \mbox{2.17 (0.85)} & \mbox{---} & \mbox{16.7 (3.93)} & \mbox{34.6 (4.52)} & \mbox{30.3 (4.19)} \\*
\texttt{DART} & \mbox{0.0810 (0.00409)} & \mbox{0 (0)} & \mbox{0 (0)} & \mbox{---} & \mbox{242 (27.3)} & \mbox{127 (27.1)} & \mbox{113 (10.5)} \\*
\texttt{SVGP-ARD} & \mbox{0.570 (0.123)} & \mbox{NA} & \mbox{NA} & \mbox{---} & \mbox{196 (46.5)} & \mbox{392 (890)} & \mbox{410 (556)} \\*
\texttt{DGP-2} & \mbox{0.135 (0.183)} & \mbox{NA} & \mbox{NA} & \mbox{---} & \mbox{798 (211)} & \mbox{653 (199)} & \mbox{676 (390)} \\*
\texttt{AS-GP} & \mbox{0.367 (0.198)} & \mbox{NA} & \mbox{NA} & \mbox{---} & \mbox{2320 (428)} & \mbox{1300 (313)} & \mbox{654 (127)} \\
\midrule
\multicolumn{8}{@{}l}{\textbf{$n=500$}}\\*
\texttt{optK}\textsubscript{0.005} & \mbox{0.0204 (0.00184)} & \mbox{0 (0)} & \mbox{1.97 (1.60)} & \mbox{14.8 (6.4)} & \mbox{15.1 (9.01)} & \mbox{21.7 (12.1)} & \mbox{20.9 (12.1)} \\*
\texttt{optK-pruned} & \mbox{0.0189 (0.00160)} & \mbox{0 (0)} & \mbox{0.77 (1.10)} & \mbox{12.8 (6.3)} & \mbox{12.9 (8.51)} & \mbox{20.8 (18.5)} & \mbox{18.4 (11.5)} \\*
\texttt{optK}\textsubscript{0.05} & \mbox{0.0185 (0.00167)} & \mbox{0 (0)} & \mbox{0.01 (0.10)} & \mbox{6.3 (0.9)} & \mbox{4.58 (0.926)} & \mbox{6.35 (1.71)} & \mbox{6.32 (2.15)} \\*
\texttt{laGP} & \mbox{0.366 (0.0856)} & \mbox{NA} & \mbox{NA} & \mbox{---} & \mbox{12.0 (4.01)} & \mbox{11.0 (3.48)} & \mbox{10.9 (3.80)} \\*
\texttt{mlegp} & \mbox{0.0958 (0.0105)} & \mbox{0 (0)} & \mbox{0.03 (0.22)} & \mbox{---} & \mbox{552 (38.0)} & \mbox{523 (40.3)} & \mbox{534 (123)} \\*
\texttt{MRFA} & \mbox{0.0658 (0.0530)} & \mbox{0 (0)} & \mbox{2.99 (0.75)} & \mbox{---} & \mbox{39.4 (9.10)} & \mbox{69.9 (10.4)} & \mbox{60.9 (9.81)} \\*
\texttt{DART} & \mbox{0.0682 (0.00221)} & \mbox{0 (0)} & \mbox{0.15 (0.39)} & \mbox{---} & \mbox{287 (35.6)} & \mbox{143 (14.6)} & \mbox{136 (18.4)} \\*
\texttt{SVGP-ARD} & \mbox{0.460 (0.102)} & \mbox{NA} & \mbox{NA} & \mbox{---} & \mbox{239 (54.9)} & \mbox{465 (933)} & \mbox{412 (411)} \\*
\texttt{DGP-2} & \mbox{0.0469 (0.00598)} & \mbox{NA} & \mbox{NA} & \mbox{---} & \mbox{1170 (150)} & \mbox{1020 (861)} & \mbox{895 (482)} \\*
\texttt{AS-GP} & \mbox{0.0978 (0.00996)} & \mbox{NA} & \mbox{NA} & \mbox{---} & \mbox{9920 (1940)} & \mbox{5370 (1410)} & \mbox{2990 (785)} \\
\midrule
\multicolumn{8}{@{}l}{\textbf{$n=1000$}}\\*
\texttt{optK}\textsubscript{0.005} & \mbox{0.0146 (0.00141)} & \mbox{0 (0)} & \mbox{1.56 (1.24)} & \mbox{16.2 (6.5)} & \mbox{73.5 (45.6)} & \mbox{81.9 (48.0)} & \mbox{73.6 (43.4)} \\*
\texttt{optK-pruned} & \mbox{0.0141 (0.00132)} & \mbox{0 (0)} & \mbox{1.26 (1.12)} & \mbox{15.5 (6.6)} & \mbox{76.7 (55.9)} & \mbox{76.8 (47.8)} & \mbox{72.3 (45.3)} \\*
\texttt{optK}\textsubscript{0.05} & \mbox{0.0141 (0.00115)} & \mbox{0 (0)} & \mbox{0.02 (0.14)} & \mbox{6.7 (1.0)} & \mbox{16.8 (3.13)} & \mbox{17.6 (3.15)} & \mbox{17.0 (4.24)} \\*
\texttt{laGP} & \mbox{0.231 (0.0430)} & \mbox{NA} & \mbox{NA} & \mbox{---} & \mbox{95.3 (34.2)} & \mbox{86.6 (33.1)} & \mbox{82.0 (28.4)} \\*
\texttt{mlegp} & \mbox{0.0455 (0.00397)} & \mbox{0 (0)} & \mbox{0 (0)} & \mbox{---} & \mbox{5220 (618)} & \mbox{4810 (708)} & \mbox{4830 (947)} \\*
\texttt{MRFA} & \mbox{0.0355 (0.0367)} & \mbox{0 (0)} & \mbox{3.50 (0.60)} & \mbox{---} & \mbox{83.3 (15.9)} & \mbox{145 (22.6)} & \mbox{128 (24.7)} \\*
\texttt{DART} & \mbox{0.0644 (0.00150)} & \mbox{0 (0)} & \mbox{1.14 (1.03)} & \mbox{---} & \mbox{375 (48.2)} & \mbox{193 (26.5)} & \mbox{174 (14.9)} \\*
\texttt{SVGP-ARD} & \mbox{0.585 (0.119)} & \mbox{NA} & \mbox{NA} & \mbox{---} & \mbox{290 (53.2)} & \mbox{455 (861)} & \mbox{864 (1840)} \\*
\texttt{DGP-2} & \mbox{0.0376 (0.00262)} & \mbox{NA} & \mbox{NA} & \mbox{---} & \mbox{3080 (1310)} & \mbox{1490 (689)} & \mbox{1480 (814)} \\*
\texttt{AS-GP} & \mbox{0.0461 (0.00385)} & \mbox{NA} & \mbox{NA} & \mbox{---} & \mbox{44300 (9430)} & \mbox{22200 (5290)} & \mbox{12200 (2830)} \\*
\end{longtable}\endgroup

\begingroup\spacingset{1}
\begin{figure}[H]\centering
\includegraphics[width=\linewidth]{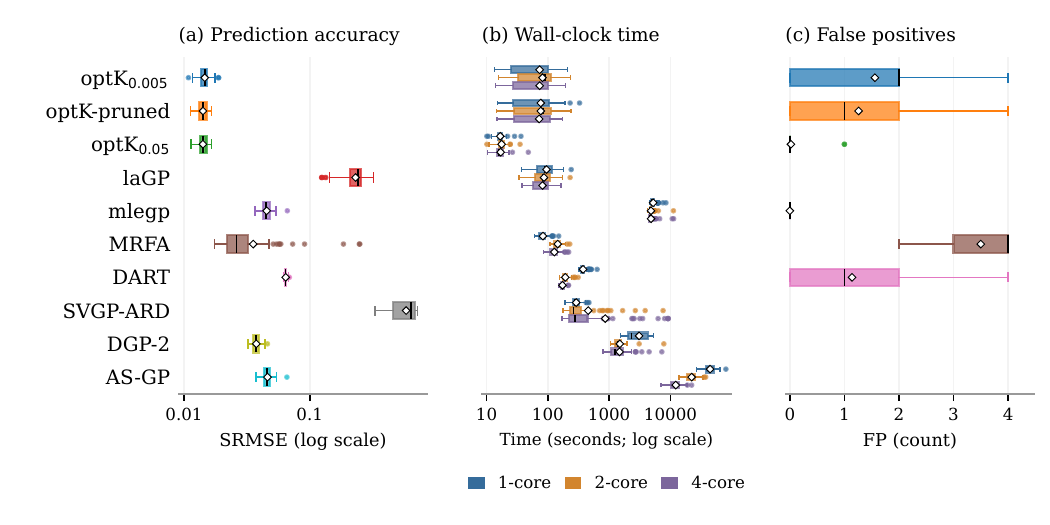}
\caption{Prediction, time and variable selection for the two-variable Michalewicz function with $\bm a=(1,2)$, $n=1000$, $d=6$, $p=2$.}\label{fig:main-michal2-n1000} 
\end{figure}
\endgroup

For the six-variable function, Table~\ref{tab:michal6} gives the $d=60$ results for $n=300,500$.
The corresponding distributions are shown in Figure~\ref{fig:main-michal6-n500} for $n=500$ and Figure~\ref{fig:s-case26} for $n=300$.
At both sample sizes, \texttt{optK-pruned} achieves the lowest mean SRMSE among all methods, and all three optK versions retain all six active inputs in every replication.
At $n=300$, \texttt{optK}\textsubscript{0.05} produces no false positives but has a higher mean SRMSE than the other two optK versions and DART.
At $n=500$, the three optK versions have the three lowest mean SRMSEs and achieve FN$=$FP$=0$.

Additional results for $d=10,20$ and $n=300,500$ are reported in Table~\ref{tab:s-variant} of the Supplementary Material.
All three optK versions have the three lowest mean SRMSEs and FN$=0$ in every configuration.
The version \texttt{optK}\textsubscript{0.005} occasionally selects irrelevant inputs. Terminal pruning reduces mean FP, while \texttt{optK}\textsubscript{0.05} achieves FP$=0$ throughout.

A smaller \texttt{DEL} allows kernels with small weights to remain during the search, potentially aiding prediction but also retaining irrelevant inputs.
A larger \texttt{DEL} generally reduces false positives, although its effect on prediction varies across settings.
In these Michalewicz examples, \texttt{optK-pruned} has lower mean SRMSE and no higher mean FP than the unpruned version with the smaller \texttt{DEL}, with FN remaining zero.

The three optK versions are among the methods with short wall-clock times in the Michalewicz examples, with \texttt{optK}\textsubscript{0.05} being the fastest of the three across all configurations and CPU budgets.
Its larger \texttt{DEL} is associated with lower search iteration counts (Iter), which may partly explain its shorter wall-clock times.

\par\begingroup\spacingset{1}\footnotesize
\setlength{\tabcolsep}{2pt}\renewcommand{\arraystretch}{1}
\def\ptmeasure#1#2{\settowidth{\dimen0}{#2}\ifdim\dimen0>#1\relax\edef#1{\the\dimen0}\fi}
\def\ptMethodWidth{0pt}
\ptmeasure\ptMethodWidth{Method}
\ptmeasure\ptMethodWidth{\texttt{AS-GP}}
\ptmeasure\ptMethodWidth{\texttt{DART}}
\ptmeasure\ptMethodWidth{\texttt{DGP-2}}
\ptmeasure\ptMethodWidth{\texttt{MRFA}}
\ptmeasure\ptMethodWidth{\texttt{SVGP-ARD}}
\ptmeasure\ptMethodWidth{\texttt{laGP}}
\ptmeasure\ptMethodWidth{\texttt{mlegp}}
\ptmeasure\ptMethodWidth{\texttt{optK-pruned}}
\ptmeasure\ptMethodWidth{\texttt{optK}\textsubscript{0.005}}
\ptmeasure\ptMethodWidth{\texttt{optK}\textsubscript{0.05}}
\def\ptSrmseWidth{0pt}
\ptmeasure\ptSrmseWidth{SRMSE}
\ptmeasure\ptSrmseWidth{\mbox{0.0264 (0.00485)}}
\ptmeasure\ptSrmseWidth{\mbox{0.0283 (0.0160)}}
\ptmeasure\ptSrmseWidth{\mbox{0.0338 (0.00498)}}
\ptmeasure\ptSrmseWidth{\mbox{0.0835 (0.00296)}}
\ptmeasure\ptSrmseWidth{\mbox{0.0877 (0.0317)}}
\ptmeasure\ptSrmseWidth{\mbox{0.0946 (0.0308)}}
\ptmeasure\ptSrmseWidth{\mbox{0.0997 (0.0223)}}
\ptmeasure\ptSrmseWidth{\mbox{0.104 (0.00554)}}
\ptmeasure\ptSrmseWidth{\mbox{0.182 (0.0981)}}
\ptmeasure\ptSrmseWidth{\mbox{0.238 (0.0489)}}
\ptmeasure\ptSrmseWidth{\mbox{0.943 (0.0164)}}
\ptmeasure\ptSrmseWidth{\mbox{0.963 (0.0162)}}
\ptmeasure\ptSrmseWidth{\mbox{0.965 (0.0322)}}
\ptmeasure\ptSrmseWidth{\mbox{0.984 (0.0238)}}
\ptmeasure\ptSrmseWidth{\mbox{1.00 (0.0146)}}
\ptmeasure\ptSrmseWidth{\mbox{1.02 (0.0357)}}
\ptmeasure\ptSrmseWidth{\mbox{1.03 (0.0208)}}
\ptmeasure\ptSrmseWidth{\mbox{1.03 (0.0421)}}
\ptmeasure\ptSrmseWidth{\mbox{1.38 (0.0515)}}
\ptmeasure\ptSrmseWidth{\mbox{1.50 (0.132)}}
\def\ptFnWidth{0pt}
\ptmeasure\ptFnWidth{FN}
\ptmeasure\ptFnWidth{\mbox{0 (0)}}
\ptmeasure\ptFnWidth{\mbox{NA}}
\def\ptFpWidth{0pt}
\ptmeasure\ptFpWidth{FP}
\ptmeasure\ptFpWidth{\mbox{0 (0)}}
\ptmeasure\ptFpWidth{\mbox{0.01 (0.10)}}
\ptmeasure\ptFpWidth{\mbox{0.13 (0.44)}}
\ptmeasure\ptFpWidth{\mbox{11.75 (2.65)}}
\ptmeasure\ptFpWidth{\mbox{13.17 (2.60)}}
\ptmeasure\ptFpWidth{\mbox{53.95 (0.33)}}
\ptmeasure\ptFpWidth{\mbox{53.97 (0.17)}}
\ptmeasure\ptFpWidth{\mbox{NA}}
\def\ptIterWidth{0pt}
\ptmeasure\ptIterWidth{Iter}
\ptmeasure\ptIterWidth{\mbox{---}}
\ptmeasure\ptIterWidth{\mbox{15.4 (1.1)}}
\ptmeasure\ptIterWidth{\mbox{15.8 (0.5)}}
\ptmeasure\ptIterWidth{\mbox{16.8 (0.9)}}
\ptmeasure\ptIterWidth{\mbox{17.2 (0.9)}}
\ptmeasure\ptIterWidth{\mbox{17.5 (2.0)}}
\ptmeasure\ptIterWidth{\mbox{17.6 (2.1)}}
\def\ptTimeWidth{0pt}
\ptmeasure\ptTimeWidth{\makecell[r]{1-core\\(Sec)}}
\ptmeasure\ptTimeWidth{\makecell[r]{2-core\\(Sec)}}
\ptmeasure\ptTimeWidth{\makecell[r]{4-core\\(Sec)}}
\ptmeasure\ptTimeWidth{\mbox{104 (16.9)}}
\ptmeasure\ptTimeWidth{\mbox{109 (19.1)}}
\ptmeasure\ptTimeWidth{\mbox{125 (17.1)}}
\ptmeasure\ptTimeWidth{\mbox{132 (16.0)}}
\ptmeasure\ptTimeWidth{\mbox{134 (17.7)}}
\ptmeasure\ptTimeWidth{\mbox{13800 (1950)}}
\ptmeasure\ptTimeWidth{\mbox{144 (44.1)}}
\ptmeasure\ptTimeWidth{\mbox{148 (22.4)}}
\ptmeasure\ptTimeWidth{\mbox{15.9 (4.07)}}
\ptmeasure\ptTimeWidth{\mbox{150 (23.2)}}
\ptmeasure\ptTimeWidth{\mbox{1550 (217)}}
\ptmeasure\ptTimeWidth{\mbox{159 (52.1)}}
\ptmeasure\ptTimeWidth{\mbox{16.2 (2.59)}}
\ptmeasure\ptTimeWidth{\mbox{160 (53.9)}}
\ptmeasure\ptTimeWidth{\mbox{165 (33.0)}}
\ptmeasure\ptTimeWidth{\mbox{167 (61.5)}}
\ptmeasure\ptTimeWidth{\mbox{169 (16.9)}}
\ptmeasure\ptTimeWidth{\mbox{170 (16.4)}}
\ptmeasure\ptTimeWidth{\mbox{18.7 (3.97)}}
\ptmeasure\ptTimeWidth{\mbox{18.8 (5.29)}}
\ptmeasure\ptTimeWidth{\mbox{184 (20.0)}}
\ptmeasure\ptTimeWidth{\mbox{19.0 (2.90)}}
\ptmeasure\ptTimeWidth{\mbox{19.6 (3.14)}}
\ptmeasure\ptTimeWidth{\mbox{20.7 (5.96)}}
\ptmeasure\ptTimeWidth{\mbox{20.9 (4.00)}}
\ptmeasure\ptTimeWidth{\mbox{22.0 (5.20)}}
\ptmeasure\ptTimeWidth{\mbox{223 (343)}}
\ptmeasure\ptTimeWidth{\mbox{235 (532)}}
\ptmeasure\ptTimeWidth{\mbox{258 (20.5)}}
\ptmeasure\ptTimeWidth{\mbox{26.0 (8.81)}}
\ptmeasure\ptTimeWidth{\mbox{26.4 (9.01)}}
\ptmeasure\ptTimeWidth{\mbox{28.4 (9.50)}}
\ptmeasure\ptTimeWidth{\mbox{283 (21.5)}}
\ptmeasure\ptTimeWidth{\mbox{3010 (495)}}
\ptmeasure\ptTimeWidth{\mbox{32.2 (8.44)}}
\ptmeasure\ptTimeWidth{\mbox{35.2 (10.2)}}
\ptmeasure\ptTimeWidth{\mbox{36.5 (9.85)}}
\ptmeasure\ptTimeWidth{\mbox{3780 (589)}}
\ptmeasure\ptTimeWidth{\mbox{38.9 (6.03)}}
\ptmeasure\ptTimeWidth{\mbox{40.5 (8.71)}}
\ptmeasure\ptTimeWidth{\mbox{42.3 (7.05)}}
\ptmeasure\ptTimeWidth{\mbox{44.7 (8.92)}}
\ptmeasure\ptTimeWidth{\mbox{52.0 (7.63)}}
\ptmeasure\ptTimeWidth{\mbox{54.1 (8.43)}}
\ptmeasure\ptTimeWidth{\mbox{55.0 (7.90)}}
\ptmeasure\ptTimeWidth{\mbox{582 (295)}}
\ptmeasure\ptTimeWidth{\mbox{59.0 (9.41)}}
\ptmeasure\ptTimeWidth{\mbox{5970 (801)}}
\ptmeasure\ptTimeWidth{\mbox{627 (254)}}
\ptmeasure\ptTimeWidth{\mbox{671 (180)}}
\ptmeasure\ptTimeWidth{\mbox{6750 (850)}}
\ptmeasure\ptTimeWidth{\mbox{695 (599)}}
\ptmeasure\ptTimeWidth{\mbox{789 (66.5)}}
\ptmeasure\ptTimeWidth{\mbox{793 (71.7)}}
\ptmeasure\ptTimeWidth{\mbox{831 (77.0)}}
\ptmeasure\ptTimeWidth{\mbox{86.4 (31.3)}}
\ptmeasure\ptTimeWidth{\mbox{87.7 (33.4)}}
\ptmeasure\ptTimeWidth{\mbox{893 (215)}}
\ptmeasure\ptTimeWidth{\mbox{93.1 (34.0)}}
\ptmeasure\ptTimeWidth{\mbox{930 (219)}}
\def\ptUpdateWidth{\edef\ptTableWidth{\the\dimexpr\ptMethodWidth+\ptSrmseWidth+\ptFnWidth+\ptFpWidth+\ptIterWidth+\ptTimeWidth+\ptTimeWidth+\ptTimeWidth+14\tabcolsep\relax}}
\ptUpdateWidth\ifdim\ptTableWidth>\linewidth\setlength{\tabcolsep}{1pt}\ptUpdateWidth\fi
\typeout{PERFORMANCE-WIDTH main_michal6: table=\ptTableWidth; time=\ptTimeWidth; linewidth=\the\linewidth}
\ifdim\ptTableWidth>\linewidth\PackageError{performance-tables}{Table exceeds linewidth}{Check the table layout.}\fi
\setlength{\LTcapwidth}{\ptTableWidth}
\Needspace{430pt}
\begin{longtable}{@{}>{\raggedright\arraybackslash}p{\ptMethodWidth}>{\raggedleft\arraybackslash}p{\ptSrmseWidth}>{\raggedleft\arraybackslash}p{\ptFnWidth}>{\raggedleft\arraybackslash}p{\ptFpWidth}>{\raggedleft\arraybackslash}p{\ptIterWidth}>{\raggedleft\arraybackslash}p{\ptTimeWidth}>{\raggedleft\arraybackslash}p{\ptTimeWidth}>{\raggedleft\arraybackslash}p{\ptTimeWidth}@{}}
\caption{Six-variable Michalewicz variant, $\bm a=(1,2,1,2,1,2)$; $d=60$, $p=6$, $m=3481$, $B=100$.}\label{tab:michal6}\\
\toprule
Method & SRMSE & FN & FP & Iter & \makecell[r]{1-core\\(Sec)} & \makecell[r]{2-core\\(Sec)} & \makecell[r]{4-core\\(Sec)} \\
\midrule\endfirsthead
\caption[]{Six-variable Michalewicz variant, $\bm a=(1,2,1,2,1,2)$; $d=60$, $p=6$, $m=3481$, $B=100$. (continued).}\\\toprule
Method & SRMSE & FN & FP & Iter & \makecell[r]{1-core\\(Sec)} & \makecell[r]{2-core\\(Sec)} & \makecell[r]{4-core\\(Sec)} \\
\midrule\endhead
\midrule
\multicolumn{8}{@{}p{\ptTableWidth}@{}}{\footnotesize Values are reported as mean (sample SD). Prediction, FN/FP and Iter use 1-core. An em dash in Iter denotes inapplicability. NA denotes no approved binary selection rule.}\\
\multicolumn{8}{r@{}}{\footnotesize Continued on next page.}\\\endfoot
\bottomrule
\endlastfoot
\multicolumn{8}{@{}l}{\textbf{$n=300$}}\\*
\texttt{optK}\textsubscript{0.005} & \mbox{0.0946 (0.0308)} & \mbox{0 (0)} & \mbox{0.13 (0.44)} & \mbox{17.5 (2.0)} & \mbox{19.0 (2.90)} & \mbox{20.9 (4.00)} & \mbox{18.8 (5.29)} \\*
\texttt{optK-pruned} & \mbox{0.0877 (0.0317)} & \mbox{0 (0)} & \mbox{0.01 (0.10)} & \mbox{17.6 (2.1)} & \mbox{19.6 (3.14)} & \mbox{22.0 (5.20)} & \mbox{20.7 (5.96)} \\*
\texttt{optK}\textsubscript{0.05} & \mbox{0.182 (0.0981)} & \mbox{0 (0)} & \mbox{0 (0)} & \mbox{15.4 (1.1)} & \mbox{16.2 (2.59)} & \mbox{18.7 (3.97)} & \mbox{15.9 (4.07)} \\*
\texttt{laGP} & \mbox{1.03 (0.0208)} & \mbox{NA} & \mbox{NA} & \mbox{---} & \mbox{28.4 (9.50)} & \mbox{26.4 (9.01)} & \mbox{26.0 (8.81)} \\*
\texttt{mlegp} & \mbox{0.963 (0.0162)} & \mbox{0 (0)} & \mbox{53.97 (0.17)} & \mbox{---} & \mbox{184 (20.0)} & \mbox{170 (16.4)} & \mbox{169 (16.9)} \\*
\texttt{MRFA} & \mbox{0.238 (0.0489)} & \mbox{0 (0)} & \mbox{11.75 (2.65)} & \mbox{---} & \mbox{38.9 (6.03)} & \mbox{104 (16.9)} & \mbox{109 (19.1)} \\*
\texttt{DART} & \mbox{0.104 (0.00554)} & \mbox{0 (0)} & \mbox{0 (0)} & \mbox{---} & \mbox{258 (20.5)} & \mbox{132 (16.0)} & \mbox{125 (17.1)} \\*
\texttt{SVGP-ARD} & \mbox{0.984 (0.0238)} & \mbox{NA} & \mbox{NA} & \mbox{---} & \mbox{144 (44.1)} & \mbox{235 (532)} & \mbox{159 (52.1)} \\*
\texttt{DGP-2} & \mbox{1.03 (0.0421)} & \mbox{NA} & \mbox{NA} & \mbox{---} & \mbox{893 (215)} & \mbox{695 (599)} & \mbox{582 (295)} \\*
\texttt{AS-GP} & \mbox{1.50 (0.132)} & \mbox{NA} & \mbox{NA} & \mbox{---} & \mbox{5970 (801)} & \mbox{3010 (495)} & \mbox{1550 (217)} \\
\midrule
\multicolumn{8}{@{}l}{\textbf{$n=500$}}\\*
\texttt{optK}\textsubscript{0.005} & \mbox{0.0338 (0.00498)} & \mbox{0 (0)} & \mbox{0 (0)} & \mbox{16.8 (0.9)} & \mbox{54.1 (8.43)} & \mbox{42.3 (7.05)} & \mbox{35.2 (10.2)} \\*
\texttt{optK-pruned} & \mbox{0.0264 (0.00485)} & \mbox{0 (0)} & \mbox{0 (0)} & \mbox{17.2 (0.9)} & \mbox{55.0 (7.90)} & \mbox{44.7 (8.92)} & \mbox{36.5 (9.85)} \\*
\texttt{optK}\textsubscript{0.05} & \mbox{0.0283 (0.0160)} & \mbox{0 (0)} & \mbox{0 (0)} & \mbox{15.8 (0.5)} & \mbox{52.0 (7.63)} & \mbox{40.5 (8.71)} & \mbox{32.2 (8.44)} \\*
\texttt{laGP} & \mbox{1.00 (0.0146)} & \mbox{NA} & \mbox{NA} & \mbox{---} & \mbox{93.1 (34.0)} & \mbox{87.7 (33.4)} & \mbox{86.4 (31.3)} \\*
\texttt{mlegp} & \mbox{0.943 (0.0164)} & \mbox{0 (0)} & \mbox{53.95 (0.33)} & \mbox{---} & \mbox{831 (77.0)} & \mbox{793 (71.7)} & \mbox{789 (66.5)} \\*
\texttt{MRFA} & \mbox{0.0997 (0.0223)} & \mbox{0 (0)} & \mbox{13.17 (2.60)} & \mbox{---} & \mbox{59.0 (9.41)} & \mbox{150 (23.2)} & \mbox{165 (33.0)} \\*
\texttt{DART} & \mbox{0.0835 (0.00296)} & \mbox{0 (0)} & \mbox{0 (0)} & \mbox{---} & \mbox{283 (21.5)} & \mbox{148 (22.4)} & \mbox{134 (17.7)} \\*
\texttt{SVGP-ARD} & \mbox{0.965 (0.0322)} & \mbox{NA} & \mbox{NA} & \mbox{---} & \mbox{160 (53.9)} & \mbox{223 (343)} & \mbox{167 (61.5)} \\*
\texttt{DGP-2} & \mbox{1.02 (0.0357)} & \mbox{NA} & \mbox{NA} & \mbox{---} & \mbox{930 (219)} & \mbox{671 (180)} & \mbox{627 (254)} \\*
\texttt{AS-GP} & \mbox{1.38 (0.0515)} & \mbox{NA} & \mbox{NA} & \mbox{---} & \mbox{13800 (1950)} & \mbox{6750 (850)} & \mbox{3780 (589)} \\
\end{longtable}\endgroup

\begingroup\spacingset{1}
\begin{figure}[H]\centering
\includegraphics[width=\linewidth]{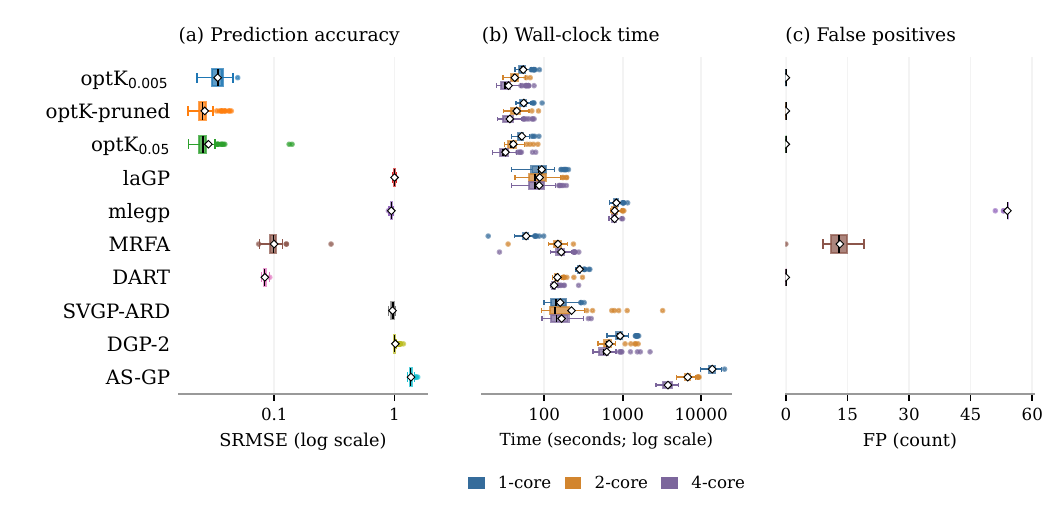}
\caption{Prediction, time and variable selection for the six-variable Michalewicz function with $\bm a=(1,2,1,2,1,2)$, $n=500$, $d=60$, $p=6$. }\label{fig:main-michal6-n500}
\end{figure}
\endgroup

\subsection{Borehole Function}\label{sec:borehole}
The borehole function models the water flow rate through a borehole and is widely used to assess GP regression \citep{morris1993bayesian}:
\[
f(\bm x)=\frac{2\pi T_u(H_u-H_l)}
{\ln(r/r_w)\left(1+\dfrac{2LT_u}{\ln(r/r_w)r_w^2K_w}+\dfrac{T_u}{T_l}\right)}.
\]
The response is the volumetric flow rate, measured in $\mathrm{m}^3/\mathrm{yr}$. All eight inputs appearing in this expression form the true active set. 
Their physical definitions and ranges are given in Table~\ref{table-4}, in the same order as input columns $1$--$8$. 
The function is non-additive, providing a different structure from the preceding example.

We conduct simulations in a manner similar to the previous example.
The study uses $n=200,500$, $d=20,40,60$ and $m=1000$, with $B=100$ replications per configuration.
For each replication, we generate a maximin Latin hypercube training design and a random Latin hypercube test design on $[0,1]^d$.
Training columns are further rescaled using their minima and maxima, while the test design is used without this additional rescaling.
To evaluate the function, the first eight inputs are mapped to the physical ranges and the remaining $d-8$ inputs are irrelevant.
All $d$ inputs are supplied to every method.
Independent noise is added to training responses, and test responses are noiseless values.

Table~\ref{tab:borehole} reports the $d=60$ results for $n=200,500$.
The corresponding distributions are shown in Figure~\ref{fig:main-borehole-n500} for $n=500$ and Figure~\ref{fig:s-case14} for $n=200$.
At $n=200$, the three optK versions have the three lowest mean SRMSEs, with \texttt{optK-pruned} achieving the lowest mean.
At $n=500$, AS-GP has the lowest mean SRMSE, followed by \texttt{optK}\textsubscript{0.005} and \texttt{optK-pruned}.

Additional results for $d=20,40$ and $n=200,500$ are reported in Table~\ref{tab:s-borehole} of the Supplementary Material.
AS-GP has the lowest mean SRMSE in all four configurations.
However, it also has the longest reported mean wall-clock time across all six Borehole configurations and CPU budgets.

For the $d-8$ added irrelevant inputs, terminal pruning reduces mean FP counts relative to \texttt{optK}\textsubscript{0.005}.
The version \texttt{optK}\textsubscript{0.05} has the lowest mean FP counts among the three optK versions, although it does not eliminate FP counts entirely and has a somewhat higher SRMSE than the other two optK versions.

The FN counts of \texttt{optK}\textsubscript{0.005} and \texttt{optK-pruned} are often close to 3 because the two versions of \texttt{optK} consistently retain $r_w,H_u,H_l,L,$ and $K_w$, while frequently omitting $r,T_u,$ and $T_l$.
This pattern is qualitatively consistent with the sensitivity analysis reported by \citet{osti_5355549}.
The larger deletion threshold in \texttt{optK}\textsubscript{0.05} occasionally also excludes $K_w$, resulting in higher mean FN.
This suggests that optK tends to identify variables with sufficiently strong influence under the given experimental design and sample size, rather than necessarily recovering every analytically active variable.
Nevertheless, since all eight inputs enter the Borehole function, the omitted variables are correctly counted as false negatives in our evaluation.

\par\begingroup\spacingset{1}\footnotesize
\setlength{\tabcolsep}{2pt}\renewcommand{\arraystretch}{1}
\def\ptmeasure#1#2{\settowidth{\dimen0}{#2}\ifdim\dimen0>#1\relax\edef#1{\the\dimen0}\fi}
\def\ptMethodWidth{0pt}
\ptmeasure\ptMethodWidth{Method}
\ptmeasure\ptMethodWidth{\texttt{AS-GP}}
\ptmeasure\ptMethodWidth{\texttt{DART}}
\ptmeasure\ptMethodWidth{\texttt{DGP-2}}
\ptmeasure\ptMethodWidth{\texttt{MRFA}}
\ptmeasure\ptMethodWidth{\texttt{SVGP-ARD}}
\ptmeasure\ptMethodWidth{\texttt{laGP}}
\ptmeasure\ptMethodWidth{\texttt{mlegp}}
\ptmeasure\ptMethodWidth{\texttt{optK-pruned}}
\ptmeasure\ptMethodWidth{\texttt{optK}\textsubscript{0.005}}
\ptmeasure\ptMethodWidth{\texttt{optK}\textsubscript{0.05}}
\def\ptSrmseWidth{0pt}
\ptmeasure\ptSrmseWidth{SRMSE}
\ptmeasure\ptSrmseWidth{\mbox{0.0177 (0.00547)}}
\ptmeasure\ptSrmseWidth{\mbox{0.0325 (0.00806)}}
\ptmeasure\ptSrmseWidth{\mbox{0.0336 (0.00902)}}
\ptmeasure\ptSrmseWidth{\mbox{0.0453 (0.0268)}}
\ptmeasure\ptSrmseWidth{\mbox{0.0560 (0.00335)}}
\ptmeasure\ptSrmseWidth{\mbox{0.0561 (0.0413)}}
\ptmeasure\ptSrmseWidth{\mbox{0.0675 (0.0190)}}
\ptmeasure\ptSrmseWidth{\mbox{0.0712 (0.0178)}}
\ptmeasure\ptSrmseWidth{\mbox{0.0784 (0.0946)}}
\ptmeasure\ptSrmseWidth{\mbox{0.0820 (0.0326)}}
\ptmeasure\ptSrmseWidth{\mbox{0.0871 (0.00749)}}
\ptmeasure\ptSrmseWidth{\mbox{0.0943 (0.00671)}}
\ptmeasure\ptSrmseWidth{\mbox{0.0949 (0.0623)}}
\ptmeasure\ptSrmseWidth{\mbox{0.112 (0.0304)}}
\ptmeasure\ptSrmseWidth{\mbox{0.145 (0.0126)}}
\ptmeasure\ptSrmseWidth{\mbox{0.156 (0.0243)}}
\ptmeasure\ptSrmseWidth{\mbox{0.197 (0.0504)}}
\ptmeasure\ptSrmseWidth{\mbox{0.307 (0.333)}}
\ptmeasure\ptSrmseWidth{\mbox{0.936 (0.205)}}
\ptmeasure\ptSrmseWidth{\mbox{0.988 (0.0886)}}
\def\ptFnWidth{0pt}
\ptmeasure\ptFnWidth{FN}
\ptmeasure\ptFnWidth{\mbox{0.05 (0.41)}}
\ptmeasure\ptFnWidth{\mbox{1.08 (1.52)}}
\ptmeasure\ptFnWidth{\mbox{2.32 (0.68)}}
\ptmeasure\ptFnWidth{\mbox{2.34 (0.73)}}
\ptmeasure\ptFnWidth{\mbox{2.79 (0.46)}}
\ptmeasure\ptFnWidth{\mbox{2.91 (0.29)}}
\ptmeasure\ptFnWidth{\mbox{2.95 (0.22)}}
\ptmeasure\ptFnWidth{\mbox{2.96 (0.20)}}
\ptmeasure\ptFnWidth{\mbox{3.00 (0.00)}}
\ptmeasure\ptFnWidth{\mbox{3.07 (0.26)}}
\ptmeasure\ptFnWidth{\mbox{3.12 (0.36)}}
\ptmeasure\ptFnWidth{\mbox{NA}}
\def\ptFpWidth{0pt}
\ptmeasure\ptFpWidth{FP}
\ptmeasure\ptFpWidth{\mbox{0.01 (0.10)}}
\ptmeasure\ptFpWidth{\mbox{0.08 (0.31)}}
\ptmeasure\ptFpWidth{\mbox{0.14 (0.40)}}
\ptmeasure\ptFpWidth{\mbox{0.62 (0.71)}}
\ptmeasure\ptFpWidth{\mbox{0.80 (0.89)}}
\ptmeasure\ptFpWidth{\mbox{1.55 (1.34)}}
\ptmeasure\ptFpWidth{\mbox{1.57 (1.20)}}
\ptmeasure\ptFpWidth{\mbox{10.28 (2.06)}}
\ptmeasure\ptFpWidth{\mbox{10.40 (3.08)}}
\ptmeasure\ptFpWidth{\mbox{35.45 (21.90)}}
\ptmeasure\ptFpWidth{\mbox{4.05 (2.65)}}
\ptmeasure\ptFpWidth{\mbox{51.40 (5.00)}}
\ptmeasure\ptFpWidth{\mbox{NA}}
\def\ptIterWidth{0pt}
\ptmeasure\ptIterWidth{Iter}
\ptmeasure\ptIterWidth{\mbox{---}}
\ptmeasure\ptIterWidth{\mbox{18.4 (3.1)}}
\ptmeasure\ptIterWidth{\mbox{19.3 (3.4)}}
\ptmeasure\ptIterWidth{\mbox{52.1 (6.5)}}
\ptmeasure\ptIterWidth{\mbox{52.9 (5.5)}}
\ptmeasure\ptIterWidth{\mbox{56.8 (7.5)}}
\ptmeasure\ptIterWidth{\mbox{57.4 (7.8)}}
\def\ptTimeWidth{0pt}
\ptmeasure\ptTimeWidth{\makecell[r]{1-core\\(Sec)}}
\ptmeasure\ptTimeWidth{\makecell[r]{2-core\\(Sec)}}
\ptmeasure\ptTimeWidth{\makecell[r]{4-core\\(Sec)}}
\ptmeasure\ptTimeWidth{\mbox{1010 (210)}}
\ptmeasure\ptTimeWidth{\mbox{12.9 (4.33)}}
\ptmeasure\ptTimeWidth{\mbox{120 (11.0)}}
\ptmeasure\ptTimeWidth{\mbox{13.5 (4.69)}}
\ptmeasure\ptTimeWidth{\mbox{1410 (646)}}
\ptmeasure\ptTimeWidth{\mbox{149 (71.5)}}
\ptmeasure\ptTimeWidth{\mbox{154 (74.2)}}
\ptmeasure\ptTimeWidth{\mbox{1620 (1050)}}
\ptmeasure\ptTimeWidth{\mbox{168 (64.0)}}
\ptmeasure\ptTimeWidth{\mbox{16800 (3050)}}
\ptmeasure\ptTimeWidth{\mbox{183 (24.6)}}
\ptmeasure\ptTimeWidth{\mbox{192 (116)}}
\ptmeasure\ptTimeWidth{\mbox{196 (109)}}
\ptmeasure\ptTimeWidth{\mbox{1960 (433)}}
\ptmeasure\ptTimeWidth{\mbox{20.2 (5.48)}}
\ptmeasure\ptTimeWidth{\mbox{205 (148)}}
\ptmeasure\ptTimeWidth{\mbox{216 (130)}}
\ptmeasure\ptTimeWidth{\mbox{234 (77.4)}}
\ptmeasure\ptTimeWidth{\mbox{244 (331)}}
\ptmeasure\ptTimeWidth{\mbox{25.2 (6.86)}}
\ptmeasure\ptTimeWidth{\mbox{2720 (1190)}}
\ptmeasure\ptTimeWidth{\mbox{30.8 (7.57)}}
\ptmeasure\ptTimeWidth{\mbox{325 (450)}}
\ptmeasure\ptTimeWidth{\mbox{327 (588)}}
\ptmeasure\ptTimeWidth{\mbox{34.5 (25.9)}}
\ptmeasure\ptTimeWidth{\mbox{36.3 (32.6)}}
\ptmeasure\ptTimeWidth{\mbox{36.9 (10.2)}}
\ptmeasure\ptTimeWidth{\mbox{37.5 (28.3)}}
\ptmeasure\ptTimeWidth{\mbox{3720 (795)}}
\ptmeasure\ptTimeWidth{\mbox{38.9 (10.2)}}
\ptmeasure\ptTimeWidth{\mbox{4.32 (0.870)}}
\ptmeasure\ptTimeWidth{\mbox{4.46 (1.13)}}
\ptmeasure\ptTimeWidth{\mbox{4.80 (1.65)}}
\ptmeasure\ptTimeWidth{\mbox{413 (139)}}
\ptmeasure\ptTimeWidth{\mbox{419 (137)}}
\ptmeasure\ptTimeWidth{\mbox{46.3 (7.27)}}
\ptmeasure\ptTimeWidth{\mbox{47.9 (3.46)}}
\ptmeasure\ptTimeWidth{\mbox{48.3 (8.61)}}
\ptmeasure\ptTimeWidth{\mbox{50.3 (8.94)}}
\ptmeasure\ptTimeWidth{\mbox{5010 (1580)}}
\ptmeasure\ptTimeWidth{\mbox{52.0 (9.41)}}
\ptmeasure\ptTimeWidth{\mbox{568 (1010)}}
\ptmeasure\ptTimeWidth{\mbox{57.2 (6.21)}}
\ptmeasure\ptTimeWidth{\mbox{575 (432)}}
\ptmeasure\ptTimeWidth{\mbox{598 (472)}}
\ptmeasure\ptTimeWidth{\mbox{628 (262)}}
\ptmeasure\ptTimeWidth{\mbox{63.9 (12.6)}}
\ptmeasure\ptTimeWidth{\mbox{650 (241)}}
\ptmeasure\ptTimeWidth{\mbox{7.15 (1.97)}}
\ptmeasure\ptTimeWidth{\mbox{727 (68.5)}}
\ptmeasure\ptTimeWidth{\mbox{739 (70.5)}}
\ptmeasure\ptTimeWidth{\mbox{74.0 (18.1)}}
\ptmeasure\ptTimeWidth{\mbox{774 (65.7)}}
\ptmeasure\ptTimeWidth{\mbox{83.1 (23.0)}}
\ptmeasure\ptTimeWidth{\mbox{86.6 (9.29)}}
\ptmeasure\ptTimeWidth{\mbox{898 (326)}}
\ptmeasure\ptTimeWidth{\mbox{920 (279)}}
\ptmeasure\ptTimeWidth{\mbox{922 (292)}}
\ptmeasure\ptTimeWidth{\mbox{9240 (2270)}}
\ptmeasure\ptTimeWidth{\mbox{94.6 (15.0)}}
\def\ptUpdateWidth{\edef\ptTableWidth{\the\dimexpr\ptMethodWidth+\ptSrmseWidth+\ptFnWidth+\ptFpWidth+\ptIterWidth+\ptTimeWidth+\ptTimeWidth+\ptTimeWidth+14\tabcolsep\relax}}
\ptUpdateWidth\ifdim\ptTableWidth>\linewidth\setlength{\tabcolsep}{1pt}\ptUpdateWidth\fi
\typeout{PERFORMANCE-WIDTH main_borehole: table=\ptTableWidth; time=\ptTimeWidth; linewidth=\the\linewidth}
\ifdim\ptTableWidth>\linewidth\PackageError{performance-tables}{Table exceeds linewidth}{Check the table layout.}\fi
\setlength{\LTcapwidth}{\ptTableWidth}
\Needspace{430pt}
\begin{longtable}{@{}>{\raggedright\arraybackslash}p{\ptMethodWidth}>{\raggedleft\arraybackslash}p{\ptSrmseWidth}>{\raggedleft\arraybackslash}p{\ptFnWidth}>{\raggedleft\arraybackslash}p{\ptFpWidth}>{\raggedleft\arraybackslash}p{\ptIterWidth}>{\raggedleft\arraybackslash}p{\ptTimeWidth}>{\raggedleft\arraybackslash}p{\ptTimeWidth}>{\raggedleft\arraybackslash}p{\ptTimeWidth}@{}}
\caption{Borehole function; $d=60$, $p=8$, $m=1000$, $B=100$.}\label{tab:borehole}\\
\toprule
Method & SRMSE & FN & FP & Iter & \makecell[r]{1-core\\(Sec)} & \makecell[r]{2-core\\(Sec)} & \makecell[r]{4-core\\(Sec)} \\
\midrule\endfirsthead
\caption[]{Borehole function; $d=60$, $p=8$, $m=1000$, $B=100$. (continued).}\\\toprule
Method & SRMSE & FN & FP & Iter & \makecell[r]{1-core\\(Sec)} & \makecell[r]{2-core\\(Sec)} & \makecell[r]{4-core\\(Sec)} \\
\midrule\endhead
\midrule
\multicolumn{8}{@{}p{\ptTableWidth}@{}}{\footnotesize Values are reported as mean (sample SD). Prediction, FN/FP and Iter use 1-core. An em dash in Iter denotes inapplicability. NA denotes no approved binary selection rule.}\\
\multicolumn{8}{r@{}}{\footnotesize Continued on next page.}\\\endfoot
\bottomrule
\endlastfoot
\multicolumn{8}{@{}l}{\textbf{$n=200$}}\\*
\texttt{optK}\textsubscript{0.005} & \mbox{0.0712 (0.0178)} & \mbox{2.79 (0.46)} & \mbox{4.05 (2.65)} & \mbox{52.9 (5.5)} & \mbox{216 (130)} & \mbox{196 (109)} & \mbox{154 (74.2)} \\*
\texttt{optK-pruned} & \mbox{0.0675 (0.0190)} & \mbox{2.91 (0.29)} & \mbox{1.55 (1.34)} & \mbox{52.1 (6.5)} & \mbox{205 (148)} & \mbox{192 (116)} & \mbox{149 (71.5)} \\*
\texttt{optK}\textsubscript{0.05} & \mbox{0.0820 (0.0326)} & \mbox{3.12 (0.36)} & \mbox{0.14 (0.40)} & \mbox{19.3 (3.4)} & \mbox{7.15 (1.97)} & \mbox{13.5 (4.69)} & \mbox{12.9 (4.33)} \\*
\texttt{laGP} & \mbox{0.145 (0.0126)} & \mbox{NA} & \mbox{NA} & \mbox{---} & \mbox{4.80 (1.65)} & \mbox{4.32 (0.870)} & \mbox{4.46 (1.13)} \\*
\texttt{mlegp} & \mbox{0.197 (0.0504)} & \mbox{0.05 (0.41)} & \mbox{51.40 (5.00)} & \mbox{---} & \mbox{52.0 (9.41)} & \mbox{47.9 (3.46)} & \mbox{48.3 (8.61)} \\*
\texttt{MRFA} & \mbox{0.156 (0.0243)} & \mbox{2.32 (0.68)} & \mbox{10.28 (2.06)} & \mbox{---} & \mbox{20.2 (5.48)} & \mbox{46.3 (7.27)} & \mbox{50.3 (8.94)} \\*
\texttt{DART} & \mbox{0.0871 (0.00749)} & \mbox{3.00 (0.00)} & \mbox{0.01 (0.10)} & \mbox{---} & \mbox{120 (11.0)} & \mbox{63.9 (12.6)} & \mbox{57.2 (6.21)} \\*
\texttt{SVGP-ARD} & \mbox{0.936 (0.205)} & \mbox{NA} & \mbox{NA} & \mbox{---} & \mbox{168 (64.0)} & \mbox{244 (331)} & \mbox{327 (588)} \\*
\texttt{DGP-2} & \mbox{0.988 (0.0886)} & \mbox{NA} & \mbox{NA} & \mbox{---} & \mbox{920 (279)} & \mbox{598 (472)} & \mbox{575 (432)} \\*
\texttt{AS-GP} & \mbox{0.112 (0.0304)} & \mbox{NA} & \mbox{NA} & \mbox{---} & \mbox{3720 (795)} & \mbox{1960 (433)} & \mbox{1010 (210)} \\
\midrule
\multicolumn{8}{@{}l}{\textbf{$n=500$}}\\*
\texttt{optK}\textsubscript{0.005} & \mbox{0.0325 (0.00806)} & \mbox{2.95 (0.22)} & \mbox{1.57 (1.20)} & \mbox{57.4 (7.8)} & \mbox{922 (292)} & \mbox{650 (241)} & \mbox{419 (137)} \\*
\texttt{optK-pruned} & \mbox{0.0336 (0.00902)} & \mbox{2.95 (0.22)} & \mbox{0.62 (0.71)} & \mbox{56.8 (7.5)} & \mbox{898 (326)} & \mbox{628 (262)} & \mbox{413 (139)} \\*
\texttt{optK}\textsubscript{0.05} & \mbox{0.0453 (0.0268)} & \mbox{3.07 (0.26)} & \mbox{0.08 (0.31)} & \mbox{18.4 (3.1)} & \mbox{38.9 (10.2)} & \mbox{30.8 (7.57)} & \mbox{25.2 (6.86)} \\*
\texttt{laGP} & \mbox{0.0943 (0.00671)} & \mbox{NA} & \mbox{NA} & \mbox{---} & \mbox{37.5 (28.3)} & \mbox{36.3 (32.6)} & \mbox{34.5 (25.9)} \\*
\texttt{mlegp} & \mbox{0.0949 (0.0623)} & \mbox{1.08 (1.52)} & \mbox{35.45 (21.90)} & \mbox{---} & \mbox{774 (65.7)} & \mbox{739 (70.5)} & \mbox{727 (68.5)} \\*
\texttt{MRFA} & \mbox{0.0561 (0.0413)} & \mbox{2.34 (0.73)} & \mbox{10.40 (3.08)} & \mbox{---} & \mbox{36.9 (10.2)} & \mbox{74.0 (18.1)} & \mbox{83.1 (23.0)} \\*
\texttt{DART} & \mbox{0.0560 (0.00335)} & \mbox{2.96 (0.20)} & \mbox{0.80 (0.89)} & \mbox{---} & \mbox{183 (24.6)} & \mbox{94.6 (15.0)} & \mbox{86.6 (9.29)} \\*
\texttt{SVGP-ARD} & \mbox{0.307 (0.333)} & \mbox{NA} & \mbox{NA} & \mbox{---} & \mbox{234 (77.4)} & \mbox{325 (450)} & \mbox{568 (1010)} \\*
\texttt{DGP-2} & \mbox{0.0784 (0.0946)} & \mbox{NA} & \mbox{NA} & \mbox{---} & \mbox{2720 (1190)} & \mbox{1410 (646)} & \mbox{1620 (1050)} \\*
\texttt{AS-GP} & \mbox{0.0177 (0.00547)} & \mbox{NA} & \mbox{NA} & \mbox{---} & \mbox{16800 (3050)} & \mbox{9240 (2270)} & \mbox{5010 (1580)} \\
\end{longtable}\endgroup

\begingroup\spacingset{1}
\begin{figure}[H]\centering
\includegraphics[width=\linewidth]{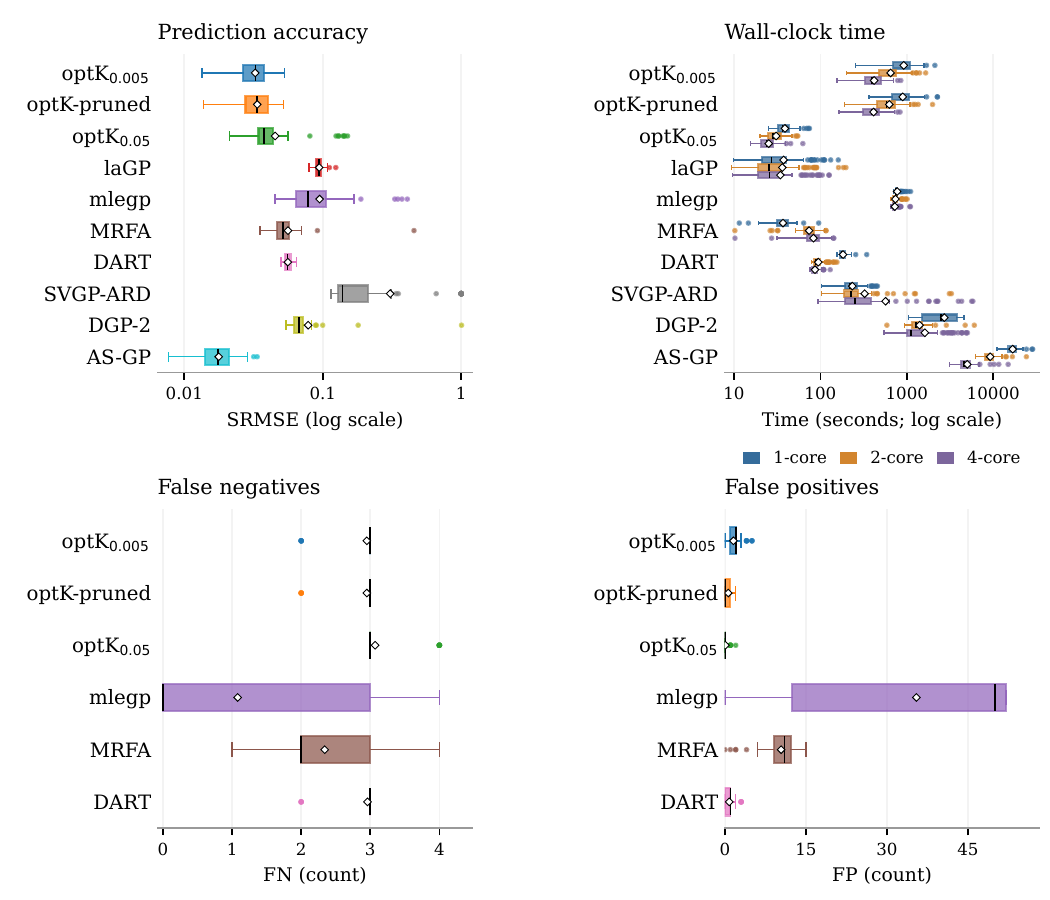}
\caption{Prediction, time and variable selection for the Borehole function; $n=500$, $d=60$, $p=8$, $m=1000$, $B=100$. }\label{fig:main-borehole-n500}
\end{figure}
\endgroup

\subsection{Predicting Satellite Drag}\label{sec:satellite}
Accurate estimation of satellite drag coefficients in low Earth orbit is important for precise positioning and collision avoidance. 
The computational cost of direct simulation motivates surrogate models for drag prediction \citep{mehta2014modeling}. 
We use the GRACE (Gravity Recovery and Climate Experiment) data described in Section~2.3 of \citet{gramacy2020surrogates}, generated with a Test Particle Monte Carlo (TPMC) simulator developed at Los Alamos National Laboratory (LANL). 
We model the drag coefficient $C_D$ separately for three pure species: helium (He), molecular nitrogen ($\mathrm{N}_2$) and molecular oxygen ($\mathrm{O}_2$). 
The seven physical inputs describe relative velocity, surface and atmospheric temperatures, yaw and pitch angles, and normal energy and tangential momentum accommodation coefficients. 
The original simulation study restricts the ranges of yaw and pitch angles.

For each response and replication, $n=200$ or $500$ training observations are sampled without replacement from the original 1000-observation pool, and predictions are evaluated on the existing separate 100-observation test set. 
Seven independent Uniform$[0,1]$ inputs are appended as irrelevant dimensions to the seven physical inputs, giving $d=14$, to assess prediction and variable selection in the presence of irrelevant inputs. 
The inputs are rescaled using training data only, and the stored simulator responses are used without added noise.
Data provenance and preprocessing details are provided in Section~\ref{ssec:s3-implementation}.

For variable-selection evaluation, we treat the seven physical inputs as the reference active set and the seven added inputs as inactive. 
FN and FP count omitted physical inputs and selected added inputs.

Table~\ref{tab:satellite} reports the $n=500$ results for the three gases, with the corresponding distributions shown in Figures~\ref{fig:main-satellite-he}--\ref{fig:main-satellite-o2}.
AS-GP achieves the lowest mean SRMSE for He, while \texttt{optK}\textsubscript{0.005} and \texttt{optK-pruned} achieve the lowest means for $\mathrm{N}_2$ and $\mathrm{O}_2$, respectively.
Additional results for $n=200$ are reported in Table~\ref{tab:s-satellite} of the Supplementary Material.
For all three gases, \texttt{optK-pruned} achieves the lowest mean SRMSE, followed by \texttt{optK}\textsubscript{0.005}.
Across the six configurations, terminal pruning lowers mean SRMSE relative to \texttt{optK}\textsubscript{0.005}, except for $\mathrm{N}_2$ at $n=500$.
The version \texttt{optK}\textsubscript{0.05} has higher mean SRMSE than the other two optK versions in every configuration.

For the seven added irrelevant inputs, terminal pruning reduces mean FP relative to \texttt{optK}\textsubscript{0.005} in all six configurations.
The version \texttt{optK}\textsubscript{0.05} has the lowest mean FP among the three optK versions, although it does not eliminate false positives.

In terms of computation, \texttt{optK}\textsubscript{0.05} has the shortest reported wall-clock time among the three optK versions across all configurations and CPU budgets, while AS-GP has the longest wall-clock time.

Terminal pruning and the larger \texttt{DEL} increase FN relative to \texttt{optK}\textsubscript{0.005}.
These omissions may reflect weak effects of some physical inputs for a particular gas, or limitations of variable selection at the available sample size.
\par\begingroup\spacingset{1}\footnotesize
\setlength{\tabcolsep}{2pt}\renewcommand{\arraystretch}{1}
\def\ptmeasure#1#2{\settowidth{\dimen0}{#2}\ifdim\dimen0>#1\relax\edef#1{\the\dimen0}\fi}
\def\ptMethodWidth{0pt}
\ptmeasure\ptMethodWidth{Method}
\ptmeasure\ptMethodWidth{\texttt{AS-GP}}
\ptmeasure\ptMethodWidth{\texttt{DART}}
\ptmeasure\ptMethodWidth{\texttt{DGP-2}}
\ptmeasure\ptMethodWidth{\texttt{MRFA}}
\ptmeasure\ptMethodWidth{\texttt{SVGP-ARD}}
\ptmeasure\ptMethodWidth{\texttt{laGP}}
\ptmeasure\ptMethodWidth{\texttt{mlegp}}
\ptmeasure\ptMethodWidth{\texttt{optK-pruned}}
\ptmeasure\ptMethodWidth{\texttt{optK}\textsubscript{0.005}}
\ptmeasure\ptMethodWidth{\texttt{optK}\textsubscript{0.05}}
\def\ptSrmseWidth{0pt}
\ptmeasure\ptSrmseWidth{SRMSE}
\ptmeasure\ptSrmseWidth{\mbox{0.0443 (0.00328)}}
\ptmeasure\ptSrmseWidth{\mbox{0.0512 (0.00880)}}
\ptmeasure\ptSrmseWidth{\mbox{0.0521 (0.00515)}}
\ptmeasure\ptSrmseWidth{\mbox{0.0549 (0.00463)}}
\ptmeasure\ptSrmseWidth{\mbox{0.0566 (0.00645)}}
\ptmeasure\ptSrmseWidth{\mbox{0.0575 (0.00642)}}
\ptmeasure\ptSrmseWidth{\mbox{0.0575 (0.00809)}}
\ptmeasure\ptSrmseWidth{\mbox{0.0577 (0.00410)}}
\ptmeasure\ptSrmseWidth{\mbox{0.0592 (0.00462)}}
\ptmeasure\ptSrmseWidth{\mbox{0.0667 (0.00668)}}
\ptmeasure\ptSrmseWidth{\mbox{0.0688 (0.00536)}}
\ptmeasure\ptSrmseWidth{\mbox{0.0702 (0.0262)}}
\ptmeasure\ptSrmseWidth{\mbox{0.0715 (0.0199)}}
\ptmeasure\ptSrmseWidth{\mbox{0.0738 (0.00570)}}
\ptmeasure\ptSrmseWidth{\mbox{0.0750 (0.00666)}}
\ptmeasure\ptSrmseWidth{\mbox{0.0776 (0.00560)}}
\ptmeasure\ptSrmseWidth{\mbox{0.0779 (0.00643)}}
\ptmeasure\ptSrmseWidth{\mbox{0.0802 (0.00645)}}
\ptmeasure\ptSrmseWidth{\mbox{0.0850 (0.00833)}}
\ptmeasure\ptSrmseWidth{\mbox{0.0919 (0.00715)}}
\ptmeasure\ptSrmseWidth{\mbox{0.0943 (0.0396)}}
\ptmeasure\ptSrmseWidth{\mbox{0.0973 (0.00749)}}
\ptmeasure\ptSrmseWidth{\mbox{0.103 (0.00761)}}
\ptmeasure\ptSrmseWidth{\mbox{0.107 (0.00975)}}
\ptmeasure\ptSrmseWidth{\mbox{0.150 (0.00923)}}
\ptmeasure\ptSrmseWidth{\mbox{0.151 (0.00957)}}
\ptmeasure\ptSrmseWidth{\mbox{0.163 (0.0943)}}
\ptmeasure\ptSrmseWidth{\mbox{0.166 (0.0108)}}
\ptmeasure\ptSrmseWidth{\mbox{0.167 (0.104)}}
\ptmeasure\ptSrmseWidth{\mbox{0.169 (0.0783)}}
\def\ptFnWidth{0pt}
\ptmeasure\ptFnWidth{FN}
\ptmeasure\ptFnWidth{\mbox{0.49 (0.52)}}
\ptmeasure\ptFnWidth{\mbox{0.50 (0.59)}}
\ptmeasure\ptFnWidth{\mbox{0.53 (0.63)}}
\ptmeasure\ptFnWidth{\mbox{0.59 (0.49)}}
\ptmeasure\ptFnWidth{\mbox{0.61 (0.49)}}
\ptmeasure\ptFnWidth{\mbox{0.63 (0.49)}}
\ptmeasure\ptFnWidth{\mbox{0.69 (0.46)}}
\ptmeasure\ptFnWidth{\mbox{0.74 (0.44)}}
\ptmeasure\ptFnWidth{\mbox{0.85 (0.36)}}
\ptmeasure\ptFnWidth{\mbox{0.86 (0.35)}}
\ptmeasure\ptFnWidth{\mbox{0.94 (0.24)}}
\ptmeasure\ptFnWidth{\mbox{0.95 (0.22)}}
\ptmeasure\ptFnWidth{\mbox{0.97 (0.22)}}
\ptmeasure\ptFnWidth{\mbox{0.98 (0.14)}}
\ptmeasure\ptFnWidth{\mbox{1.02 (0.20)}}
\ptmeasure\ptFnWidth{\mbox{1.21 (0.43)}}
\ptmeasure\ptFnWidth{\mbox{NA}}
\def\ptFpWidth{0pt}
\ptmeasure\ptFpWidth{FP}
\ptmeasure\ptFpWidth{\mbox{0.10 (0.30)}}
\ptmeasure\ptFpWidth{\mbox{0.15 (0.36)}}
\ptmeasure\ptFpWidth{\mbox{0.15 (0.39)}}
\ptmeasure\ptFpWidth{\mbox{0.16 (0.37)}}
\ptmeasure\ptFpWidth{\mbox{0.42 (0.54)}}
\ptmeasure\ptFpWidth{\mbox{0.52 (0.64)}}
\ptmeasure\ptFpWidth{\mbox{0.59 (0.68)}}
\ptmeasure\ptFpWidth{\mbox{0.61 (0.62)}}
\ptmeasure\ptFpWidth{\mbox{1.67 (1.18)}}
\ptmeasure\ptFpWidth{\mbox{1.88 (1.34)}}
\ptmeasure\ptFpWidth{\mbox{1.92 (1.21)}}
\ptmeasure\ptFpWidth{\mbox{2.34 (1.07)}}
\ptmeasure\ptFpWidth{\mbox{2.39 (1.32)}}
\ptmeasure\ptFpWidth{\mbox{2.43 (1.27)}}
\ptmeasure\ptFpWidth{\mbox{3.87 (2.25)}}
\ptmeasure\ptFpWidth{\mbox{3.92 (1.76)}}
\ptmeasure\ptFpWidth{\mbox{4.17 (1.83)}}
\ptmeasure\ptFpWidth{\mbox{NA}}
\def\ptIterWidth{0pt}
\ptmeasure\ptIterWidth{Iter}
\ptmeasure\ptIterWidth{\mbox{---}}
\ptmeasure\ptIterWidth{\mbox{14.2 (2.3)}}
\ptmeasure\ptIterWidth{\mbox{15.0 (3.0)}}
\ptmeasure\ptIterWidth{\mbox{17.4 (2.7)}}
\ptmeasure\ptIterWidth{\mbox{34.6 (5.3)}}
\ptmeasure\ptIterWidth{\mbox{35.5 (4.9)}}
\ptmeasure\ptIterWidth{\mbox{36.8 (3.9)}}
\ptmeasure\ptIterWidth{\mbox{37.2 (3.4)}}
\ptmeasure\ptIterWidth{\mbox{39.2 (3.5)}}
\ptmeasure\ptIterWidth{\mbox{39.3 (3.9)}}
\def\ptTimeWidth{0pt}
\ptmeasure\ptTimeWidth{\makecell[r]{1-core\\(Sec)}}
\ptmeasure\ptTimeWidth{\makecell[r]{2-core\\(Sec)}}
\ptmeasure\ptTimeWidth{\makecell[r]{4-core\\(Sec)}}
\ptmeasure\ptTimeWidth{\mbox{1160 (2460)}}
\ptmeasure\ptTimeWidth{\mbox{128 (66.7)}}
\ptmeasure\ptTimeWidth{\mbox{12900 (2270)}}
\ptmeasure\ptTimeWidth{\mbox{13000 (2480)}}
\ptmeasure\ptTimeWidth{\mbox{131 (20.5)}}
\ptmeasure\ptTimeWidth{\mbox{131 (20.6)}}
\ptmeasure\ptTimeWidth{\mbox{131 (62.7)}}
\ptmeasure\ptTimeWidth{\mbox{13200 (2500)}}
\ptmeasure\ptTimeWidth{\mbox{133 (18.4)}}
\ptmeasure\ptTimeWidth{\mbox{148 (90.6)}}
\ptmeasure\ptTimeWidth{\mbox{1540 (393)}}
\ptmeasure\ptTimeWidth{\mbox{16.6 (5.23)}}
\ptmeasure\ptTimeWidth{\mbox{1600 (354)}}
\ptmeasure\ptTimeWidth{\mbox{1640 (337)}}
\ptmeasure\ptTimeWidth{\mbox{17.1 (5.28)}}
\ptmeasure\ptTimeWidth{\mbox{170 (77.6)}}
\ptmeasure\ptTimeWidth{\mbox{172 (288)}}
\ptmeasure\ptTimeWidth{\mbox{177 (81.4)}}
\ptmeasure\ptTimeWidth{\mbox{183 (84.4)}}
\ptmeasure\ptTimeWidth{\mbox{187 (151)}}
\ptmeasure\ptTimeWidth{\mbox{19.1 (5.94)}}
\ptmeasure\ptTimeWidth{\mbox{19.7 (6.56)}}
\ptmeasure\ptTimeWidth{\mbox{20.0 (5.87)}}
\ptmeasure\ptTimeWidth{\mbox{202 (94.4)}}
\ptmeasure\ptTimeWidth{\mbox{204 (130)}}
\ptmeasure\ptTimeWidth{\mbox{22.1 (4.42)}}
\ptmeasure\ptTimeWidth{\mbox{22.5 (6.51)}}
\ptmeasure\ptTimeWidth{\mbox{220 (128)}}
\ptmeasure\ptTimeWidth{\mbox{229 (160)}}
\ptmeasure\ptTimeWidth{\mbox{23.0 (6.97)}}
\ptmeasure\ptTimeWidth{\mbox{23.2 (8.33)}}
\ptmeasure\ptTimeWidth{\mbox{23.8 (8.14)}}
\ptmeasure\ptTimeWidth{\mbox{24.2 (8.70)}}
\ptmeasure\ptTimeWidth{\mbox{24.3 (8.79)}}
\ptmeasure\ptTimeWidth{\mbox{24.6 (9.39)}}
\ptmeasure\ptTimeWidth{\mbox{242 (47.0)}}
\ptmeasure\ptTimeWidth{\mbox{244 (111)}}
\ptmeasure\ptTimeWidth{\mbox{244 (45.4)}}
\ptmeasure\ptTimeWidth{\mbox{250 (130)}}
\ptmeasure\ptTimeWidth{\mbox{250 (64.9)}}
\ptmeasure\ptTimeWidth{\mbox{251 (126)}}
\ptmeasure\ptTimeWidth{\mbox{26.2 (9.61)}}
\ptmeasure\ptTimeWidth{\mbox{266 (133)}}
\ptmeasure\ptTimeWidth{\mbox{30.1 (22.2)}}
\ptmeasure\ptTimeWidth{\mbox{30.1 (9.20)}}
\ptmeasure\ptTimeWidth{\mbox{30.4 (25.4)}}
\ptmeasure\ptTimeWidth{\mbox{315 (140)}}
\ptmeasure\ptTimeWidth{\mbox{32.7 (23.7)}}
\ptmeasure\ptTimeWidth{\mbox{339 (142)}}
\ptmeasure\ptTimeWidth{\mbox{36.1 (5.98)}}
\ptmeasure\ptTimeWidth{\mbox{36.3 (7.00)}}
\ptmeasure\ptTimeWidth{\mbox{36.7 (6.91)}}
\ptmeasure\ptTimeWidth{\mbox{388 (597)}}
\ptmeasure\ptTimeWidth{\mbox{40.1 (9.61)}}
\ptmeasure\ptTimeWidth{\mbox{40.4 (7.04)}}
\ptmeasure\ptTimeWidth{\mbox{41.4 (8.06)}}
\ptmeasure\ptTimeWidth{\mbox{4110 (1520)}}
\ptmeasure\ptTimeWidth{\mbox{4200 (1330)}}
\ptmeasure\ptTimeWidth{\mbox{4200 (1580)}}
\ptmeasure\ptTimeWidth{\mbox{435 (701)}}
\ptmeasure\ptTimeWidth{\mbox{45.0 (22.5)}}
\ptmeasure\ptTimeWidth{\mbox{45.5 (23.9)}}
\ptmeasure\ptTimeWidth{\mbox{466 (870)}}
\ptmeasure\ptTimeWidth{\mbox{471 (788)}}
\ptmeasure\ptTimeWidth{\mbox{49.0 (25.1)}}
\ptmeasure\ptTimeWidth{\mbox{509 (859)}}
\ptmeasure\ptTimeWidth{\mbox{537 (997)}}
\ptmeasure\ptTimeWidth{\mbox{558 (107)}}
\ptmeasure\ptTimeWidth{\mbox{560 (53.3)}}
\ptmeasure\ptTimeWidth{\mbox{560 (80.1)}}
\ptmeasure\ptTimeWidth{\mbox{567 (109)}}
\ptmeasure\ptTimeWidth{\mbox{567 (80.4)}}
\ptmeasure\ptTimeWidth{\mbox{567 (95.9)}}
\ptmeasure\ptTimeWidth{\mbox{590 (99.2)}}
\ptmeasure\ptTimeWidth{\mbox{591 (56.5)}}
\ptmeasure\ptTimeWidth{\mbox{604 (66.1)}}
\ptmeasure\ptTimeWidth{\mbox{61.0 (6.40)}}
\ptmeasure\ptTimeWidth{\mbox{61.1 (6.02)}}
\ptmeasure\ptTimeWidth{\mbox{62.3 (8.97)}}
\ptmeasure\ptTimeWidth{\mbox{66.5 (17.1)}}
\ptmeasure\ptTimeWidth{\mbox{67.4 (7.46)}}
\ptmeasure\ptTimeWidth{\mbox{68.2 (12.1)}}
\ptmeasure\ptTimeWidth{\mbox{7110 (1350)}}
\ptmeasure\ptTimeWidth{\mbox{7260 (1860)}}
\ptmeasure\ptTimeWidth{\mbox{7400 (1940)}}
\ptmeasure\ptTimeWidth{\mbox{900 (640)}}
\ptmeasure\ptTimeWidth{\mbox{911 (469)}}
\ptmeasure\ptTimeWidth{\mbox{933 (211)}}
\ptmeasure\ptTimeWidth{\mbox{938 (112)}}
\ptmeasure\ptTimeWidth{\mbox{957 (205)}}
\def\ptUpdateWidth{\edef\ptTableWidth{\the\dimexpr\ptMethodWidth+\ptSrmseWidth+\ptFnWidth+\ptFpWidth+\ptIterWidth+\ptTimeWidth+\ptTimeWidth+\ptTimeWidth+14\tabcolsep\relax}}
\ptUpdateWidth\ifdim\ptTableWidth>\linewidth\setlength{\tabcolsep}{1pt}\ptUpdateWidth\fi
\typeout{PERFORMANCE-WIDTH main_satellite: table=\ptTableWidth; time=\ptTimeWidth; linewidth=\the\linewidth}
\ifdim\ptTableWidth>\linewidth\PackageError{performance-tables}{Table exceeds linewidth}{Check the table layout.}\fi
\setlength{\LTcapwidth}{\ptTableWidth}
\Needspace{570pt}
\begin{longtable}{@{}>{\raggedright\arraybackslash}p{\ptMethodWidth}>{\raggedleft\arraybackslash}p{\ptSrmseWidth}>{\raggedleft\arraybackslash}p{\ptFnWidth}>{\raggedleft\arraybackslash}p{\ptFpWidth}>{\raggedleft\arraybackslash}p{\ptIterWidth}>{\raggedleft\arraybackslash}p{\ptTimeWidth}>{\raggedleft\arraybackslash}p{\ptTimeWidth}>{\raggedleft\arraybackslash}p{\ptTimeWidth}@{}}
\caption{Satellite drag, three responses fitted separately; $n=500$, $d=14$, $m=100$, $B=100$. The reference set contains the seven physical inputs.}\label{tab:satellite}\\
\toprule
Method & SRMSE & FN & FP & Iter & \makecell[r]{1-core\\(Sec)} & \makecell[r]{2-core\\(Sec)} & \makecell[r]{4-core\\(Sec)} \\
\midrule\endfirsthead
\caption[]{Satellite drag, three responses fitted separately; $n=500$, $d=14$, $m=100$, $B=100$. The reference set contains the seven physical inputs. (continued).}\\\toprule
Method & SRMSE & FN & FP & Iter & \makecell[r]{1-core\\(Sec)} & \makecell[r]{2-core\\(Sec)} & \makecell[r]{4-core\\(Sec)} \\
\midrule\endhead
\midrule
\multicolumn{8}{@{}p{\ptTableWidth}@{}}{\footnotesize Values are reported as mean (sample SD). Prediction, FN/FP and Iter use 1-core. An em dash in Iter denotes inapplicability. NA denotes no approved binary selection rule.}\\
\multicolumn{8}{r@{}}{\footnotesize Continued on next page.}\\\endfoot
\bottomrule
\endlastfoot
\multicolumn{8}{@{}l}{\textbf{Response: $\mathrm{He}$}}\\*
\texttt{optK}\textsubscript{0.005} & \mbox{0.0575 (0.00642)} & \mbox{0.63 (0.49)} & \mbox{1.88 (1.34)} & \mbox{39.3 (3.9)} & \mbox{315 (140)} & \mbox{250 (130)} & \mbox{170 (77.6)} \\*
\texttt{optK-pruned} & \mbox{0.0566 (0.00645)} & \mbox{0.94 (0.24)} & \mbox{0.61 (0.62)} & \mbox{39.2 (3.5)} & \mbox{339 (142)} & \mbox{266 (133)} & \mbox{183 (84.4)} \\*
\texttt{optK}\textsubscript{0.05} & \mbox{0.0715 (0.0199)} & \mbox{0.97 (0.22)} & \mbox{0.16 (0.37)} & \mbox{17.4 (2.7)} & \mbox{30.1 (9.20)} & \mbox{23.8 (8.14)} & \mbox{20.0 (5.87)} \\*
\texttt{laGP} & \mbox{0.0973 (0.00749)} & \mbox{NA} & \mbox{NA} & \mbox{---} & \mbox{49.0 (25.1)} & \mbox{45.5 (23.9)} & \mbox{45.0 (22.5)} \\*
\texttt{mlegp} & \mbox{0.169 (0.0783)} & \mbox{0.49 (0.52)} & \mbox{3.87 (2.25)} & \mbox{---} & \mbox{604 (66.1)} & \mbox{560 (53.3)} & \mbox{558 (107)} \\*
\texttt{MRFA} & \mbox{0.0850 (0.00833)} & \mbox{0.61 (0.49)} & \mbox{2.39 (1.32)} & \mbox{---} & \mbox{22.1 (4.42)} & \mbox{36.7 (6.91)} & \mbox{40.1 (9.61)} \\*
\texttt{DART} & \mbox{0.0919 (0.00715)} & \mbox{0.95 (0.22)} & \mbox{0.42 (0.54)} & \mbox{---} & \mbox{133 (18.4)} & \mbox{67.4 (7.46)} & \mbox{62.3 (8.97)} \\*
\texttt{SVGP-ARD} & \mbox{0.150 (0.00923)} & \mbox{NA} & \mbox{NA} & \mbox{---} & \mbox{250 (64.9)} & \mbox{388 (597)} & \mbox{471 (788)} \\*
\texttt{DGP-2} & \mbox{0.0738 (0.00570)} & \mbox{NA} & \mbox{NA} & \mbox{---} & \mbox{1540 (393)} & \mbox{933 (211)} & \mbox{900 (640)} \\*
\texttt{AS-GP} & \mbox{0.0443 (0.00328)} & \mbox{NA} & \mbox{NA} & \mbox{---} & \mbox{12900 (2270)} & \mbox{7110 (1350)} & \mbox{4200 (1580)} \\*
\midrule
\multicolumn{8}{@{}l}{\textbf{Response: $\mathrm{N}_2$}}\\*
\texttt{optK}\textsubscript{0.005} & \mbox{0.0549 (0.00463)} & \mbox{0.74 (0.44)} & \mbox{1.67 (1.18)} & \mbox{36.8 (3.9)} & \mbox{244 (111)} & \mbox{204 (130)} & \mbox{172 (288)} \\*
\texttt{optK-pruned} & \mbox{0.0575 (0.00809)} & \mbox{0.95 (0.22)} & \mbox{0.52 (0.64)} & \mbox{35.5 (4.9)} & \mbox{229 (160)} & \mbox{177 (81.4)} & \mbox{131 (62.7)} \\*
\texttt{optK}\textsubscript{0.05} & \mbox{0.0943 (0.0396)} & \mbox{1.21 (0.43)} & \mbox{0.15 (0.36)} & \mbox{15.0 (3.0)} & \mbox{23.2 (8.33)} & \mbox{19.7 (6.56)} & \mbox{17.1 (5.28)} \\*
\texttt{laGP} & \mbox{0.107 (0.00975)} & \mbox{NA} & \mbox{NA} & \mbox{---} & \mbox{32.7 (23.7)} & \mbox{30.4 (25.4)} & \mbox{30.1 (22.2)} \\*
\texttt{mlegp} & \mbox{0.163 (0.0943)} & \mbox{0.53 (0.63)} & \mbox{3.92 (1.76)} & \mbox{---} & \mbox{590 (99.2)} & \mbox{567 (80.4)} & \mbox{567 (109)} \\*
\texttt{MRFA} & \mbox{0.0667 (0.00668)} & \mbox{0.85 (0.36)} & \mbox{2.34 (1.07)} & \mbox{---} & \mbox{23.0 (6.97)} & \mbox{36.1 (5.98)} & \mbox{40.4 (7.04)} \\*
\texttt{DART} & \mbox{0.0779 (0.00643)} & \mbox{0.98 (0.14)} & \mbox{0.15 (0.39)} & \mbox{---} & \mbox{131 (20.5)} & \mbox{66.5 (17.1)} & \mbox{61.1 (6.02)} \\*
\texttt{SVGP-ARD} & \mbox{0.166 (0.0108)} & \mbox{NA} & \mbox{NA} & \mbox{---} & \mbox{244 (45.4)} & \mbox{466 (870)} & \mbox{509 (859)} \\*
\texttt{DGP-2} & \mbox{0.0802 (0.00645)} & \mbox{NA} & \mbox{NA} & \mbox{---} & \mbox{1600 (354)} & \mbox{957 (205)} & \mbox{911 (469)} \\*
\texttt{AS-GP} & \mbox{0.0577 (0.00410)} & \mbox{NA} & \mbox{NA} & \mbox{---} & \mbox{13000 (2480)} & \mbox{7400 (1940)} & \mbox{4110 (1520)} \\
\midrule
\multicolumn{8}{@{}l}{\textbf{Response: $\mathrm{O}_2$}}\\*
\texttt{optK}\textsubscript{0.005} & \mbox{0.0521 (0.00515)} & \mbox{0.69 (0.46)} & \mbox{1.92 (1.21)} & \mbox{37.2 (3.4)} & \mbox{251 (126)} & \mbox{202 (94.4)} & \mbox{148 (90.6)} \\*
\texttt{optK-pruned} & \mbox{0.0512 (0.00880)} & \mbox{0.86 (0.35)} & \mbox{0.59 (0.68)} & \mbox{34.6 (5.3)} & \mbox{220 (128)} & \mbox{187 (151)} & \mbox{128 (66.7)} \\*
\texttt{optK}\textsubscript{0.05} & \mbox{0.0702 (0.0262)} & \mbox{1.02 (0.20)} & \mbox{0.10 (0.30)} & \mbox{14.2 (2.3)} & \mbox{24.6 (9.39)} & \mbox{19.1 (5.94)} & \mbox{16.6 (5.23)} \\*
\texttt{laGP} & \mbox{0.103 (0.00761)} & \mbox{NA} & \mbox{NA} & \mbox{---} & \mbox{26.2 (9.61)} & \mbox{24.3 (8.79)} & \mbox{24.2 (8.70)} \\*
\texttt{mlegp} & \mbox{0.167 (0.104)} & \mbox{0.50 (0.59)} & \mbox{4.17 (1.83)} & \mbox{---} & \mbox{591 (56.5)} & \mbox{567 (95.9)} & \mbox{560 (80.1)} \\*
\texttt{MRFA} & \mbox{0.0688 (0.00536)} & \mbox{0.59 (0.49)} & \mbox{2.43 (1.27)} & \mbox{---} & \mbox{22.5 (6.51)} & \mbox{36.3 (7.00)} & \mbox{41.4 (8.06)} \\*
\texttt{DART} & \mbox{0.0776 (0.00560)} & \mbox{0.98 (0.14)} & \mbox{0.15 (0.39)} & \mbox{---} & \mbox{131 (20.6)} & \mbox{68.2 (12.1)} & \mbox{61.0 (6.40)} \\*
\texttt{SVGP-ARD} & \mbox{0.151 (0.00957)} & \mbox{NA} & \mbox{NA} & \mbox{---} & \mbox{242 (47.0)} & \mbox{435 (701)} & \mbox{537 (997)} \\*
\texttt{DGP-2} & \mbox{0.0750 (0.00666)} & \mbox{NA} & \mbox{NA} & \mbox{---} & \mbox{1640 (337)} & \mbox{938 (112)} & \mbox{1160 (2460)} \\*
\texttt{AS-GP} & \mbox{0.0592 (0.00462)} & \mbox{NA} & \mbox{NA} & \mbox{---} & \mbox{13200 (2500)} & \mbox{7260 (1860)} & \mbox{4200 (1330)} \\
\end{longtable}\endgroup

\begingroup\spacingset{1}
\begin{figure}[H]\centering
\includegraphics[width=\linewidth]{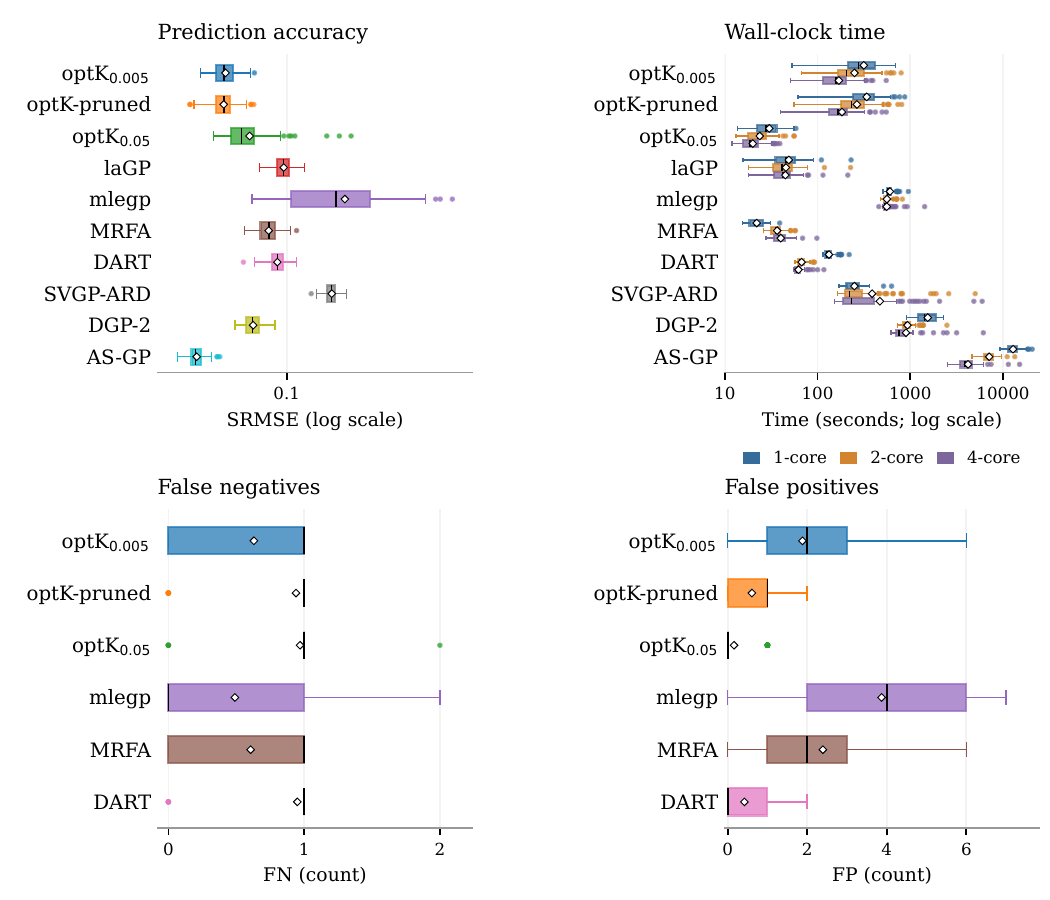}
\caption{Prediction, time and variable selection for Satellite He; $n=500$, $d=14$, $m=100$.}\label{fig:main-satellite-he}
\end{figure}
\begin{figure}[H]\centering
\includegraphics[width=\linewidth]{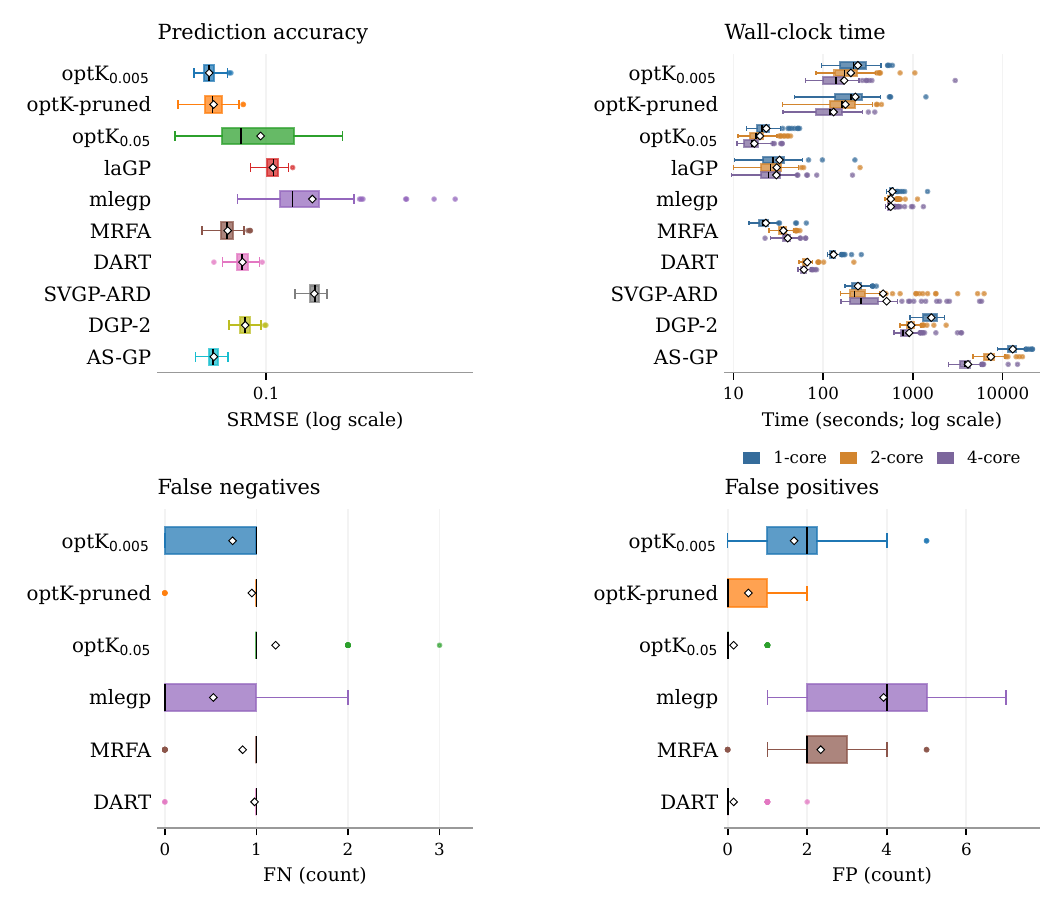}
\caption{Prediction, time and variable selection for Satellite $\mathrm{N}_2$; $n=500$, $d=14$, $m=100$.}\label{fig:main-satellite-n2}
\end{figure}
\begin{figure}[H]\centering
\includegraphics[width=\linewidth]{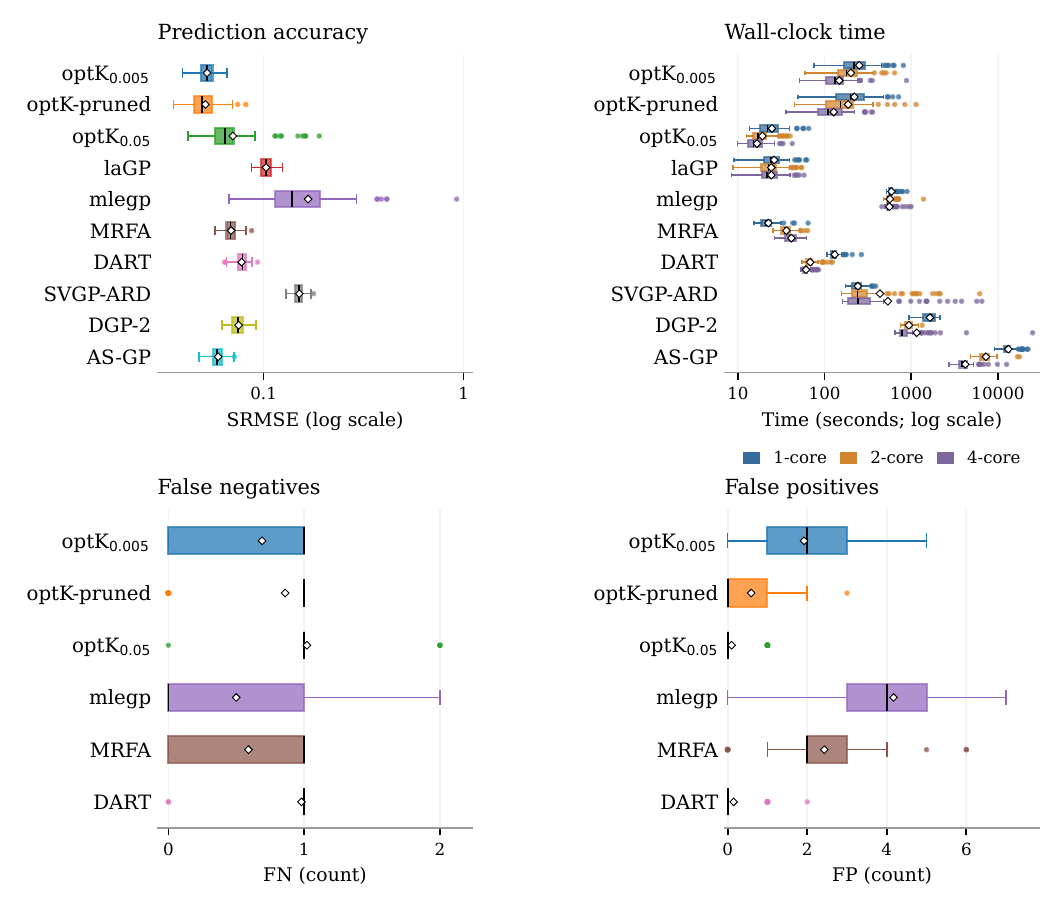}
\caption{Prediction, time and variable selection for Satellite $\mathrm{O}_2$; $n=500$, $d=14$, $m=100$.}\label{fig:main-satellite-o2}
\end{figure}
\endgroup

\section{Conclusion}\label{sec:end}

In this paper, we propose an optimal kernel learning approach to overcome the bottleneck of high-dimensional input for GP regression models. 
It approximates the covariance kernel function of the original input variables via a convex combination of kernel functions of lower-dimensional input variables. 
This is equivalent to approximating the original GP by functional ANOVA decomposition, which leads to an additive GP model with each added component being a GP of low-dimensional input space.
The optimal kernel learning problem is formulated as a minimization of a regularized least-squares loss function. 
Inspired by a similar optimization problem and the Fedorov-Wynn-type of algorithm in optimal design, we develop an optimal kernel learning algorithm that sequentially selects kernels from the candidate set and simultaneously updates the weights of all selected kernels. 
For high-dimensional input data, the candidate set of low-dimensional kernels can be large, so the computation needed to prepare the candidate kernels also increases drastically.
To overcome this issue, we follow the effect heredity principle from the design and analysis of experiments literature, and construct the candidate set of kernel functions stage-wise based on the previously selected input variables. 

GP regression is one special case of the kernel methods that compute inner products between data points using kernel functions and enable efficient non-linear learning across diverse domains. 
By mapping data to spaces where linear separation becomes possible, kernel methods provide remarkable flexibility, computational efficiency, and robust generalization capabilities. 
Selecting appropriate kernel functions and associated parameters continues to be a critical theoretical and computational problem. 
The performance of kernel methods heavily depends on factors such as function smoothness and bandwidth parameters. 
This proposed optimal kernel learning framework can be widely applied across kernel-based techniques, including support vector machine classification, spline methods, and even deep kernel networks. 
Another possible extension is to multi-output GP models, where the covariance structure must account for both input dependence and cross-output correlations.
Extending the proposed kernel learning framework to such settings, particularly with complex temporal dependence, would require additional covariance modeling and optimization developments.
We are going to pursue these directions in the future and address the fundamental bottleneck in kernel-based statistical learning methods: effectively handling high-dimensional input data.

\noindent{\bf \large Declaration of Generative AI Use}

Generative artificial intelligence (AI) tools were used in the preparation of this work. Specifically, GPT-5.6 Sol was used to assist with polishing the writing, drafting large simulation procedures and job-submission scripts, and formatting simulation results as LaTeX tables. All AI-generated content was reviewed, edited, and verified by the authors. The authors take full responsibility for the accuracy, integrity, and final presentation of the work.

\bibliography{Ref}
 
\newpage 

\setcounter{page}{1}
\setcounter{figure}{0}
\setcounter{table}{0}

\setcounter{section}{0}
\renewcommand{\thesection}{S\arabic{section}}

\makeatletter 
\renewcommand{\thefigure}{S\@arabic\c@figure}
\renewcommand{\thetable}{S\@arabic\c@table}

\makeatother

\begin{center}

{\large\bf SUPPLEMENTARY MATERIAL}

\end{center}

The supplementary materials include the following parts: (1) an optimal weight procedure given support kernels summarized in Algorithm \ref{alg:multiplicative}, which is based on the corollary of Theorem \ref{thm:GE}; (2) detailed discussion on tuning parameters; (3) proofs, and derivations; (4) discussion on kernel parameters; and (5) extra figures for the examples. 
The codes for all the examples are available from this link \url{https://github.com/tonyxms/ConvexKernel}. 

\section{An Optimal Weight Procedure Given Support Kernels}\label{Ssec:multiplicative}

After a new support kernel is added to the current design, we need to update the weights of all the current support kernels. 
We first obtain the conditions of optimal weights presented in the following Corollary \ref{thm:weight} and then develop Algorithm \ref{alg:multiplicative} to update the weights given the support kernels. 

\begin{corollary}[Conditions of Optimal Weights]\label{thm:weight}
Restrict the set of basic kernel $\mathcal{G}$ to be a finite set, $\mathcal{G}=\{K_1,\ldots, K_M\}$ and $\Xi$ is the class of discrete measure on $\mathcal{G}$. 
For any $\xi\in \Xi$, the corresponding weight vector $\bm \lambda=[\lambda_1,\ldots, \lambda_M]^\top$ with $0\leq \lambda_i\leq 1$ becomes the only variable that decides $Q_{\eta}(\xi)$. 
The following two conditions on the optimal design $\xi^*$ and its weight vector $\bm \lambda^*$ are equivalent. 
 \begin{enumerate}
\item The weight vector $\bm{\lambda}^{*}$ minimizes $Q_{\eta}(\xi)$;
\item For all $K_i$ with $\lambda_i^*>0$, $\phi(K_i,\xi^*)=0$; for all $K_i$ with $\lambda_i^{*}=0$, $\phi(K_i,\xi^*)\geq 0$.
\end{enumerate}
\end{corollary} 

Consider any design $\xi$ in Corollary \ref{thm:weight} with weight vector $\bm \lambda$. 
Its corresponding kernel function is $K(\xi)$ and the kernel matrix is $\bm K$.
As $K(\xi)=\sum_{i=1}^M \lambda_iK_i$, so is true for $\bm K=\sum_{i=1}^M \lambda_i \bm K_i$. 
For such a design $\xi$, we define 
\begin{equation}
\label{eq:d-fun}
d(K_i,\xi):=\bm y^\top((\bm K+\eta \bm I_n)^{-1}\bm K_i(\bm K+\eta \bm I_n)^{-1})\bm y.
\end{equation}

Corollary \ref{thm:weight} is a special case of Theorem \ref{thm:GE} when $\mathcal{G}$ is finite, but it can be extended to the more general case where $\mathcal{G}$ is not finite but countable. 
Based on it, given the set of kernels $K_1,\cdots,K_M$, a sufficient condition that $\bm{\lambda}^{*}$ minimizes $Q_{\eta}(\xi)$ is $\phi(K_i,\xi^*)=0$ for $i=1,\cdots,M$, or equivalently,
\begin{equation}\label{eq:corov2}
\bm y^\top (\bm K^*+\eta \bm I_n)^{-1}\bm K_i(\bm K^*+\eta \bm I_n)^{-1}\bm y=\bm y^\top (\bm K^*+\eta \bm I_n)^{-1}\bm K^*(\bm K^*+\eta \bm I_n)^{-1}\bm y.
\end{equation}
Here $\bm K^*$ is the kernel matrix obtained by evaluating the kernel function $K(\xi^*)$ on $\mathcal{X}$ and $\bm K_i$ is the kernel matrix of the kernel function $K_i$. 
It is easy to see that \eqref{eq:corov2} is equivalent to 
\begin{equation}\label{weight equation}
\sum_{j=1}^M \lambda_j^* d(K_j,\xi^*)=d(K_i,\xi^*).
\end{equation}

To obtain the optimal weights $\bm{\lambda}^{*}$, the non-optimal weights of a design could be adjusted according to two sides of \eqref{weight equation}.
At the $k$th iteration, denote the current design $\xi^k$ with weights $\bm \lambda^{(k)}$ for the basic kernels $\mathcal{G}=\{K_1,\ldots, K_M\}$. 
The corresponding kernel function is $K(\xi^k)=\sum_{i=1}^M \lambda_i^{(k)}K_i$ and the kernel matrix is $\bm K^{(k)}=\sum_{i=1}^M \lambda_i^{(k)} \bm K_i$. 
If $d(K_i,\xi^k)>\sum_{j=1}^M \lambda_j d(K_j,\xi^k)$, then the weight of kernel $K_i$ should be increased based on \eqref{weight equation}.
Otherwise, if $d(K_i,\xi^k)<\sum_{j=1}^M \lambda_j d(K_j,\xi^k)$ , the weight of kernel $K_i$ should be decreased.
Therefore, following the similar idea in the multiplicative procedure used in \cite{silvey1978algorithm, yu2010monotonic, li2022maximin}, the ratio $\left[d(K_i,\xi^k)/\sum_{j=1}^M\lambda_j^{(k)}(d(K_j,\xi^k))\right]^{\delta}$
would be a good adjustment for the weight of kernel $K_i$. 
Here $\delta\in (0, 1]$ is a tuning parameter that controls the speed of convergence.
We update the weight of kernel $K_i$ at the $k$th iteration by
\begin{equation}\label{update-1}
{\widetilde\lambda_i^{(k+1)}}=\lambda_i^{(k)}\left[\frac{d(K_i,\xi^k)}{\sum_{j=1}^M\lambda_j^{(k)}d(K_j,\xi^k)}\right]^{\delta}.
\end{equation}
Then we normalize the weights to ensure the sum to one condition by
\begin{equation}\label{update-2}
\lambda_i^{(k+1)}=\frac{{\widetilde\lambda_i^{(k+1)}}}{\sum_{j=1}^M{\widetilde\lambda_j^{(k+1)}}}
\end{equation}
Plugging \eqref{update-1} into \eqref{update-2}, we get \eqref{eq:update} in Algorithm \ref{alg:multiplicative}.

\begin{algorithm}[H]
\caption{Optimal Weight Procedure: A Modified Multiplicative Algorithm \label{alg:multiplicative}}
\begin{algorithmic}[1]
\State {\bf Setup:}  Given the set of basic kernels $\mathcal{G}=\{K_1,\ldots,K_M\}$ of size $M$, set the following parameters.
\begin{itemize}
\item $\delta\in (0, 1]$: controls the speed of convergence, and we set $\delta=1$ by default;
\item \texttt{Tol}: a small positive value to check convergence, which can be the same as in Algorithm \ref{alg:fed-wynn};
\item \texttt{MaxIter}$_0$: a large integer as the maximum iterations allowed.
\end{itemize} 
\State {\bf Initiate:} Assign a uniform initial weight vector $\bm{\lambda}^{(0)}=[\lambda_1^{(0)},\cdots,\lambda_M^{(0)}]^\top$. Set the counter $k=0$ and initial value \texttt{change}$_0$=1. 
\While {\texttt{change}$_0 > $ \texttt{Tol} and $k < $ \texttt{MaxIter}$_0$}
\For {$i=1,\ldots,M$}
\State Update the weight of kernel $K_i$:
\begin{equation}\label{eq:update}
\begin{aligned}
\lambda_i^{(k+1)}&=\lambda_i^{(k)}\frac{\left[d(K_i,\xi^k)\right]^{\delta}}{\sum\limits_{j=1}^M\lambda_j^{(k)}\left[d(K_j,\xi^k)\right]^{\delta}}\\
&=\lambda_i^{(k)}\frac{\left[\bm y^\top(\bm K^{(k)}+\eta \bm I_n)^{-1}\bm K_i(\bm K^{(k)}+\eta \bm I_n)^{-1}\bm y\right]^{\delta}}{\sum\limits_{j=1}^M\lambda_j^{(k)}\left[\bm y^\top(\bm K^{(k)}+\eta \bm I_n)^{-1}\bm K_j(\bm K^{(k)}+\eta \bm I_n)^{-1}\bm y\right]^{\delta}}
\end{aligned}
\end{equation}
\EndFor
\State \texttt{change}$_0=\bigg|\frac{Q_\eta(\xi^{k+1})-Q_\eta(\xi^{k})}{Q_\eta(\xi^{k})}\bigg|$
\State $k \leftarrow k+1$
\EndWhile
\end{algorithmic}
\end{algorithm}

It is easy to see that Algorithm \ref{alg:multiplicative} solves a convex minimization problem since $Q_{\eta}$ is convex w.r.t. $\bm \lambda$ and the constraints $\sum_{i=1}^M \lambda_i=1$ and $\lambda_i\geq 0$ for all $i$'s are convex. 
Therefore, convex optimization methods other than the multiplicative algorithm can be used here.
We chose the multiplicative algorithm because \cite{yu2010monotonic} established the monotonic convergence for a general class of multiplicative algorithms. 
Theorem \ref{thm:alg3} gives monotonic convergence of Algorithm \ref{alg:multiplicative}. 
\begin{theorem}[Monotonic convergence of Algorithm \ref{alg:multiplicative}]\label{thm:alg3}
Let $\mathcal G=\{K_1,\ldots,K_M\}$ be a fixed finite collection of basic kernels, with corresponding strictly positive definite kernel matrices $\bm K_1,\ldots, \bm K_M$.
Define the feasible weight set
\[
\Lambda_M = \left\{ \bm\lambda\in\mathbb R^M: \lambda_i\geq 0,\quad \sum_{i=1}^M\lambda_i=1 \right\}.
\]
For any $\bm\lambda\in\Lambda_M$, let
\[
\bm K(\bm\lambda)=\sum_{i=1}^M\lambda_i \bm K_i, \qquad Q_\eta(\bm\lambda) = \eta\mathbf y^\top \left(\bm K(\bm\lambda)+\eta \bm I_n\right)^{-1} \mathbf y.
\]
Suppose that $\eta>0$ and $\mathbf y\neq\mathbf 0$. 
Fix $\delta\in(0,1]$ and an initial vector $\bm\lambda^{(0)}\in\Lambda_M$ with $\lambda_i^{(0)}>0$ for every $i$.
Let $\{\bm\lambda^{(k)}\}_{k\geq0}$ be generated by the simultaneous multiplicative update
\[
\lambda_i^{(k+1)}
=\frac{\lambda_i^{(k)} [d_i(\bm\lambda^{(k)})]^\delta}{\sum_{j=1}^M\lambda_j^{(k)}[d_j(\bm\lambda^{(k)})]^\delta}, \qquad i=1,\ldots,M,
\]
where
\[
d_i(\bm\lambda) = \mathbf y^\top \left(\bm K(\bm\lambda)+\eta\bm I_n\right)^{-1} \bm K_i \left(\bm K(\bm\lambda)+\eta\bm I_n\right)^{-1} \mathbf y.
\]
The iteration is performed without finite stopping criteria or deletion of components.
Then the following statements hold:
\begin{enumerate}
\item[(1)]
The objective values are monotonically nonincreasing:
\[
Q_\eta(\bm\lambda^{(k+1)}) \leq Q_\eta(\bm\lambda^{(k)}), \qquad k\geq 0.
\]
Moreover, the inequality is strict whenever $\bm\lambda^{(k+1)}\neq\bm\lambda^{(k)}$.
\item[(2)]
The objective values converge to the global minimum:
\[
\lim_{k\to\infty} Q_\eta(\bm\lambda^{(k)}) = \min_{\bm\lambda\in\Lambda_M} Q_\eta(\bm\lambda).
\]
\item[(3)] Every accumulation point of $\{\bm\lambda^{(k)}\}_{k\geq 0}$ is a global minimizer of $Q_\eta$ over $\Lambda_M$.
If the global minimizer $\bm\lambda^*$ is unique, then
\[
\lim_{k\to\infty}\bm\lambda^{(k)} = \bm\lambda^*.
\]
\end{enumerate}
\end{theorem}

\begin{proof}
For notational convenience, write
\[
\bm A(\bm\lambda) =\bm K(\bm\lambda)+\eta\bm I_n, \qquad \bm\alpha(\bm\lambda) =\bm A(\bm\lambda)^{-1}\mathbf y,
\]
and define
\[
d_i(\bm\lambda) =\bm\alpha(\bm\lambda)^\top \bm K_i\bm\alpha(\bm\lambda), \qquad 
Z(\bm\lambda) =\sum_{j=1}^M\lambda_jd_j(\bm\lambda)^\delta.
\]
Since $\bm K_i\succ0$ and $\mathbf y\neq\mathbf 0$, each $d_i$ is positive and continuous on $\Lambda_M$.
Thus the update map
\begin{equation}
T_i(\bm\lambda) =\frac{\lambda_i d_i(\bm\lambda)^\delta} {Z(\bm\lambda)}, \qquad i=1,\ldots,M,
\label{eq:alg3_update_proof}
\end{equation}
is continuous on $\Lambda_M$ and preserves the support of $\bm\lambda$. In particular, strictly positive initial weights remain strictly positive at every finite iteration.

\noindent\emph{Step 1: A majorization inequality.}
Let $\bm\lambda,\bm\mu\in\Lambda_M$ have the same support $I=\{i:\lambda_i>0\}$, and put $\bm\alpha=\bm\alpha(\bm\lambda)$.
Define
\[
\mathbf z_0=\eta\bm\alpha, \qquad \mathbf z_i=\lambda_i\bm K_i\bm\alpha, \quad i\in I.
\]
Then $\mathbf z_0+\sum_{i\in I}\mathbf z_i=\mathbf y$.
For positive definite matrices $\bm B_i$, the quadratic variational identity
\begin{equation}
\mathbf y^\top \left(\sum_i\bm B_i\right)^{-1}\mathbf y = \min_{\sum_i\mathbf u_i=\mathbf y} \sum_i\mathbf u_i^\top\bm B_i^{-1}\mathbf u_i
\label{eq:quadratic_variational_identity}
\end{equation}
has the unique minimizer $\mathbf u_i=\bm B_i(\sum_j\bm B_j)^{-1}\mathbf y$.
Applying this identity with $\bm B_0=\eta\bm I_n$ and $\bm B_i=\mu_i\bm K_i$ for $i\in I$, and evaluating the objective at the vectors $\mathbf z_i$, gives
\begin{equation}
\mathbf y^\top\bm A(\bm\mu)^{-1}\mathbf y \leq \eta\|\bm\alpha\|^2 +\sum_{i\in I} \frac{\lambda_i^2d_i(\bm\lambda)}{\mu_i}.
\label{eq:alg3_majorization}
\end{equation}
Equality holds when $\bm\mu=\bm\lambda$.

\noindent\emph{Step 2: Monotonicity and strict descent.}
For any $\bm\lambda\in\Lambda_M$, put $\bm\lambda^+=T(\bm\lambda)$, $d_i=d_i(\bm\lambda)$, and $Z=Z(\bm\lambda)$.
Substituting \eqref{eq:alg3_update_proof} into \eqref{eq:alg3_majorization} yields
\begin{align*}
\mathbf y^\top\bm A(\bm\lambda^+)^{-1}\mathbf y &\leq \eta\|\bm\alpha\|^2 +\left(\sum_{i\in I}\lambda_i d_i^\delta\right) \left(\sum_{i\in I}\lambda_i d_i^{1-\delta}\right)\nonumber\\
&\leq \eta\|\bm\alpha\|^2+\sum_{i\in I}\lambda_i d_i = \mathbf y^\top\bm A(\bm\lambda)^{-1}\mathbf y.
\end{align*}
The second inequality follows from Jensen's inequality when $0<\delta<1$ and is an equality when $\delta=1$.
Multiplying by $\eta>0$ proves monotonicity.

For $0<\delta<1$, equality in Jensen's inequality requires all $d_i$, $i\in I$, to be equal, which implies $\bm\lambda^+=\bm\lambda$.
For $\delta=1$, equality of the objective values requires equality in \eqref{eq:alg3_majorization}.
Uniqueness of the minimizer in \eqref{eq:quadratic_variational_identity} then gives
\[
\eta\bm\alpha =\eta\bm A(\bm\lambda^+)^{-1}\mathbf y, \qquad \lambda_i\bm K_i\bm\alpha =\lambda_i^+\bm K_i \bm A(\bm\lambda^+)^{-1}\mathbf y, \quad i\in I.
\]
Hence $(\lambda_i-\lambda_i^+)\bm K_i\bm\alpha=\mathbf 0$. 
Since $\bm K_i$ is invertible and $\bm\alpha\neq\mathbf 0$, we obtain $\lambda_i=\lambda_i^+$ for every $i\in I$.
Therefore, on the entire simplex,
\begin{equation}
Q_\eta(T(\bm\lambda))<Q_\eta(\bm\lambda) \quad\text{whenever}\quad T(\bm\lambda)\neq\bm\lambda.
\label{eq:alg3_strict_descent}
\end{equation}
This establishes part (1).

\noindent \emph{Step 3: Fixed points and accumulation points.}
By monotonicity and nonnegativity, there exists $Q_\infty\geq0$ such that
\begin{equation}
Q_\eta(\bm\lambda^{(k)})\longrightarrow Q_\infty.
\label{eq:alg3_objective_limit}
\end{equation}
Compactness of $\Lambda_M$ ensures the existence of accumulation points.
If $\bm\lambda^{(k_\ell)}\to\bar{\bm\lambda}$, continuity of $T$ and $Q_\eta$ gives
\[
Q_\eta(\bar{\bm\lambda}) =Q_\infty =Q_\eta(T(\bar{\bm\lambda})).
\]
By \eqref{eq:alg3_strict_descent},
\begin{equation}
T(\bar{\bm\lambda})=\bar{\bm\lambda}.
\label{eq:alg3_fixed_point}
\end{equation}
Thus every accumulation point is a fixed point.
Continuity and compactness also imply
\[
\|\bm\lambda^{(k+1)}-\bm\lambda^{(k)}\| \longrightarrow0;
\]
otherwise, a convergent subsequence would yield an accumulation point violating \eqref{eq:alg3_fixed_point}.

For later use, differentiation gives
\begin{equation}
\frac{\partial Q_\eta(\bm\lambda)}{\partial\lambda_i} =-\eta d_i(\bm\lambda).
\label{eq:alg3_gradient}
\end{equation}
Moreover, for $\bm H=\sum_i h_i\bm K_i$, $\bm A_t=\bm A(\bm\lambda+t\bm h)$, and $\bm\alpha_t=\bm A_t^{-1}\mathbf y$,
\begin{equation}
\frac{d^2}{dt^2}Q_\eta(\bm\lambda+t\bm h) = 2\eta(\bm H\bm\alpha_t)^\top \bm A_t^{-1}(\bm H\bm\alpha_t) \geq0.
\label{eq:alg3_convexity}
\end{equation}
Thus $Q_\eta$ is convex, which is also claimed in Lemma \ref{lemma:convex}. 
At any fixed point, the values $d_i$ are equal over its positive-weight coordinates. 
By \eqref{eq:alg3_gradient} and convexity, that fixed point minimizes $Q_\eta$ over the face determined by its support.

\noindent \emph{Step 4: Convergence of the inverse-weighted response.}
We first show that all minimizers over any given face have the same vector $\bm\alpha(\bm\lambda)$.
Write
\[
F(\bm\lambda)=\frac{Q_\eta(\bm\lambda)}{\eta} =\max_{\mathbf a\in\mathbb R^n} g_{\bm\lambda}(\mathbf a),
\qquad 
g_{\bm\lambda}(\mathbf a) =2\mathbf y^\top\mathbf a -\mathbf a^\top\bm A(\bm\lambda)\mathbf a.
\]
The maximizer is unique and equals $\bm\alpha(\bm\lambda)$.
Suppose $\bm\lambda$ and $\bm\mu$ minimize $F$ over the same face, with minimum value $q$.
Their midpoint $\bm\nu=(\bm\lambda+\bm\mu)/2$ also has value $q$.
For $\mathbf a_\nu=\bm\alpha(\bm\nu)$, affinity of $\bm A$ gives
\[
q =g_{\bm\nu}(\mathbf a_\nu) =\frac12g_{\bm\lambda}(\mathbf a_\nu)+\frac12g_{\bm\mu}(\mathbf a_\nu) \leq \frac12F(\bm\lambda)+\frac12F(\bm\mu) =q.
\]
Equality implies that $\mathbf a_\nu$ maximizes both $g_{\bm\lambda}$ and $g_{\bm\mu}$.
By uniqueness, $\bm\alpha(\bm\lambda)=\bm\alpha(\bm\mu)$.

Since the simplex has finitely many faces, Step 3 therefore implies that $\{\bm\alpha(\bm\lambda^{(k)})\}$ has only finitely many possible accumulation points.
Uniform continuity of $\bm\alpha$ and Step 3 give
\[
\|\bm\alpha(\bm\lambda^{(k+1)})-\bm\alpha(\bm\lambda^{(k)})\| \longrightarrow 0.
\]
A bounded sequence with vanishing successive differences and finitely many accumulation points must converge: distinct accumulation points can be enclosed in disjoint neighborhoods, and sufficiently small successive steps cannot pass between them without producing an additional accumulation point outside those neighborhoods.
Consequently,
\[
\bm\alpha(\bm\lambda^{(k)})\longrightarrow\bm\alpha^*
\]
for some $\bm\alpha^*$.

\noindent \emph{Step 5: Global optimality and convergence.}
Define
\[
d_i^*=(\bm\alpha^*)^\top\bm K_i\bm\alpha^*,
\qquad
b^*=(\bm\alpha^*)^\top\mathbf y -\eta\|\bm\alpha^*\|^2.
\]
For every $\bm\lambda\in\Lambda_M$,
\[
\sum_i\lambda_i d_i(\bm\lambda)
=\bm\alpha(\bm\lambda)^\top\mathbf y -\eta\|\bm\alpha(\bm\lambda)\|^2.
\]
Let $\bar{\bm\lambda}$ be any accumulation point.
Then $\bm\alpha(\bar{\bm\lambda})=\bm\alpha^*$, and the fixed-point property implies
\[
d_i^*=b^* \quad\text{whenever}\quad \bar\lambda_i>0.
\]
In particular, $b^*=(\bm\alpha^*)^\top \bm K(\bar{\bm\lambda})\bm\alpha^*>0$, and
\[
Z(\bar{\bm\lambda})=(b^*)^\delta.
\]
Since this holds at every accumulation point, compactness and continuity imply
\[
Z(\bm\lambda^{(k)})\longrightarrow(b^*)^\delta.
\]
If $d_i^*>b^*$ for some $i$, strict positivity of the finite-iteration weights gives
\[
\frac{\lambda_i^{(k+1)}}{\lambda_i^{(k)}} =\frac{d_i(\bm\lambda^{(k)})^\delta} {Z(\bm\lambda^{(k)})}
\longrightarrow \left(\frac{d_i^*}{b^*}\right)^\delta>1.
\]
This would force $\lambda_i^{(k)}$ eventually to grow geometrically, contradicting $\lambda_i^{(k)}\leq1$.
Therefore,
\begin{equation*}
d_i^*\leq b^*\quad\text{for every }i, \qquad d_i^*=b^*\quad\text{if }\bar\lambda_i>0.
\end{equation*}
For any $\bm\mu\in\Lambda_M$, convexity and
\eqref{eq:alg3_gradient} now yield
\begin{align*}
Q_\eta(\bm\mu) &\geq Q_\eta(\bar{\bm\lambda}) -\eta\sum_i d_i^*(\mu_i-\bar\lambda_i)\\
&= Q_\eta(\bar{\bm\lambda}) +\eta\left(b^*-\sum_i\mu_i d_i^*\right) \geq Q_\eta(\bar{\bm\lambda}).
\end{align*}
Thus every accumulation point is a global minimizer, proving the first assertion of part (3).
By continuity and \eqref{eq:alg3_objective_limit},
\[
Q_\infty =Q_\eta(\bar{\bm\lambda}) =\min_{\bm\lambda\in\Lambda_M}Q_\eta(\bm\lambda),
\]
which proves part (2).
Finally, if the global minimizer $\bm\lambda^*$ is unique, every accumulation point equals $\bm\lambda^*$; compactness then implies $\bm\lambda^{(k)}\to\bm\lambda^*$.
\end{proof}

\section{Tuning Parameters}\label{Ssec:tuningpara}

Parameters such as \texttt{Tol}, \texttt{MaxIter}, and \texttt{MaxIter}$_0$ are stopping rules and their settings should be chosen based on trial-and-error and the available computing time, which is similar to most optimization algorithms. 
We set $\texttt{Tol}=0.005$ and $\texttt{MaxIter}=\texttt{MaxIter}_0=1000$ in all the examples in Section \ref{sec:num}. 
The parameter \texttt{DEL} is the threshold for removing a certain support kernel if its weight is too small. 
A proper \texttt{DEL} setting can produce a sparse model with fewer support kernels in the optimal kernel. 
On the other hand, if \texttt{DEL} is too large, it leads to a sub-optimal solution and less accurate approximation of the underlying function $f(\bm x)$. 
In our simulations, we have tried a series of \texttt{DEL} values in $[0.01, 0.1]$ and discovered that $\texttt{DEL}=0.05$ gives sufficiently accurate results. 
In all the examples, we have compared three versions of \texttt{optK} as explained in Section \ref{sec:num}. 
It is not surprising that $\texttt{DEL}=0.005$ takes more computation time, but it leads to bigger false positive and comparable prediction accuracy compared with $\texttt{DEL}=0.05$.

The parameter $\delta \in (0,1]$ controls the speed of convergence.
Originally, $\delta$ was introduced in the classical multiplicative algorithm \citep{silvey1978algorithm}, and some settings were suggested based on different optimal design criteria \citep{torsney1983moment}. 
Unfortunately, such rules-of-thumb are unsuitable for our objective function, but our numerical simulations have shown that $\delta=1$ leads to sufficiently fast convergence. 

\begin{figure}[H]
\centering
\includegraphics[width=0.5\linewidth]{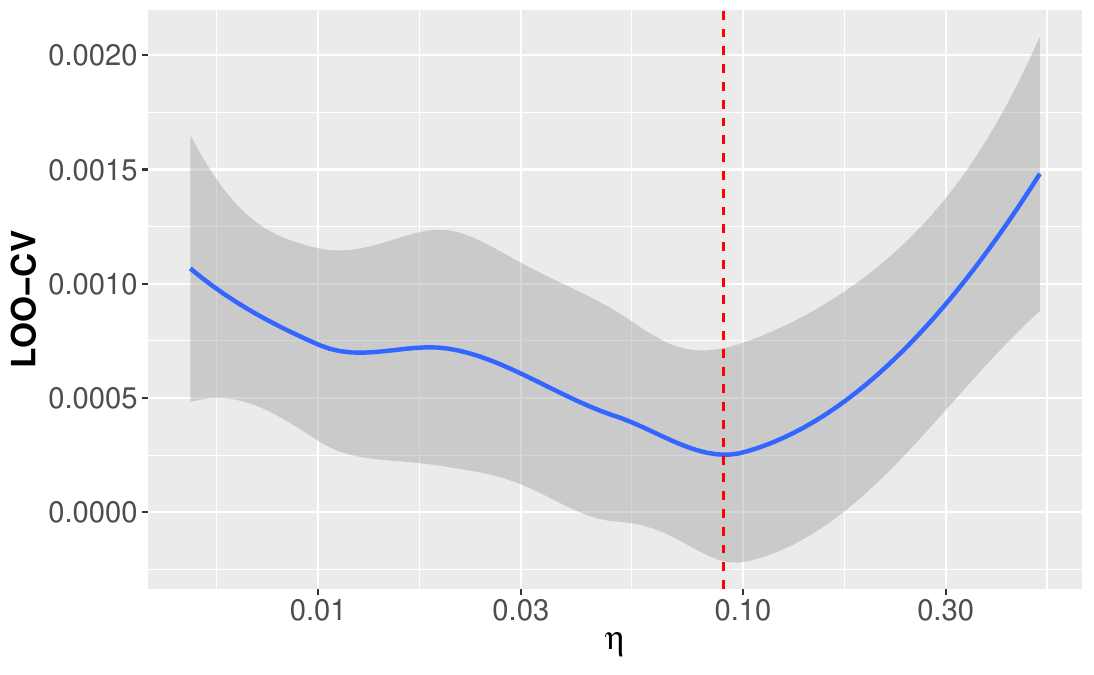}
\caption{Leave-one-out cross-validation error w.r.t. $\eta$.\label{fig:loocv}}
\end{figure}

We only tune $\eta$, the nugget effect in the probabilistic GP model, which is also the regularization parameter in nonparametric RKHS regression. 
Since the optimal kernel learning is from the nonparametric perspective, we use the leave-one-out cross-validation (LOO-CV) procedure to choose the optimal $\eta$. 
It entails computing the LOO-CV error for a given $\eta$ value with the RKHS regression whose kernel is obtained from the proposed Algorithm \ref{alg:fed-wynn}.  
To compute the LOO-CV error, the existing short-cut formula for GP regression \citep{dubrule1983cross} can be used to save computation. 
To show the effectiveness of the LOO-CV procedure, we use the Michalewicz test function with $p=2$ true active input variables. 
The surface of a two-dimensional Michalewicz function is shown in Figure \ref{fig:michale2d}. 
The training data has $n=200$ sample size and $d=6$ input dimension (four extra ``fake'' input dimensions).
The LOO-CV error curve in Figure \ref{fig:loocv} is plotted at the mean value of $10$ simulations for each $\eta$ value. 
The curve's minimum (marked by the red dashed vertical line) leads to the optimal $\eta$ value. 

\section{Proofs and Derivations}\label{Ssec:proofs}

\subsection{Proof of Lemma \ref{lemma:convex}}

\begin{proof}
Following the notation and assumption in Section \ref{sec:opt}, we have 
\[
Q_{\eta}((1-\alpha)K+\alpha K')\leq (1-\alpha)Q_{\eta}(K)+\alpha Q_{\eta}(K'), \quad \forall \alpha \in [0,1],
\]
due to the convexity of $Q_{\eta}(\cdot)$ in $\mathcal{K}$. 
Since $\mathcal{K}=\overline{\text{conv}(\mathcal{G})}$, let $\xi$ and $\xi'$ in $\Xi$ such that $K=\int G\xi(\dd G)$ and $K'=\int G\xi'(\dd G)$. 
Then, 
\begin{align*}
&Q_{\eta}((1-\alpha)\xi+\alpha \xi')=Q_{\eta}\left(K((1-\alpha)\xi+\alpha \xi')\right)=Q_{\eta}\left(\int G ((1-\alpha)\xi+\alpha \xi')(\dd G)\right)\\
=&Q_{\eta}\left(\int G ((1-\alpha)\xi)(\dd G)+\int G (\alpha \xi')(\dd G)\right)=Q_{\eta}\left((1-\alpha)\int G \xi(\dd G)+\alpha \int G \xi'(\dd G)\right)\\
=&Q_{\eta}\left((1-\alpha)K(\xi)+\alpha K(\xi')\right)\\
\leq & (1-\alpha)Q_{\eta}(K(\xi))+\alpha Q_{\eta}(K(\xi'))=(1-\alpha)Q_{\eta}(\xi)+\alpha Q_{\eta}(\xi').
\end{align*}
Therefore, $Q_{\eta}(\cdot)$ is also convex in $\Xi$. 
\end{proof}

\subsection{Proof of Proposition \ref{prop:dir-diff}}

\begin{proof}
To simplify the notation, we omit the design and let $K=K(\xi)$ and $K'=K(\xi')$, with corresponding kernel matrices $\bm K=\bm K_\xi$ and $\bm K'=\bm K_{\xi'}$. 
For fixed $\eta>0$, the matrix $\bm K+\eta\bm I_n$ is symmetric positive definite.
With $\bm c^*=(\bm K+\eta\bm I_n)^{-1}\bm y$, we have
\[
\bm y-\bm K\bm c^*=\bigl(\bm I_n-\bm K(\bm K+\eta\bm I_n)^{-1}\bigr)\bm y=\eta(\bm K+\eta\bm I_n)^{-1}\bm y.
\]
Substituting into \eqref{eq:min-minQL2} gives
\begin{align*}
Q_\eta(K)
&=\|\bm y-\bm K\bm c^*\|_2^2+\eta(\bm c^*)^\top\bm K\bm c^*\\
&=\eta^2\bm y^\top(\bm K+\eta\bm I_n)^{-2}\bm y+\eta\bm y^\top(\bm K+\eta\bm I_n)^{-1}\bm K(\bm K+\eta\bm I_n)^{-1}\bm y\\
&=\eta\bm y^\top(\bm K+\eta\bm I_n)^{-1}(\bm K+\eta\bm I_n)(\bm K+\eta\bm I_n)^{-1}\bm y\\
&=\eta\bm y^\top(\bm K+\eta\bm I_n)^{-1}\bm y.
\end{align*}

For a differentiable kernel path $K=K(\alpha)$ with corresponding kernel matrix $\bm K=\bm K(\alpha)$, the inverse-matrix derivative identity gives
\[
\frac{dQ_\eta(K)}{d\alpha}
=-\eta\bm y^\top(\bm K+\eta\bm I_n)^{-1}
\frac{d\bm K}{d\alpha}(\bm K+\eta\bm I_n)^{-1}\bm y.
\]
Now consider the path $(1-\alpha)K+\alpha K'$ for fixed $K,K'\in\mathcal K$ and $\alpha\in[0,1]$. We have
\[
\frac{d\bigl((1-\alpha)K+\alpha K'\bigr)}{d\alpha}=K'-K.
\]
The corresponding derivative of the kernel matrix is
\[
\frac{d\bigl((1-\alpha)\bm K+\alpha\bm K'\bigr)}{d\alpha}=\bm K'-\bm K.
\]
Evaluating the derivative along this path at $\alpha=0^+$ yields
\begin{align*}
\phi(K',K)
&=\left.\frac{d}{d\alpha}Q_\eta(K)\right|_{\alpha=0^+}\\
&=-\eta \bm y^\top ((\bm K+\eta \bm I_n)^{-1}(\bm K^{\prime}-\bm K)(\bm K+\eta \bm I_n)^{-1})\bm y.
\end{align*}
$\phi(\xi', \xi)$ follows by definition. 
\end{proof}

\subsection{Proof of Theorem \ref{thm:GE}}

\begin{proof}
We show the equivalence of the three statements by showing (i) the equivalence of (1) and (2); (ii) (1) is sufficient for (3); and (iii) (3) is sufficient for (2).
\begin{enumerate}
\item (1) $\leftrightarrow$ (2): Based on the convexity of $Q_{\eta}(\xi)$ in terms of $\xi$, the equivalence of (1) and (2) is a direct result. 
\item (1) $\to$ (3): Since $\phi(G, \xi^*)$ is simply $\phi(\xi_G, \xi^*)$, a special case of $\phi(\xi', \xi^*)$, and thus if (1) is true, we have $\phi(G,\xi^*)\geq 0$ for any $G\in \mathcal{G}$.

Next, we need to show that given (1), if $G$ is a support kernel of $\xi^*$, then $\phi(G,\xi^*)=0$. 
We assume that the design $\xi^*$ consists of support kernel functions $\{K_1,\ldots, K_m\}$ which is a subset of $\mathcal{G}$. 
The corresponding weights are $\{\lambda_1,\ldots, \lambda_m\}$ and for $i=1,\ldots, m$, $0<\lambda_i<1$ if $m\geq 2$. 
Without loss of generality, we only need to show that $\phi(K_1,\xi^*)=0$. 
From Section \ref{sec:learning}, we know that $m$ is finite, but the following proof still applies if $m$ is infinite.

We can write $K(\xi^*)$ as
\[K^*=\lambda_1K_1+\sum_{i>1} \lambda_i K_i=\lambda_1K_1+(1-\lambda_1)K^*_{-1},\]
where $K_{-1}^*=\sum_{i>1}\frac{\lambda_i}{1-\lambda_1} K_i$, and thus $K_{-1}^*$ is a convex combination of $K_2,\ldots, K_m$ and belongs to $\mathcal{K}$. 
Denote the design corresponding to $K_{-1}^*$ by $\xi_{-1}^*$, which has support kernels $\{K_2,\ldots,K_m\}$ and weights $\{\lambda_2/(1-\lambda_1),\ldots, \lambda_m/(1-\lambda_1)\}$. 
Next we prove $\phi(K_1,\xi^*)=0$ by contradiction.

Given (1), we have $\phi(\xi_{-1}^*, \xi^*) \geq 0$. Following Proposition \ref{prop:dir-diff}, 
\begin{align}\nonumber
& \bm y^\top (\bm K^*+\eta \bm I_n)^{-1}(\bm K_{-1}^* - \bm K^*)(\bm K^*+\eta \bm I_n)^{-1}\bm y\leq 0\\\label{eq:neq1}
\Longleftrightarrow \quad & \bm y^\top (\bm K^*+\eta \bm I_n)^{-1}\bm K_{-1}^* (\bm K^*+\eta \bm I_n)^{-1}\bm y \leq  \bm y^\top (\bm K^*+\eta \bm I_n)^{-1} \bm K^*(\bm K^*+\eta \bm I_n)^{-1}\bm y.
\end{align}
Also, $\bm K_1$ is a positive definite matrix and thus $\bm y^\top (\bm K^*+\eta \bm I_n)^{-1}\bm K_1(\bm K^*+\eta \bm I_n)^{-1}\bm y\geq 0$.

If $\phi(K_1,\xi^*) >0$ instead of $\phi(K_1,\xi^*)=0$, then 
\begin{equation}\label{eq:neq2}
\bm y^\top (\bm K^*+\eta \bm I_n)^{-1}\bm K_1(\bm K^*+\eta \bm I_n)^{-1}\bm y < \bm y^\top (\bm K^*+\eta \bm I_n)^{-1}\bm K^*(\bm K^*+\eta \bm I_n)^{-1}\bm y.
\end{equation}
From the two inequalities \eqref{eq:neq1} and \eqref{eq:neq2},
\begin{align*}
&\bm y^\top (\bm K^*+\eta \bm I_n)^{-1}\bm K^* (\bm K^*+\eta \bm I_n)^{-1}\bm y\\
=&\bm y^\top (\bm K^*+\eta \bm I_n)^{-1}(\lambda_1 \bm K_1+(1-\lambda_1)\bm K_{-1}^*)(\bm K^*+\eta \bm I_n)^{-1}\bm y\\
=&\lambda_1 \bm y^\top (\bm K^*+\eta \bm I_n)^{-1}\bm K_1(\bm K^*+\eta \bm I_n)^{-1}\bm y+(1-\lambda_1)\bm y^\top (\bm K^* \\
 &+\eta \bm I_n)^{-1}\bm K_{-1}^*(\bm K^*+\eta \bm I_n)^{-1}\bm y \\
< & \lambda_1 \bm y^\top (\bm K^*+\eta \bm I_n)^{-1}\bm K^*(\bm K^*+\eta \bm I_n)^{-1}\bm y \\
 & +(1-\lambda_1)\bm y^\top (\bm K^*+\eta \bm I_n)^{-1}\bm K^* (\bm K^*+\eta \bm I_n)^{-1}\bm y\\
=& \bm y^\top (\bm K^*+\eta \bm I_n)^{-1}\bm K^* (\bm K^*+\eta \bm I_n)^{-1}\bm y,
\end{align*}
which is a contradiction. 
Thus, it is only possible that $\phi(K_1,\xi^*)=0$.

\item (3)$\to$(2): if $\phi(G,\xi^*)\geq 0$ holds for any $G\in \mathcal{G}$, then
\begin{align*}
\phi(G, \xi^*)=-\eta \bm y^\top ((\bm K^*+\eta \bm I_n)^{-1}(\bm G-\bm K^*)(\bm K^*+\eta \bm I_n)^{-1})\bm y \geq 0.
\end{align*}
Let $\xi'$ be any design in $\Xi$. 
The corresponding kernel matrix $\bm K_{\xi'}=\int \bm G\xi'(\dd \bm G)$ because 
$K(\xi')=\int G\xi'(\dd G)$ and $\bm G$ is the kernel matrix of the kernel function $G$ evaluated at $\mathcal{X}$. 
Therefore, 
\begin{align*}
\phi(\xi',\xi^*)&=-\eta \bm y^\top ((\bm K^*+\eta \bm I_n)^{-1}(\bm K_{\xi'}-\bm K^*)(\bm K^*+\eta \bm I_n)^{-1})\bm y \\
&=-\eta \bm y^\top ((\bm K^*+\eta \bm I_n)^{-1}(\int \bm G\xi'(\dd \bm G)-\bm K^*)(\bm K^*+\eta \bm I_n)^{-1})\bm y\\
&=\int -\eta \bm y^\top ((\bm K^*+\eta \bm I_n)^{-1}(\bm G-\bm K^*)(\bm K^*+\eta \bm I_n)^{-1})\bm y \xi'(\dd \bm G)\geq 0.
\end{align*}
\end{enumerate}
\end{proof}

\subsection{Proof of Theorem \ref{thm:convergence}}

\noindent{\bf Proof of Lemma \ref{lem:inequality}}
\begin{proof}
Denote $K$ as the solution $K=\argmin_{G\in \mathcal{G}}\phi(G,\xi)$. 
Thus, for any $G\in \mathcal{G}$, $\phi(K,\xi)\leq \phi(G,\xi)$. 
The optimal design $\xi^*$ has the support kernel set $\mathcal{S}=\{K_1,\ldots, K_m\}$ and the corresponding optimal weights $\bm \lambda^*=[\lambda^*_1,\ldots, \lambda_m^*]^\top$ such that $0<\lambda_i^*\leq 1$ and $\sum_{i=1}^m \lambda_i^*=1$. 
Based on the definition of $K$, we also have $\phi(K,\xi)\leq \phi(K_i,\xi)$ for any $K_i\in \mathcal{S}$. 
From the formula \eqref{eq:dir-diff} to compute directional derivative, it is straightforward to obtain $\phi(\xi^*, \xi)=\sum_{i=1}^m \lambda_{i}^*\phi(K_i, \xi)$. 
Therefore, 
\[
\phi(K,\xi)=\sum_{i=1}^m \lambda_i^* \phi(K,\xi)\leq \sum_{i=1}^m \lambda_i^* \phi(K_i,\xi)=\phi(\xi^*,\xi).
\]
Furthermore, based on the definition of directional derivative in Definition \ref{def:dir-diff}, 
\begin{align*}
\phi(\xi^*,\xi)&=\lim_{\alpha \to 0^+} \frac{1}{\alpha}\left[Q_{\eta}((1-\alpha)\xi+\alpha \xi^*)-Q_{\eta}(\xi)\right]\\
&\leq \lim_{\alpha \to 0^+} \frac{1}{\alpha} \left[(1-\alpha)Q_{\eta}(\xi)+\alpha Q_{\eta}(\xi^*)-Q_{\eta}(\xi)\right] \\
&=Q_{\eta}(\xi^*)-Q_{\eta}(\xi).
\end{align*}
The above inequality is due to the convexity of $Q_{\eta}(\xi)$ in Lemma \ref{lemma:convex}.
Since $\xi^*$ is the optimal design minimizing $Q_{\eta}$, $Q_{\eta}(\xi^*)-Q_{\eta}(\xi)\leq 0$. 
\end{proof}

\noindent{\bf Proof of Theorem \ref{thm:convergence}}
\begin{proof}
By Theorem \ref{thm:alg3}, the multiplicative procedure converges to the global minimum of each restricted weight problem. 
In the idealized algorithm considered here, this minimum is attained at every outer iteration.

For any iteration $r+1\geq1$, since $\mathcal S^r\subseteq\mathcal S^{r+1}$ and the weight optimization is exact, the design $\xi^{r+1}$ cannot be worse than the previous design $\xi^r$, i.e.,
\[
Q_\eta(\xi^{r+1})\leq Q_\eta(\xi^r).
\]
Since $Q_\eta$ is nonnegative, the objective values have a finite limit.

We prove the result by contradiction.
Suppose that
\[
\lim_{r\to\infty}Q_\eta(\xi^r)>Q_\eta(\xi^*).
\]
Then there exists $c>0$ such that, for all $r\geq0$,
\[
Q_\eta(\xi^r)>Q_\eta(\xi^*)+c.
\]
Recall from Algorithm \ref{alg:fed-wynn} that
\[
K_{r+1}\in
\argmin_{G\in\mathcal G\setminus\mathcal S^r}
\phi(G,\xi^r).
\]
Exact weight optimization implies $\phi(G,\xi^r)\geq0$ for every $G\in\mathcal S^r$.
By Lemma \ref{lem:inequality}, the minimum derivative over $\mathcal G$ is negative under the contradiction hypothesis and therefore is attained outside $\mathcal S^r$. 
Consequently,
\[
\phi(K_{r+1},\xi^r) \leq\phi(\xi^*,\xi^r) \leq Q_\eta(\xi^*)-Q_\eta(\xi^r) <-c.
\]

We next obtain a uniform bound on the second-order directional derivative.
By compactness and continuity of kernel evaluation,
\[
L=\max_{G\in\mathcal G}\|\bm K_G\|_2<\infty.
\]
Put
\[
\bm H_r=\bm K_{K_{r+1}}-\bm K_{\xi^r}, \qquad \bm A_t=\bm K_{\xi^r}+t\bm H_r+\eta\bm I_n, \qquad 0\leq t\leq1.
\]
Since $\|\bm H_r\|_2\leq2L$ and $\bm A_t\succeq\eta\bm I_n$,
\[
0\leq 2\eta\mathbf y^\top \bm A_t^{-1}\bm H_r\bm A_t^{-1} \bm H_r\bm A_t^{-1}\mathbf y \leq \frac{8L^2\|\mathbf y\|^2}{\eta^2} <U,
\]
where
\[
U=1+\frac{8L^2\|\mathbf y\|^2}{\eta^2}
\]
is independent of $r$ and $t$.

Here we slightly abuse the notation and identify $K_{r+1}$ with the design assigning unit probability to that kernel.
For any $0<\alpha\leq1$, Taylor's theorem gives
\begin{align*}
Q_\eta((1-\alpha)\xi^r+\alpha K_{r+1})
&= Q_\eta(\xi^r) +\phi(K_{r+1},\xi^r)\alpha +\frac{u}{2}\alpha^2\\ 
&< Q_\eta(\xi^r)-c\alpha+\frac{U}{2}\alpha^2,
\end{align*}
where $0\leq u\leq U$.

Since the weights on the enlarged support set are optimized exactly,
\begin{align*}
Q_\eta(\xi^{r+1}) &\leq Q_\eta((1-\alpha)\xi^r+\alpha K_{r+1})\\
&< Q_\eta(\xi^r)-c\alpha+\frac{U}{2}\alpha^2.
\end{align*}
Choose the fixed step
\[
\alpha_0=\min\{1,c/U\}>0.
\]
Then $U\alpha_0\leq c$, and hence
\[
Q_\eta(\xi^{r+1})-Q_\eta(\xi^r) < -\frac{c\alpha_0}{2}<0 \qquad\text{for every }r\geq0.
\]
Summing these inequalities gives
\[
Q_\eta(\xi^r) \leq Q_\eta(\xi^0)-\frac{rc\alpha_0}{2},
\]
which contradicts the nonnegativity of $Q_\eta$.
Therefore,
\[
\lim_{r\to\infty}Q_\eta(\xi^r)=Q_\eta(\xi^*).
\]
\end{proof}

\section{Discussions on Kernel Parameters}\label{Ssec:kernelpara}
In this section, we provide more in-depth discussion on the choice of the parameters of the kernel function $\theta$, also known as the inverse length-scale parameter. 
First, we use a small-scale experiment to compare the isotropic and anisotropic kernel. 
Then, we look into how sensitive \texttt{optK} is to the discrete grid set $\Theta$ for $\theta$ when constructing the basic kernels. 

\subsection{Isotropic v.s. Anisotropic Kernel Functions}\label{Ssecsub:isotropic}
Independent coordinate scales can be accommodated by expanding the basic-kernel dictionary:
\[
K_{\theta_j,\theta_k}(\mathbf{x},\mathbf{x}') =\exp\!\left[-\theta_j(x_j-x'_j)^2-\theta_k(x_k-x'_k)^2\right],
\]
with isotropy corresponding to $\theta_j=\theta_k$. 
The existing optimization of mixture weights $\lambda$ is retained. 
With 32 candidate scales per coordinate, full enumeration gives 32 isotropic versus $32^2=1024$ anisotropic candidates per eligible variable pair $(x_j, x_k)$. 
These counts refer to our dictionary implementation.

We conducted an illustrative comparison for the Borehole example at $n=200$ and $500$, $d=20$, with one dataset per sample size ($B=1$) and deletion thresholds
$\texttt{DEL}=0.005$ and $0.05$. 
For \texttt{optK} using isotropic and anisotropic kernels, we use $\texttt{MaxDim}=2$, limiting each basic kernel to two inputs at most. 
The Borehole function has $p=8$ active inputs among the $d=20$ inputs.
For each combination of $n$ and $\texttt{DEL}$, data, initialization and optimization settings are shared between the isotropic and anisotropic runs.
Both methods use strong heredity and no terminal pruning.
All eight workflows completed on the same Macbook Air equipped with an Apple M2 chip and $16$ GB of memory, using one worker thread.
The following table shows the comparison results. 

\begin{center}
\centering
\small
\begin{tabular}{@{}rr r r r r r@{}}
\hline
\texttt{DEL} & $n$ & \multicolumn{2}{c}{SRMSE} &
\multicolumn{2}{c}{Wall-clock time (s)} & Time ratio\\
 & & ISO & ANISO & ISO & ANISO & ANISO/ISO\\
\hline
0.005 & 200 & 0.061042 & 0.060868 & 4.609 & 174.613 & 37.9\\
0.005 & 500 & 0.046912 & 0.046590 & 55.467 & 1,291.041 & 23.3\\
0.05 & 200 & 0.056005 & 0.049679 & 1.150 & 31.945 & 27.8\\
0.05 & 500 & 0.050027 & 0.049925 & 4.371 & 66.273 & 15.2\\
\hline
\end{tabular}
\end{center}

For each setting, we only run $B=1$ simulation, and thus SD is unavailable. 
Wall clock time reports the elapsed time for the training procedure for the chosen optimal $\eta$ value.

Anisotropic kernels reduced SRMSE in all four settings, but the magnitude varied: reductions were below 0.7\% in three settings and approximately 11.3\% at $n=200$, $\texttt{DEL}=0.05$. 
The fitting time took approximately 15--38 times as long with anisotropic kernels.


These observations support isotropic blocks as a computationally economical default for the simulation setting, which uses $d=20$ and $\texttt{MaxDim}=2$, without establishing a general advantage over anisotropic kernels.
Under full grid enumeration, extending independent coordinate scales to a
$k$-variable block would require $32^k$ candidates, compared with $32$
under isotropy.
This exponential growth with block order, together with more eligible
blocks as $d$ increases, can substantially increase computation and
storage demands.
Anisotropic kernels may still be useful when they substantially improve prediction.

\subsection{Discrete Grid Set for Kernel Parameter}\label{ssubsec:Theta}

For the Gaussian kernel, larger values of $\theta$ correspond to shorter length-scales.
In this part, we investigate how \texttt{optK} is affected by the range, resolution, and size of the grid $\Theta$. 

We examined three independent paired datasets for the positive $p=6$ Michalewicz function in \eqref{eq:michal-new} with $a=(1,2,3,4,5,6)$, $n=500$ and $d=20$. 
Larger values of $a_j$ make the corresponding component oscillate more rapidly, so the response can vary sharply over short intervals of $x_j$.
All comparisons use $\texttt{DEL}=0.05$ and $\texttt{MaxDim}=4$. 

We compare four grids.
G0 and G1 use the following grids, containing $25$ and $30$ values, respectively:
\begin{equation*}
\begin{aligned}
\Theta_0 &=\left\{c\times 10^b: c\in\{1,3,5,7,9\},\ b\in\{-2,-1,0,1,2\}\right\},\\
\Theta_1 &=\Theta_0\cup\{1000,3000,5000,7000,9000\}.
\end{aligned}
\end{equation*}
Thus, G1 extends G0's upper bound from 900 to 9000.
The 32-value dictionary used in Section~\ref{sec:num} is $\Theta_1\cup\{10000,30000\}$.
G2 and G3 add all adjacent geometric midpoints to $\Theta_0$ and $\Theta_1$, respectively, giving 49 and 59 grid points.
These comparisons assess whether adding more grid points improves prediction without changing the range of $\theta$.

\begin{center}
\begin{minipage}{\linewidth}
\centering
\begingroup
\small
\setlength{\tabcolsep}{5pt}
\begin{tabular}{lrrrr}
\hline
Dictionary & Points & Maximum $\theta$ & SRMSE & Wall-clock time (s)\\
\hline
G0: original & 25 & 900 & 0.337461 & 7.33\\
G1: extended & 30 & 9000 & 0.118538 & 8.49\\
G2: G0 dense & 49 & 900 & 0.339794 & 12.20\\
G3: G1 dense & 59 & 9000 & 0.118871 & 14.45\\
\hline
\end{tabular}
\par\smallskip
\begin{minipage}{0.98\linewidth}
\footnotesize\raggedright
SRMSE and Wall-clock time are means across three datasets ($B=3$).
\end{minipage}
\endgroup
\end{minipage}
\end{center}

Extending the upper bound from 900 to 9000 (G0 to G1) improved prediction in all three datasets, reducing mean SRMSE by 64.9\% with a 15.7\% increase in mean wall-clock time.
Adding geometric midpoints within either range (G0 to G2 and G1 to G3) did not consistently improve prediction and increased computational cost.
Additional Michalewicz and Borehole experiments also showed no consistent benefit from further increasing the upper bound.
These results suggest that expanding the grid range can be more beneficial than adding points within a fixed range, especially when the response varies rapidly over short input intervals.
However, further increasing the upper bound may offer little additional benefit if the response does not vary substantially at the corresponding finer scales.

In addition, we examined three strategies for searching and refining the candidate dictionary.
Full-dictionary search (F) accesses all candidates in the 32-value dictionary used in Section~\ref{sec:num}.
Restricted search (S), after an unrestricted second selection, permits only additions with $\theta$ strictly smaller than that of the last successfully added kernel.
Adaptive refinement (A) first runs F at each $\eta$, evaluates geometric midpoints between adjacent grid values, and adds up to four candidates with the most negative directional derivatives.
If candidates are added, the model is refitted using the expanded dictionary and the original initialization.

The experiments used noiseless one-dimensional GPs whose covariance functions combine two or four Gaussian kernel components with distinct values of the kernel parameter $\theta$ ($n=80$).
Results were averaged over three initializations within each of 20 paired replications.
In the on-grid simulations, all generating $\theta$ values were included in this 32-value dictionary.
Mean SRMSE increased from $0.00509$ for F to $0.02472$ for S in the two-component setting, and from $0.00638$ to $0.02305$ in the four-component setting.
A did not consistently improve prediction over F.

Under the same setup, off-grid simulations placed the generating $\theta$ values at geometric midpoints or one-third positions between adjacent grid points on the log scale.
We also applied full-dictionary search after adding either all geometric midpoints or the true generating $\theta$ values to the original dictionary, with the latter serving as a diagnostic reference.
Across the off-grid settings, the two dictionary extensions and A produced only small changes in mean SRMSE, with no consistent improvement over F using the original dictionary.
These small changes may partly reflect the ability of kernel mixtures to approximate kernels with off-grid $\theta$ values without exactly recovering the generating covariance.
Overall, these results favor searching the full dictionary without imposing a decreasing order on $\theta$.
Adaptive boundary extensions and multivariate refinement remain future work.

\section{Additional Details and Results for the Examples}\label{Ssec:extrasimulation}
This section provides implementation settings and additional results organized by example.
Eleven configurations are tabulated only here: four six-variable Michalewicz cases with $d=10,20$, four Borehole cases with $d=20,40$, and three Satellite responses at $n=200$. Supplementary plots show only configurations not plotted in the main text.

\subsection{Implementation and Data Details}\label{ssec:s3-implementation}
The experiments used R 4.5.3, Python 3.11.16, \texttt{mlegp} 3.1.9, \texttt{laGP} 1.5-9, \texttt{MRFA} 0.6, \texttt{BART} 2.9.10, PyTorch 2.8.0 (CPU-only build), GPyTorch 1.15.2 and Pymanopt 2.2.1. 
Test responses are used only to evaluate predictive performance.

\textbf{optK.} The algorithm is initialized separately for each candidate $\eta$, with no warm starts across candidates.
For \texttt{optK-pruned}, any weight optimization after terminal pruning starts from uniform weights and precedes the leave-one-out cross validation evaluation. 
Selected variables are the union of inputs in the selected model's nonzero support kernels.

\textbf{mlegp and laGP.} The mlegp call uses a constant mean, a separable Gaussian correlation function and an estimated nugget (initial value 1 and lower bound 0).
Default optimization settings are used.
The strict binary variable selection rule is $\beta_j>0.01$ where $\beta_j$ is the inverse length-scale of the $j$th input. 
For laGP, we fit a GP with a separable Gaussian kernel.
Length scales and the nugget are jointly estimated by maximum likelihood, using package-provided initial values and bounds and a maximum $1000$ iterations.
Responses are standardized using the training mean and sample SD, and predictions are transformed back to the original scale.

\textbf{MRFA and DART.} MRFA uses the minimum-penalty endpoint without additional cross-validation, selecting inputs in the union of aggregated nonzero basis supports.
DART uses 200 trees and two chains, each with 2000 burn-in and 8000 retained draws without thinning.
Within each chain, PIP is the fraction of retained draws containing an input.
PIPs and predictions are averaged equally across chains. 
Inputs with PIP $>0.5$ are selected.

\textbf{Variational GPs.} SVGP-ARD uses a constant mean, a scaled ARD Gaussian kernel and a Gaussian likelihood.
DGP-2 has a five-output hidden GP layer with linear means and a scalar output GP layer with a constant mean.
Both layers use scaled ARD Gaussian kernels, and the output layer receives only the hidden layer outputs as inputs.
Both methods use up to $64$ inducing points, with locations optimized during training.
Inputs and responses are standardized within each training subset.
Predictions are restored to the original scale.

Using an $80/20$ training-validation split, three initializations are trained for $4000$ full-batch Adam epochs.
The epoch minimizing validation MSE is recorded for each initialization.
We then refit the model on all training data from three new initializations, each trained for the median selected number of epochs, and retain the valid fit with the smallest negative ELBO.
DGP-2 uses $10$ Monte Carlo samples for training and $128$ for validation, final model selection and prediction.

\textbf{AS-GP.} An orthonormal projection and Gaussian GP hyperparameters are jointly estimated by minimizing the training negative log likelihood (NLL).
Three-fold cross-validation MSE selects the projection dimension $r\in\{1,\ldots,\min(8,d)\}$.
Each fold uses three initializations, and the selected dimension is refitted on all training data from five initializations.
In both stages, the valid fit with the lowest training NLL is retained.
Optimization uses Pymanopt conjugate gradients, with at most 3000 iterations and gradient tolerance $10^{-6}$.
Standardization is performed separately within each fitting subset.

For laGP, SVGP-ARD and DGP-2, no threshold is applied to the fitted length scales to select variables in this study.
AS-GP estimates a projection subspace rather than an input subset. 
FN/FP are therefore reported as NA for these four implementations.

\textbf{Borehole inputs.} The physical input definitions and ranges for the Borehole experiments in Section~\ref{sec:num} are given in Table~\ref{table-4}.

\begin{table}[H]\centering
\caption{Input variables and their ranges for the Borehole function. The row order is the order of columns 1--8 in the benchmark.}\label{table-4}
\begin{tabular}{cllc}\toprule
Symbol & Parameter & Range & Unit\\\midrule
$r_w$ & radius of borehole & $[0.05,0.15]$ & m\\
$r$ & radius of influence & $[100,50000]$ & m\\
$T_u$ & transmissivity of upper aquifer & $[63070,115600]$ & $\mathrm{m}^2/\mathrm{yr}$\\
$H_u$ & potentiometric head of upper aquifer & $[990,1110]$ & m\\
$T_l$ & transmissivity of lower aquifer & $[63.1,116]$ & $\mathrm{m}^2/\mathrm{yr}$\\
$H_l$ & potentiometric head of lower aquifer & $[700,820]$ & m\\
$L$ & length of borehole & $[1120,1680]$ & m\\
$K_w$ & hydraulic conductivity of borehole & $[9855,12045]$ & $\mathrm{m}/\mathrm{yr}$\\\bottomrule
\end{tabular}
\end{table}

\textbf{Satellite data source and version.} We use the GRACE pilot-study data accompanying Section~2.3.2 of \citet{gramacy2020surrogates}, associated with the LANL study of \citet{mehta2014modeling}. 
The training pool and test set come from \texttt{CD\_GRACE\_1000\_*.dat} and \texttt{CD\_GRACE\_100\_*.dat}, respectively, where \texttt{*} denotes \texttt{He}, \texttt{N2} or \texttt{O2}. 
We use the legacy simulator responses supplied in these \texttt{.dat} files.

Independent uniform inputs are added to the training and test data as irrelevant variables. 
Min--max scaling is applied first to the physical inputs and then to all inputs after adding the irrelevant variables. 
At each fitting stage, scaling parameters are estimated from the training data only.

\textbf{Computational resources.} For each candidate $\eta$, optK uses up to $C$ workers, each using one CPU core; search decisions and weight updates remain sequential.
mlegp and laGP use one worker regardless of $C$, while MRFA uses up to $C$ workers for parts of fitting and prediction.
DART runs its two chains sequentially at $C=1$ and in parallel at $C=2,4$.
SVGP-ARD and DGP-2 run initializations sequentially in one worker configured to use up to $C$ CPU cores.
AS-GP runs independent fits in parallel using up to $C$ workers, each using one CPU core.
We report the wall-clock training time for all methods.

\subsection{Additional Michalewicz Example}\label{sec:s3-michal}
Table~\ref{tab:s-variant} reports results for the six-variable Michalewicz function with $\bm a=(1,2,1,2,1,2)$ at $d=10,20$ and $n=300,500$. 
Figures~\ref{fig:s-case01}, \ref{fig:s-case02} and~\ref{fig:s-case26} show the two-variable cases at $n=200,500$ and the six-variable case at $d=60,n=300$, respectively. 
The larger-sample cases are plotted in Figures~\ref{fig:main-michal2-n1000} and~\ref{fig:main-michal6-n500} of the main text.
Boxplot conventions follow Figure~\ref{fig:main-michal2-n1000}.
\par\begingroup\spacingset{1}\footnotesize
\setlength{\tabcolsep}{2pt}\renewcommand{\arraystretch}{1}
\def\ptmeasure#1#2{\settowidth{\dimen0}{#2}\ifdim\dimen0>#1\relax\edef#1{\the\dimen0}\fi}
\def\ptMethodWidth{0pt}
\ptmeasure\ptMethodWidth{Method}
\ptmeasure\ptMethodWidth{\texttt{AS-GP}}
\ptmeasure\ptMethodWidth{\texttt{DART}}
\ptmeasure\ptMethodWidth{\texttt{DGP-2}}
\ptmeasure\ptMethodWidth{\texttt{MRFA}}
\ptmeasure\ptMethodWidth{\texttt{SVGP-ARD}}
\ptmeasure\ptMethodWidth{\texttt{laGP}}
\ptmeasure\ptMethodWidth{\texttt{mlegp}}
\ptmeasure\ptMethodWidth{\texttt{optK-pruned}}
\ptmeasure\ptMethodWidth{\texttt{optK}\textsubscript{0.005}}
\ptmeasure\ptMethodWidth{\texttt{optK}\textsubscript{0.05}}
\def\ptSrmseWidth{0pt}
\ptmeasure\ptSrmseWidth{SRMSE}
\ptmeasure\ptSrmseWidth{\mbox{0.0248 (0.00200)}}
\ptmeasure\ptSrmseWidth{\mbox{0.0249 (0.00235)}}
\ptmeasure\ptSrmseWidth{\mbox{0.0249 (0.00289)}}
\ptmeasure\ptSrmseWidth{\mbox{0.0250 (0.00265)}}
\ptmeasure\ptSrmseWidth{\mbox{0.0301 (0.00359)}}
\ptmeasure\ptSrmseWidth{\mbox{0.0325 (0.00416)}}
\ptmeasure\ptSrmseWidth{\mbox{0.0587 (0.0131)}}
\ptmeasure\ptSrmseWidth{\mbox{0.0667 (0.0179)}}
\ptmeasure\ptSrmseWidth{\mbox{0.0717 (0.00855)}}
\ptmeasure\ptSrmseWidth{\mbox{0.0720 (0.0155)}}
\ptmeasure\ptSrmseWidth{\mbox{0.0775 (0.0176)}}
\ptmeasure\ptSrmseWidth{\mbox{0.0792 (0.0106)}}
\ptmeasure\ptSrmseWidth{\mbox{0.0815 (0.0155)}}
\ptmeasure\ptSrmseWidth{\mbox{0.0840 (0.00322)}}
\ptmeasure\ptSrmseWidth{\mbox{0.0841 (0.00273)}}
\ptmeasure\ptSrmseWidth{\mbox{0.0930 (0.0740)}}
\ptmeasure\ptSrmseWidth{\mbox{0.103 (0.00546)}}
\ptmeasure\ptSrmseWidth{\mbox{0.104 (0.00516)}}
\ptmeasure\ptSrmseWidth{\mbox{0.184 (0.0242)}}
\ptmeasure\ptSrmseWidth{\mbox{0.200 (0.0349)}}
\ptmeasure\ptSrmseWidth{\mbox{0.735 (0.242)}}
\ptmeasure\ptSrmseWidth{\mbox{0.887 (0.0162)}}
\ptmeasure\ptSrmseWidth{\mbox{0.891 (0.0147)}}
\ptmeasure\ptSrmseWidth{\mbox{0.925 (0.0213)}}
\ptmeasure\ptSrmseWidth{\mbox{0.929 (0.0171)}}
\ptmeasure\ptSrmseWidth{\mbox{0.929 (0.0192)}}
\ptmeasure\ptSrmseWidth{\mbox{0.942 (0.0199)}}
\ptmeasure\ptSrmseWidth{\mbox{0.952 (0.0142)}}
\ptmeasure\ptSrmseWidth{\mbox{0.959 (0.0234)}}
\ptmeasure\ptSrmseWidth{\mbox{0.967 (0.0263)}}
\ptmeasure\ptSrmseWidth{\mbox{0.968 (0.0173)}}
\ptmeasure\ptSrmseWidth{\mbox{0.975 (0.0177)}}
\ptmeasure\ptSrmseWidth{\mbox{0.990 (0.0215)}}
\ptmeasure\ptSrmseWidth{\mbox{0.997 (0.0551)}}
\ptmeasure\ptSrmseWidth{\mbox{1.01 (0.0638)}}
\ptmeasure\ptSrmseWidth{\mbox{1.03 (0.0457)}}
\ptmeasure\ptSrmseWidth{\mbox{1.03 (0.0734)}}
\ptmeasure\ptSrmseWidth{\mbox{1.16 (0.0566)}}
\ptmeasure\ptSrmseWidth{\mbox{1.21 (0.0338)}}
\ptmeasure\ptSrmseWidth{\mbox{1.28 (0.0592)}}
\def\ptFnWidth{0pt}
\ptmeasure\ptFnWidth{FN}
\ptmeasure\ptFnWidth{\mbox{0 (0)}}
\ptmeasure\ptFnWidth{\mbox{0.05 (0.26)}}
\ptmeasure\ptFnWidth{\mbox{0.14 (0.43)}}
\ptmeasure\ptFnWidth{\mbox{0.76 (0.78)}}
\ptmeasure\ptFnWidth{\mbox{0.77 (0.90)}}
\ptmeasure\ptFnWidth{\mbox{NA}}
\def\ptFpWidth{0pt}
\ptmeasure\ptFpWidth{FP}
\ptmeasure\ptFpWidth{\mbox{0 (0)}}
\ptmeasure\ptFpWidth{\mbox{0.01 (0.10)}}
\ptmeasure\ptFpWidth{\mbox{0.02 (0.14)}}
\ptmeasure\ptFpWidth{\mbox{0.10 (0.39)}}
\ptmeasure\ptFpWidth{\mbox{0.12 (0.33)}}
\ptmeasure\ptFpWidth{\mbox{0.22 (0.60)}}
\ptmeasure\ptFpWidth{\mbox{1.51 (1.00)}}
\ptmeasure\ptFpWidth{\mbox{1.74 (0.97)}}
\ptmeasure\ptFpWidth{\mbox{12.34 (1.84)}}
\ptmeasure\ptFpWidth{\mbox{12.97 (1.94)}}
\ptmeasure\ptFpWidth{\mbox{2.59 (1.06)}}
\ptmeasure\ptFpWidth{\mbox{2.85 (1.03)}}
\ptmeasure\ptFpWidth{\mbox{4.31 (1.52)}}
\ptmeasure\ptFpWidth{\mbox{5.06 (1.64)}}
\ptmeasure\ptFpWidth{\mbox{NA}}
\def\ptIterWidth{0pt}
\ptmeasure\ptIterWidth{Iter}
\ptmeasure\ptIterWidth{\mbox{---}}
\ptmeasure\ptIterWidth{\mbox{15.4 (0.9)}}
\ptmeasure\ptIterWidth{\mbox{15.5 (0.9)}}
\ptmeasure\ptIterWidth{\mbox{15.7 (0.7)}}
\ptmeasure\ptIterWidth{\mbox{15.7 (1.3)}}
\ptmeasure\ptIterWidth{\mbox{16.5 (1.6)}}
\ptmeasure\ptIterWidth{\mbox{16.7 (1.6)}}
\ptmeasure\ptIterWidth{\mbox{17.0 (1.2)}}
\ptmeasure\ptIterWidth{\mbox{17.2 (1.5)}}
\ptmeasure\ptIterWidth{\mbox{17.4 (1.1)}}
\ptmeasure\ptIterWidth{\mbox{17.4 (2.4)}}
\ptmeasure\ptIterWidth{\mbox{17.8 (2.1)}}
\def\ptTimeWidth{0pt}
\ptmeasure\ptTimeWidth{\makecell[r]{1-core\\(Sec)}}
\ptmeasure\ptTimeWidth{\makecell[r]{2-core\\(Sec)}}
\ptmeasure\ptTimeWidth{\makecell[r]{4-core\\(Sec)}}
\ptmeasure\ptTimeWidth{\mbox{10.3 (2.78)}}
\ptmeasure\ptTimeWidth{\mbox{10.5 (4.75)}}
\ptmeasure\ptTimeWidth{\mbox{10.8 (3.08)}}
\ptmeasure\ptTimeWidth{\mbox{105 (13.3)}}
\ptmeasure\ptTimeWidth{\mbox{11.0 (5.20)}}
\ptmeasure\ptTimeWidth{\mbox{11.2 (4.77)}}
\ptmeasure\ptTimeWidth{\mbox{11.3 (3.48)}}
\ptmeasure\ptTimeWidth{\mbox{11.8 (3.78)}}
\ptmeasure\ptTimeWidth{\mbox{110 (17.5)}}
\ptmeasure\ptTimeWidth{\mbox{115 (10.7)}}
\ptmeasure\ptTimeWidth{\mbox{116 (17.0)}}
\ptmeasure\ptTimeWidth{\mbox{118 (16.2)}}
\ptmeasure\ptTimeWidth{\mbox{11800 (1880)}}
\ptmeasure\ptTimeWidth{\mbox{12.1 (3.31)}}
\ptmeasure\ptTimeWidth{\mbox{12.2 (3.25)}}
\ptmeasure\ptTimeWidth{\mbox{12.7 (3.60)}}
\ptmeasure\ptTimeWidth{\mbox{12.8 (3.08)}}
\ptmeasure\ptTimeWidth{\mbox{12.8 (4.53)}}
\ptmeasure\ptTimeWidth{\mbox{120 (17.4)}}
\ptmeasure\ptTimeWidth{\mbox{120 (9.51)}}
\ptmeasure\ptTimeWidth{\mbox{123 (18.6)}}
\ptmeasure\ptTimeWidth{\mbox{123 (41.7)}}
\ptmeasure\ptTimeWidth{\mbox{1230 (404)}}
\ptmeasure\ptTimeWidth{\mbox{12400 (1750)}}
\ptmeasure\ptTimeWidth{\mbox{127 (14.5)}}
\ptmeasure\ptTimeWidth{\mbox{13.1 (3.08)}}
\ptmeasure\ptTimeWidth{\mbox{13.8 (2.76)}}
\ptmeasure\ptTimeWidth{\mbox{130 (46.2)}}
\ptmeasure\ptTimeWidth{\mbox{131 (10.5)}}
\ptmeasure\ptTimeWidth{\mbox{131 (12.2)}}
\ptmeasure\ptTimeWidth{\mbox{131 (13.8)}}
\ptmeasure\ptTimeWidth{\mbox{133 (17.4)}}
\ptmeasure\ptTimeWidth{\mbox{133 (20.4)}}
\ptmeasure\ptTimeWidth{\mbox{134 (51.4)}}
\ptmeasure\ptTimeWidth{\mbox{135 (26.2)}}
\ptmeasure\ptTimeWidth{\mbox{14.1 (3.80)}}
\ptmeasure\ptTimeWidth{\mbox{144 (19.0)}}
\ptmeasure\ptTimeWidth{\mbox{144 (20.0)}}
\ptmeasure\ptTimeWidth{\mbox{144 (47.1)}}
\ptmeasure\ptTimeWidth{\mbox{145 (14.4)}}
\ptmeasure\ptTimeWidth{\mbox{1460 (239)}}
\ptmeasure\ptTimeWidth{\mbox{147 (49.0)}}
\ptmeasure\ptTimeWidth{\mbox{147 (49.7)}}
\ptmeasure\ptTimeWidth{\mbox{1480 (255)}}
\ptmeasure\ptTimeWidth{\mbox{151 (49.9)}}
\ptmeasure\ptTimeWidth{\mbox{154 (51.7)}}
\ptmeasure\ptTimeWidth{\mbox{16.5 (3.16)}}
\ptmeasure\ptTimeWidth{\mbox{18.3 (3.53)}}
\ptmeasure\ptTimeWidth{\mbox{18.7 (6.92)}}
\ptmeasure\ptTimeWidth{\mbox{19.0 (6.88)}}
\ptmeasure\ptTimeWidth{\mbox{19.1 (3.90)}}
\ptmeasure\ptTimeWidth{\mbox{20.0 (7.28)}}
\ptmeasure\ptTimeWidth{\mbox{20.9 (5.78)}}
\ptmeasure\ptTimeWidth{\mbox{21.3 (8.43)}}
\ptmeasure\ptTimeWidth{\mbox{21.8 (6.03)}}
\ptmeasure\ptTimeWidth{\mbox{211 (376)}}
\ptmeasure\ptTimeWidth{\mbox{22.0 (5.80)}}
\ptmeasure\ptTimeWidth{\mbox{224 (438)}}
\ptmeasure\ptTimeWidth{\mbox{23.0 (9.24)}}
\ptmeasure\ptTimeWidth{\mbox{23.4 (6.25)}}
\ptmeasure\ptTimeWidth{\mbox{23.8 (7.65)}}
\ptmeasure\ptTimeWidth{\mbox{23.9 (4.42)}}
\ptmeasure\ptTimeWidth{\mbox{233 (422)}}
\ptmeasure\ptTimeWidth{\mbox{24.1 (6.51)}}
\ptmeasure\ptTimeWidth{\mbox{24.3 (4.51)}}
\ptmeasure\ptTimeWidth{\mbox{25.0 (5.44)}}
\ptmeasure\ptTimeWidth{\mbox{25.4 (4.31)}}
\ptmeasure\ptTimeWidth{\mbox{25.8 (6.23)}}
\ptmeasure\ptTimeWidth{\mbox{254 (18.4)}}
\ptmeasure\ptTimeWidth{\mbox{255 (17.7)}}
\ptmeasure\ptTimeWidth{\mbox{260 (471)}}
\ptmeasure\ptTimeWidth{\mbox{27.5 (3.55)}}
\ptmeasure\ptTimeWidth{\mbox{27.5 (5.83)}}
\ptmeasure\ptTimeWidth{\mbox{27.9 (5.21)}}
\ptmeasure\ptTimeWidth{\mbox{2730 (438)}}
\ptmeasure\ptTimeWidth{\mbox{2740 (364)}}
\ptmeasure\ptTimeWidth{\mbox{275 (16.4)}}
\ptmeasure\ptTimeWidth{\mbox{278 (22.6)}}
\ptmeasure\ptTimeWidth{\mbox{29.8 (4.43)}}
\ptmeasure\ptTimeWidth{\mbox{31.7 (14.5)}}
\ptmeasure\ptTimeWidth{\mbox{32.4 (14.4)}}
\ptmeasure\ptTimeWidth{\mbox{3380 (692)}}
\ptmeasure\ptTimeWidth{\mbox{34.7 (16.6)}}
\ptmeasure\ptTimeWidth{\mbox{3430 (555)}}
\ptmeasure\ptTimeWidth{\mbox{46.2 (5.68)}}
\ptmeasure\ptTimeWidth{\mbox{49.2 (6.16)}}
\ptmeasure\ptTimeWidth{\mbox{5.63 (1.45)}}
\ptmeasure\ptTimeWidth{\mbox{5410 (895)}}
\ptmeasure\ptTimeWidth{\mbox{544 (707)}}
\ptmeasure\ptTimeWidth{\mbox{547 (93.2)}}
\ptmeasure\ptTimeWidth{\mbox{5480 (779)}}
\ptmeasure\ptTimeWidth{\mbox{560 (196)}}
\ptmeasure\ptTimeWidth{\mbox{568 (320)}}
\ptmeasure\ptTimeWidth{\mbox{571 (59.4)}}
\ptmeasure\ptTimeWidth{\mbox{574 (415)}}
\ptmeasure\ptTimeWidth{\mbox{580 (728)}}
\ptmeasure\ptTimeWidth{\mbox{581 (90.8)}}
\ptmeasure\ptTimeWidth{\mbox{6.35 (2.35)}}
\ptmeasure\ptTimeWidth{\mbox{6.68 (2.62)}}
\ptmeasure\ptTimeWidth{\mbox{6.92 (2.53)}}
\ptmeasure\ptTimeWidth{\mbox{6.97 (2.39)}}
\ptmeasure\ptTimeWidth{\mbox{6020 (969)}}
\ptmeasure\ptTimeWidth{\mbox{609 (245)}}
\ptmeasure\ptTimeWidth{\mbox{614 (71.7)}}
\ptmeasure\ptTimeWidth{\mbox{632 (56.3)}}
\ptmeasure\ptTimeWidth{\mbox{6380 (1160)}}
\ptmeasure\ptTimeWidth{\mbox{646 (106)}}
\ptmeasure\ptTimeWidth{\mbox{67.5 (8.49)}}
\ptmeasure\ptTimeWidth{\mbox{67.9 (8.76)}}
\ptmeasure\ptTimeWidth{\mbox{676 (66.7)}}
\ptmeasure\ptTimeWidth{\mbox{7.08 (2.76)}}
\ptmeasure\ptTimeWidth{\mbox{7.87 (1.40)}}
\ptmeasure\ptTimeWidth{\mbox{704 (978)}}
\ptmeasure\ptTimeWidth{\mbox{75.0 (11.4)}}
\ptmeasure\ptTimeWidth{\mbox{76.9 (12.1)}}
\ptmeasure\ptTimeWidth{\mbox{761 (383)}}
\ptmeasure\ptTimeWidth{\mbox{8.64 (1.96)}}
\ptmeasure\ptTimeWidth{\mbox{8.99 (2.18)}}
\ptmeasure\ptTimeWidth{\mbox{824 (467)}}
\ptmeasure\ptTimeWidth{\mbox{849 (195)}}
\def\ptUpdateWidth{\edef\ptTableWidth{\the\dimexpr\ptMethodWidth+\ptSrmseWidth+\ptFnWidth+\ptFpWidth+\ptIterWidth+\ptTimeWidth+\ptTimeWidth+\ptTimeWidth+14\tabcolsep\relax}}
\ptUpdateWidth\ifdim\ptTableWidth>\linewidth\setlength{\tabcolsep}{1pt}\ptUpdateWidth\fi
\typeout{PERFORMANCE-WIDTH supp_variant: table=\ptTableWidth; time=\ptTimeWidth; linewidth=\the\linewidth}
\ifdim\ptTableWidth>\linewidth\PackageError{performance-tables}{Table exceeds linewidth}{Check the table layout.}\fi
\setlength{\LTcapwidth}{\ptTableWidth}
\Needspace{570pt}
\begin{longtable}{@{}>{\raggedright\arraybackslash}p{\ptMethodWidth}>{\raggedleft\arraybackslash}p{\ptSrmseWidth}>{\raggedleft\arraybackslash}p{\ptFnWidth}>{\raggedleft\arraybackslash}p{\ptFpWidth}>{\raggedleft\arraybackslash}p{\ptIterWidth}>{\raggedleft\arraybackslash}p{\ptTimeWidth}>{\raggedleft\arraybackslash}p{\ptTimeWidth}>{\raggedleft\arraybackslash}p{\ptTimeWidth}@{}}
\caption{Six-variable Michalewicz function, $\bm a=(1,2,1,2,1,2)$; $d=10,20$, $p=6$, $m=3481$, $B=100$.}\label{tab:s-variant}\\
\toprule
Method & SRMSE & FN & FP & Iter & \makecell[r]{1-core\\(Sec)} & \makecell[r]{2-core\\(Sec)} & \makecell[r]{4-core\\(Sec)} \\
\midrule\endfirsthead
\caption[]{Six-variable Michalewicz function, $\bm a=(1,2,1,2,1,2)$; $d=10,20$, $p=6$, $m=3481$, $B=100$. (continued).}\\\toprule
Method & SRMSE & FN & FP & Iter & \makecell[r]{1-core\\(Sec)} & \makecell[r]{2-core\\(Sec)} & \makecell[r]{4-core\\(Sec)} \\
\midrule\endhead
\midrule
\multicolumn{8}{r@{}}{\footnotesize Continued on next page.}\\\endfoot
\bottomrule
\endlastfoot
\multicolumn{8}{@{}l}{\textbf{$n=300$, $d=10$}}\\*
\texttt{optK}\textsubscript{0.005} & \mbox{0.0775 (0.0176)} & \mbox{0 (0)} & \mbox{0.22 (0.60)} & \mbox{17.4 (2.4)} & \mbox{6.35 (2.35)} & \mbox{11.3 (3.48)} & \mbox{11.8 (3.78)} \\*
\texttt{optK-pruned} & \mbox{0.0667 (0.0179)} & \mbox{0 (0)} & \mbox{0.02 (0.14)} & \mbox{17.4 (2.4)} & \mbox{6.97 (2.39)} & \mbox{12.1 (3.31)} & \mbox{12.8 (4.53)} \\*
\texttt{optK}\textsubscript{0.05} & \mbox{0.0587 (0.0131)} & \mbox{0 (0)} & \mbox{0 (0)} & \mbox{15.7 (1.3)} & \mbox{5.63 (1.45)} & \mbox{10.3 (2.78)} & \mbox{10.8 (3.08)} \\*
\texttt{laGP} & \mbox{0.929 (0.0192)} & \mbox{NA} & \mbox{NA} & \mbox{---} & \mbox{7.08 (2.76)} & \mbox{6.92 (2.53)} & \mbox{6.68 (2.62)} \\*
\texttt{mlegp} & \mbox{0.925 (0.0213)} & \mbox{0.77 (0.90)} & \mbox{2.85 (1.03)} & \mbox{---} & \mbox{127 (14.5)} & \mbox{115 (10.7)} & \mbox{118 (16.2)} \\*
\texttt{MRFA} & \mbox{0.184 (0.0242)} & \mbox{0 (0)} & \mbox{1.51 (1.00)} & \mbox{---} & \mbox{27.5 (3.55)} & \mbox{67.5 (8.49)} & \mbox{67.9 (8.76)} \\*
\texttt{DART} & \mbox{0.104 (0.00516)} & \mbox{0 (0)} & \mbox{0 (0)} & \mbox{---} & \mbox{254 (18.4)} & \mbox{133 (20.4)} & \mbox{123 (18.6)} \\*
\texttt{SVGP-ARD} & \mbox{0.968 (0.0173)} & \mbox{NA} & \mbox{NA} & \mbox{---} & \mbox{123 (41.7)} & \mbox{211 (376)} & \mbox{134 (51.4)} \\*
\texttt{DGP-2} & \mbox{0.997 (0.0551)} & \mbox{NA} & \mbox{NA} & \mbox{---} & \mbox{560 (196)} & \mbox{574 (415)} & \mbox{580 (728)} \\*
\texttt{AS-GP} & \mbox{1.16 (0.0566)} & \mbox{NA} & \mbox{NA} & \mbox{---} & \mbox{5410 (895)} & \mbox{2730 (438)} & \mbox{1480 (255)} \\
\midrule
\multicolumn{8}{@{}l}{\textbf{$n=500$, $d=10$}}\\*
\texttt{optK}\textsubscript{0.005} & \mbox{0.0301 (0.00359)} & \mbox{0 (0)} & \mbox{0.10 (0.39)} & \mbox{17.8 (2.1)} & \mbox{21.3 (8.43)} & \mbox{23.8 (7.65)} & \mbox{22.0 (5.80)} \\*
\texttt{optK-pruned} & \mbox{0.0248 (0.00200)} & \mbox{0 (0)} & \mbox{0 (0)} & \mbox{17.2 (1.5)} & \mbox{20.9 (5.78)} & \mbox{24.1 (6.51)} & \mbox{23.0 (9.24)} \\*
\texttt{optK}\textsubscript{0.05} & \mbox{0.0249 (0.00235)} & \mbox{0 (0)} & \mbox{0 (0)} & \mbox{15.5 (0.9)} & \mbox{16.5 (3.16)} & \mbox{19.1 (3.90)} & \mbox{18.3 (3.53)} \\*
\texttt{laGP} & \mbox{0.891 (0.0147)} & \mbox{NA} & \mbox{NA} & \mbox{---} & \mbox{20.0 (7.28)} & \mbox{19.0 (6.88)} & \mbox{18.7 (6.92)} \\*
\texttt{mlegp} & \mbox{0.887 (0.0162)} & \mbox{0.76 (0.78)} & \mbox{2.59 (1.06)} & \mbox{---} & \mbox{614 (71.7)} & \mbox{571 (59.4)} & \mbox{581 (90.8)} \\*
\texttt{MRFA} & \mbox{0.0717 (0.00855)} & \mbox{0 (0)} & \mbox{1.74 (0.97)} & \mbox{---} & \mbox{46.2 (5.68)} & \mbox{105 (13.3)} & \mbox{110 (17.5)} \\*
\texttt{DART} & \mbox{0.0841 (0.00273)} & \mbox{0 (0)} & \mbox{0.01 (0.10)} & \mbox{---} & \mbox{275 (16.4)} & \mbox{144 (20.0)} & \mbox{131 (10.5)} \\*
\texttt{SVGP-ARD} & \mbox{0.952 (0.0142)} & \mbox{NA} & \mbox{NA} & \mbox{---} & \mbox{147 (49.7)} & \mbox{224 (438)} & \mbox{154 (51.7)} \\*
\texttt{DGP-2} & \mbox{0.735 (0.242)} & \mbox{NA} & \mbox{NA} & \mbox{---} & \mbox{1230 (404)} & \mbox{824 (467)} & \mbox{761 (383)} \\*
\texttt{AS-GP} & \mbox{1.03 (0.0734)} & \mbox{NA} & \mbox{NA} & \mbox{---} & \mbox{11800 (1880)} & \mbox{6020 (969)} & \mbox{3380 (692)} \\
\midrule
\multicolumn{8}{@{}l}{\textbf{$n=300$, $d=20$}}\\*
\texttt{optK}\textsubscript{0.005} & \mbox{0.0815 (0.0155)} & \mbox{0 (0)} & \mbox{0.12 (0.33)} & \mbox{16.5 (1.6)} & \mbox{8.64 (1.96)} & \mbox{13.1 (3.08)} & \mbox{12.8 (3.08)} \\*
\texttt{optK-pruned} & \mbox{0.0720 (0.0155)} & \mbox{0 (0)} & \mbox{0 (0)} & \mbox{16.7 (1.6)} & \mbox{8.99 (2.18)} & \mbox{13.8 (2.76)} & \mbox{14.1 (3.80)} \\*
\texttt{optK}\textsubscript{0.05} & \mbox{0.0930 (0.0740)} & \mbox{0 (0)} & \mbox{0 (0)} & \mbox{15.4 (0.9)} & \mbox{7.87 (1.40)} & \mbox{12.7 (3.60)} & \mbox{12.2 (3.25)} \\*
\texttt{laGP} & \mbox{0.967 (0.0263)} & \mbox{NA} & \mbox{NA} & \mbox{---} & \mbox{11.2 (4.77)} & \mbox{11.0 (5.20)} & \mbox{10.5 (4.75)} \\*
\texttt{mlegp} & \mbox{0.959 (0.0234)} & \mbox{0.14 (0.43)} & \mbox{12.34 (1.84)} & \mbox{---} & \mbox{144 (19.0)} & \mbox{131 (13.8)} & \mbox{133 (17.4)} \\*
\texttt{MRFA} & \mbox{0.200 (0.0349)} & \mbox{0 (0)} & \mbox{4.31 (1.52)} & \mbox{---} & \mbox{29.8 (4.43)} & \mbox{75.0 (11.4)} & \mbox{76.9 (12.1)} \\*
\texttt{DART} & \mbox{0.103 (0.00546)} & \mbox{0 (0)} & \mbox{0 (0)} & \mbox{---} & \mbox{255 (17.7)} & \mbox{131 (12.2)} & \mbox{120 (9.51)} \\*
\texttt{SVGP-ARD} & \mbox{0.990 (0.0215)} & \mbox{NA} & \mbox{NA} & \mbox{---} & \mbox{130 (46.2)} & \mbox{233 (422)} & \mbox{144 (47.1)} \\*
\texttt{DGP-2} & \mbox{1.03 (0.0457)} & \mbox{NA} & \mbox{NA} & \mbox{---} & \mbox{547 (93.2)} & \mbox{544 (707)} & \mbox{704 (978)} \\*
\texttt{AS-GP} & \mbox{1.28 (0.0592)} & \mbox{NA} & \mbox{NA} & \mbox{---} & \mbox{5480 (779)} & \mbox{2740 (364)} & \mbox{1460 (239)} \\
\midrule
\multicolumn{8}{@{}l}{\textbf{$n=500$, $d=20$}}\\*
\texttt{optK}\textsubscript{0.005} & \mbox{0.0325 (0.00416)} & \mbox{0 (0)} & \mbox{0.01 (0.10)} & \mbox{17.0 (1.2)} & \mbox{25.4 (4.31)} & \mbox{25.8 (6.23)} & \mbox{23.4 (6.25)} \\*
\texttt{optK-pruned} & \mbox{0.0250 (0.00265)} & \mbox{0 (0)} & \mbox{0 (0)} & \mbox{17.4 (1.1)} & \mbox{27.9 (5.21)} & \mbox{27.5 (5.83)} & \mbox{25.0 (5.44)} \\*
\texttt{optK}\textsubscript{0.05} & \mbox{0.0249 (0.00289)} & \mbox{0 (0)} & \mbox{0 (0)} & \mbox{15.7 (0.7)} & \mbox{24.3 (4.51)} & \mbox{23.9 (4.42)} & \mbox{21.8 (6.03)} \\*
\texttt{laGP} & \mbox{0.929 (0.0171)} & \mbox{NA} & \mbox{NA} & \mbox{---} & \mbox{34.7 (16.6)} & \mbox{32.4 (14.4)} & \mbox{31.7 (14.5)} \\*
\texttt{mlegp} & \mbox{0.942 (0.0199)} & \mbox{0.05 (0.26)} & \mbox{12.97 (1.94)} & \mbox{---} & \mbox{676 (66.7)} & \mbox{632 (56.3)} & \mbox{646 (106)} \\*
\texttt{MRFA} & \mbox{0.0792 (0.0106)} & \mbox{0 (0)} & \mbox{5.06 (1.64)} & \mbox{---} & \mbox{49.2 (6.16)} & \mbox{116 (17.0)} & \mbox{120 (17.4)} \\*
\texttt{DART} & \mbox{0.0840 (0.00322)} & \mbox{0 (0)} & \mbox{0.02 (0.14)} & \mbox{---} & \mbox{278 (22.6)} & \mbox{145 (14.4)} & \mbox{135 (26.2)} \\*
\texttt{SVGP-ARD} & \mbox{0.975 (0.0177)} & \mbox{NA} & \mbox{NA} & \mbox{---} & \mbox{147 (49.0)} & \mbox{260 (471)} & \mbox{151 (49.9)} \\*
\texttt{DGP-2} & \mbox{1.01 (0.0638)} & \mbox{NA} & \mbox{NA} & \mbox{---} & \mbox{849 (195)} & \mbox{609 (245)} & \mbox{568 (320)} \\*
\texttt{AS-GP} & \mbox{1.21 (0.0338)} & \mbox{NA} & \mbox{NA} & \mbox{---} & \mbox{12400 (1750)} & \mbox{6380 (1160)} & \mbox{3430 (555)} \\
\end{longtable}\endgroup

\begingroup\spacingset{1}
\begin{figure}[p]\centering
\includegraphics[width=\linewidth]{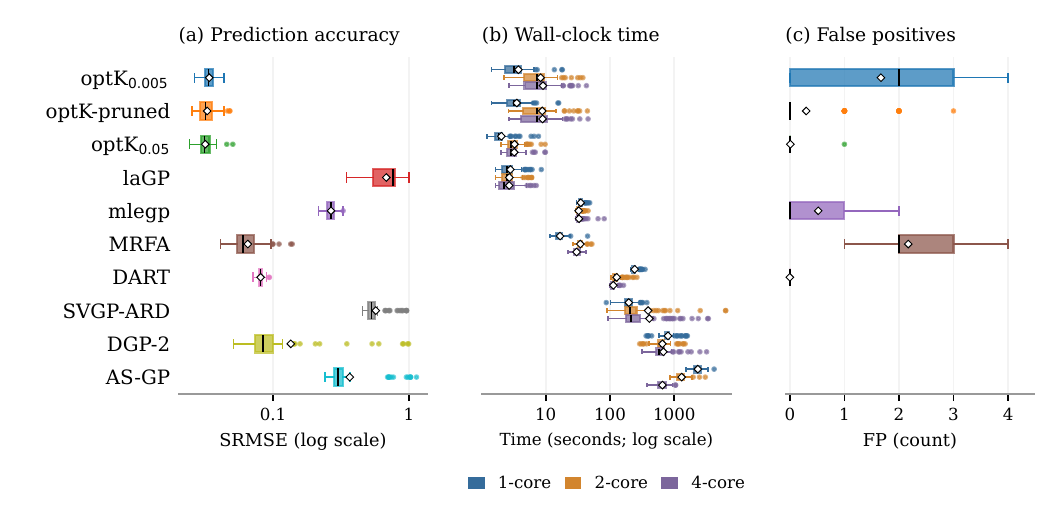}
\caption{Prediction, time and variable selection for the two-variable Michalewicz function with $\bm a=(1,2)$, $n=200$, $d=6$, $p=2$.}\label{fig:s-case01}\end{figure}
\begin{figure}[p]\centering
\includegraphics[width=\linewidth]{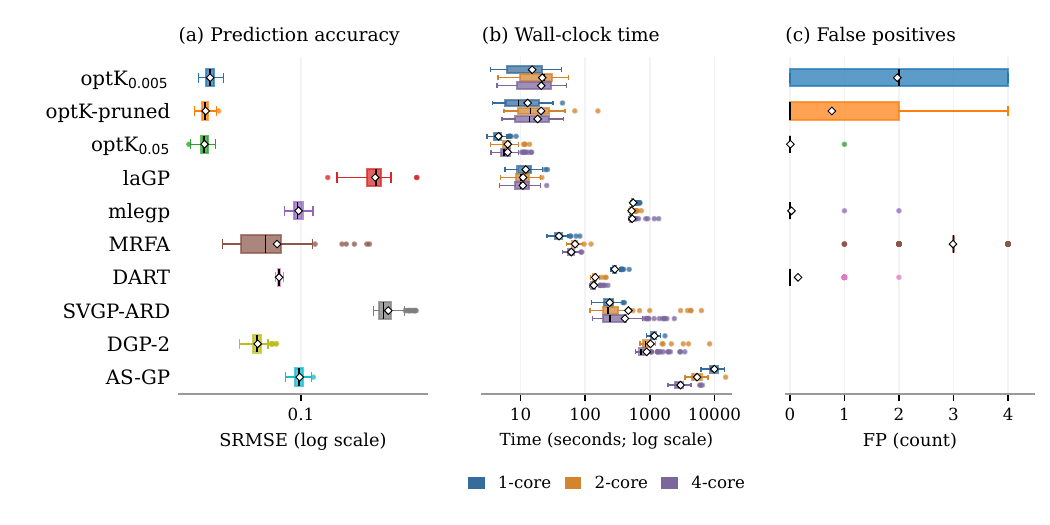}
\caption{Prediction, time and variable selection for the two-variable Michalewicz function with $\bm a=(1,2)$, $n=500$, $d=6$, $p=2$.}\label{fig:s-case02}\end{figure}
\begin{figure}[p]\centering
\includegraphics[width=\linewidth]{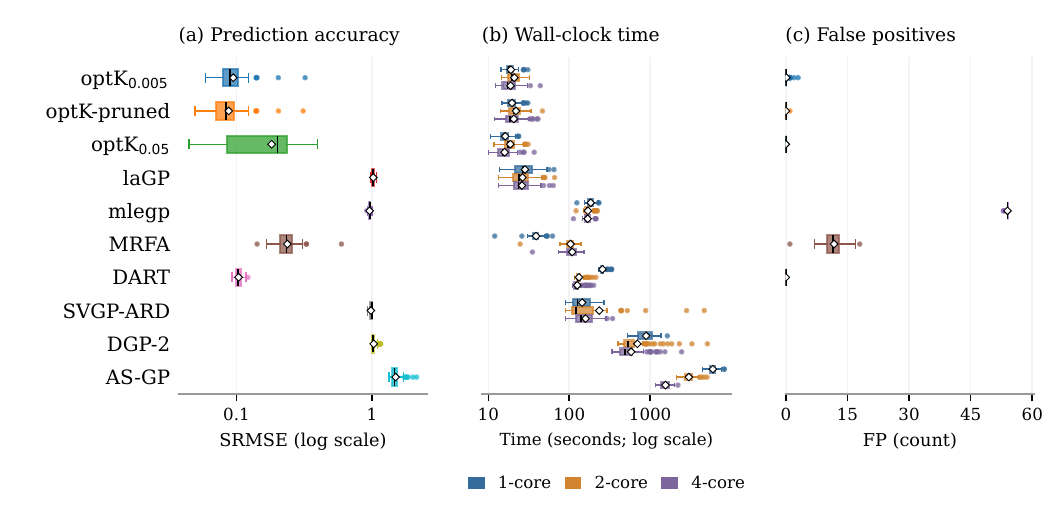}
\caption{Prediction, time and variable selection for the six-variable Michalewicz function with $\bm a=(1,2,1,2,1,2)$, $n=300$, $d=60$, $p=6$.}\label{fig:s-case26}\end{figure}\clearpage
\endgroup

\subsection{Additional Borehole Example}\label{sec:s3-borehole}
Table~\ref{tab:s-borehole} reports results for the Borehole function at $d=20,40$ and $n=200,500$. Figure~\ref{fig:s-case14} gives the SRMSE, time, FN and FP distributions at $d=60,n=200$. 
The $d=60,n=500$ distributions are shown in Figure~\ref{fig:main-borehole-n500} of the main text.
\par\begingroup\spacingset{1}\footnotesize
\setlength{\tabcolsep}{2pt}\renewcommand{\arraystretch}{1}
\def\ptmeasure#1#2{\settowidth{\dimen0}{#2}\ifdim\dimen0>#1\relax\edef#1{\the\dimen0}\fi}
\def\ptMethodWidth{0pt}
\ptmeasure\ptMethodWidth{Method}
\ptmeasure\ptMethodWidth{\texttt{AS-GP}}
\ptmeasure\ptMethodWidth{\texttt{DART}}
\ptmeasure\ptMethodWidth{\texttt{DGP-2}}
\ptmeasure\ptMethodWidth{\texttt{MRFA}}
\ptmeasure\ptMethodWidth{\texttt{SVGP-ARD}}
\ptmeasure\ptMethodWidth{\texttt{laGP}}
\ptmeasure\ptMethodWidth{\texttt{mlegp}}
\ptmeasure\ptMethodWidth{\texttt{optK-pruned}}
\ptmeasure\ptMethodWidth{\texttt{optK}\textsubscript{0.005}}
\ptmeasure\ptMethodWidth{\texttt{optK}\textsubscript{0.05}}
\def\ptSrmseWidth{0pt}
\ptmeasure\ptSrmseWidth{SRMSE}
\ptmeasure\ptSrmseWidth{\mbox{0.00160 (0.000169)}}
\ptmeasure\ptSrmseWidth{\mbox{0.00346 (0.00102)}}
\ptmeasure\ptSrmseWidth{\mbox{0.00429 (0.00135)}}
\ptmeasure\ptSrmseWidth{\mbox{0.0262 (0.00637)}}
\ptmeasure\ptSrmseWidth{\mbox{0.0279 (0.00442)}}
\ptmeasure\ptSrmseWidth{\mbox{0.0292 (0.00671)}}
\ptmeasure\ptSrmseWidth{\mbox{0.0300 (0.00776)}}
\ptmeasure\ptSrmseWidth{\mbox{0.0306 (0.00805)}}
\ptmeasure\ptSrmseWidth{\mbox{0.0447 (0.0265)}}
\ptmeasure\ptSrmseWidth{\mbox{0.0455 (0.00665)}}
\ptmeasure\ptSrmseWidth{\mbox{0.0461 (0.0539)}}
\ptmeasure\ptSrmseWidth{\mbox{0.0464 (0.00814)}}
\ptmeasure\ptSrmseWidth{\mbox{0.0473 (0.00700)}}
\ptmeasure\ptSrmseWidth{\mbox{0.0490 (0.0176)}}
\ptmeasure\ptSrmseWidth{\mbox{0.0501 (0.00339)}}
\ptmeasure\ptSrmseWidth{\mbox{0.0512 (0.0447)}}
\ptmeasure\ptSrmseWidth{\mbox{0.0521 (0.0313)}}
\ptmeasure\ptSrmseWidth{\mbox{0.0531 (0.00747)}}
\ptmeasure\ptSrmseWidth{\mbox{0.0552 (0.00297)}}
\ptmeasure\ptSrmseWidth{\mbox{0.0557 (0.00336)}}
\ptmeasure\ptSrmseWidth{\mbox{0.0574 (0.0146)}}
\ptmeasure\ptSrmseWidth{\mbox{0.0611 (0.0169)}}
\ptmeasure\ptSrmseWidth{\mbox{0.0657 (0.0149)}}
\ptmeasure\ptSrmseWidth{\mbox{0.0660 (0.0813)}}
\ptmeasure\ptSrmseWidth{\mbox{0.0691 (0.0170)}}
\ptmeasure\ptSrmseWidth{\mbox{0.0754 (0.00617)}}
\ptmeasure\ptSrmseWidth{\mbox{0.0762 (0.0231)}}
\ptmeasure\ptSrmseWidth{\mbox{0.0787 (0.0317)}}
\ptmeasure\ptSrmseWidth{\mbox{0.0787 (0.161)}}
\ptmeasure\ptSrmseWidth{\mbox{0.0849 (0.00726)}}
\ptmeasure\ptSrmseWidth{\mbox{0.0857 (0.00765)}}
\ptmeasure\ptSrmseWidth{\mbox{0.0861 (0.00765)}}
\ptmeasure\ptSrmseWidth{\mbox{0.0899 (0.00633)}}
\ptmeasure\ptSrmseWidth{\mbox{0.122 (0.0110)}}
\ptmeasure\ptSrmseWidth{\mbox{0.123 (0.0106)}}
\ptmeasure\ptSrmseWidth{\mbox{0.128 (0.0121)}}
\ptmeasure\ptSrmseWidth{\mbox{0.135 (0.0188)}}
\ptmeasure\ptSrmseWidth{\mbox{0.154 (0.0266)}}
\ptmeasure\ptSrmseWidth{\mbox{0.154 (0.160)}}
\ptmeasure\ptSrmseWidth{\mbox{0.166 (0.0185)}}
\def\ptFnWidth{0pt}
\ptmeasure\ptFnWidth{FN}
\ptmeasure\ptFnWidth{\mbox{2.01 (0.81)}}
\ptmeasure\ptFnWidth{\mbox{2.12 (0.77)}}
\ptmeasure\ptFnWidth{\mbox{2.17 (0.73)}}
\ptmeasure\ptFnWidth{\mbox{2.28 (0.73)}}
\ptmeasure\ptFnWidth{\mbox{2.30 (0.72)}}
\ptmeasure\ptFnWidth{\mbox{2.69 (0.58)}}
\ptmeasure\ptFnWidth{\mbox{2.71 (0.52)}}
\ptmeasure\ptFnWidth{\mbox{2.72 (0.45)}}
\ptmeasure\ptFnWidth{\mbox{2.83 (0.47)}}
\ptmeasure\ptFnWidth{\mbox{2.84 (0.37)}}
\ptmeasure\ptFnWidth{\mbox{2.86 (0.40)}}
\ptmeasure\ptFnWidth{\mbox{2.93 (0.29)}}
\ptmeasure\ptFnWidth{\mbox{2.95 (0.22)}}
\ptmeasure\ptFnWidth{\mbox{2.96 (0.20)}}
\ptmeasure\ptFnWidth{\mbox{2.99 (1.53)}}
\ptmeasure\ptFnWidth{\mbox{3.00 (0.00)}}
\ptmeasure\ptFnWidth{\mbox{3.02 (1.61)}}
\ptmeasure\ptFnWidth{\mbox{3.06 (0.24)}}
\ptmeasure\ptFnWidth{\mbox{3.07 (0.26)}}
\ptmeasure\ptFnWidth{\mbox{3.10 (0.33)}}
\ptmeasure\ptFnWidth{\mbox{3.11 (0.31)}}
\ptmeasure\ptFnWidth{\mbox{3.84 (1.30)}}
\ptmeasure\ptFnWidth{\mbox{4.49 (1.93)}}
\ptmeasure\ptFnWidth{\mbox{NA}}
\def\ptFpWidth{0pt}
\ptmeasure\ptFpWidth{FP}
\ptmeasure\ptFpWidth{\mbox{0 (0)}}
\ptmeasure\ptFpWidth{\mbox{0.01 (0.10)}}
\ptmeasure\ptFpWidth{\mbox{0.09 (0.29)}}
\ptmeasure\ptFpWidth{\mbox{0.10 (0.36)}}
\ptmeasure\ptFpWidth{\mbox{0.11 (0.35)}}
\ptmeasure\ptFpWidth{\mbox{0.12 (0.36)}}
\ptmeasure\ptFpWidth{\mbox{0.38 (0.55)}}
\ptmeasure\ptFpWidth{\mbox{0.52 (0.72)}}
\ptmeasure\ptFpWidth{\mbox{0.58 (0.71)}}
\ptmeasure\ptFpWidth{\mbox{0.69 (0.79)}}
\ptmeasure\ptFpWidth{\mbox{1.02 (0.91)}}
\ptmeasure\ptFpWidth{\mbox{1.13 (1.00)}}
\ptmeasure\ptFpWidth{\mbox{1.40 (1.31)}}
\ptmeasure\ptFpWidth{\mbox{1.47 (1.32)}}
\ptmeasure\ptFpWidth{\mbox{2.49 (1.28)}}
\ptmeasure\ptFpWidth{\mbox{2.69 (3.69)}}
\ptmeasure\ptFpWidth{\mbox{2.94 (9.19)}}
\ptmeasure\ptFpWidth{\mbox{3.29 (3.56)}}
\ptmeasure\ptFpWidth{\mbox{3.40 (1.54)}}
\ptmeasure\ptFpWidth{\mbox{3.42 (1.40)}}
\ptmeasure\ptFpWidth{\mbox{3.65 (2.02)}}
\ptmeasure\ptFpWidth{\mbox{3.96 (10.42)}}
\ptmeasure\ptFpWidth{\mbox{7.39 (1.92)}}
\ptmeasure\ptFpWidth{\mbox{7.58 (1.93)}}
\ptmeasure\ptFpWidth{\mbox{NA}}
\def\ptIterWidth{0pt}
\ptmeasure\ptIterWidth{Iter}
\ptmeasure\ptIterWidth{\mbox{---}}
\ptmeasure\ptIterWidth{\mbox{18.4 (3.7)}}
\ptmeasure\ptIterWidth{\mbox{18.5 (3.3)}}
\ptmeasure\ptIterWidth{\mbox{18.6 (3.5)}}
\ptmeasure\ptIterWidth{\mbox{19.2 (3.3)}}
\ptmeasure\ptIterWidth{\mbox{35.4 (3.6)}}
\ptmeasure\ptIterWidth{\mbox{35.6 (3.9)}}
\ptmeasure\ptIterWidth{\mbox{44.6 (4.1)}}
\ptmeasure\ptIterWidth{\mbox{44.8 (3.8)}}
\ptmeasure\ptIterWidth{\mbox{45.0 (5.0)}}
\ptmeasure\ptIterWidth{\mbox{46.1 (4.9)}}
\ptmeasure\ptIterWidth{\mbox{53.0 (5.9)}}
\ptmeasure\ptIterWidth{\mbox{53.0 (6.1)}}
\def\ptTimeWidth{0pt}
\ptmeasure\ptTimeWidth{\makecell[r]{1-core\\(Sec)}}
\ptmeasure\ptTimeWidth{\makecell[r]{2-core\\(Sec)}}
\ptmeasure\ptTimeWidth{\makecell[r]{4-core\\(Sec)}}
\ptmeasure\ptTimeWidth{\mbox{10.6 (3.42)}}
\ptmeasure\ptTimeWidth{\mbox{103 (72.4)}}
\ptmeasure\ptTimeWidth{\mbox{106 (74.3)}}
\ptmeasure\ptTimeWidth{\mbox{11.4 (3.78)}}
\ptmeasure\ptTimeWidth{\mbox{11.9 (4.49)}}
\ptmeasure\ptTimeWidth{\mbox{1130 (290)}}
\ptmeasure\ptTimeWidth{\mbox{1130 (810)}}
\ptmeasure\ptTimeWidth{\mbox{116 (9.36)}}
\ptmeasure\ptTimeWidth{\mbox{1170 (601)}}
\ptmeasure\ptTimeWidth{\mbox{1170 (836)}}
\ptmeasure\ptTimeWidth{\mbox{1180 (536)}}
\ptmeasure\ptTimeWidth{\mbox{120 (16.5)}}
\ptmeasure\ptTimeWidth{\mbox{130 (84.1)}}
\ptmeasure\ptTimeWidth{\mbox{143 (192)}}
\ptmeasure\ptTimeWidth{\mbox{14700 (2510)}}
\ptmeasure\ptTimeWidth{\mbox{15.4 (3.59)}}
\ptmeasure\ptTimeWidth{\mbox{151 (193)}}
\ptmeasure\ptTimeWidth{\mbox{152 (206)}}
\ptmeasure\ptTimeWidth{\mbox{16500 (2500)}}
\ptmeasure\ptTimeWidth{\mbox{17.0 (4.25)}}
\ptmeasure\ptTimeWidth{\mbox{1770 (317)}}
\ptmeasure\ptTimeWidth{\mbox{180 (27.5)}}
\ptmeasure\ptTimeWidth{\mbox{1800 (481)}}
\ptmeasure\ptTimeWidth{\mbox{182 (27.5)}}
\ptmeasure\ptTimeWidth{\mbox{1860 (475)}}
\ptmeasure\ptTimeWidth{\mbox{1890 (693)}}
\ptmeasure\ptTimeWidth{\mbox{20.1 (7.35)}}
\ptmeasure\ptTimeWidth{\mbox{215 (101)}}
\ptmeasure\ptTimeWidth{\mbox{217 (47.8)}}
\ptmeasure\ptTimeWidth{\mbox{219 (112)}}
\ptmeasure\ptTimeWidth{\mbox{22.1 (7.86)}}
\ptmeasure\ptTimeWidth{\mbox{22.4 (6.68)}}
\ptmeasure\ptTimeWidth{\mbox{224 (49.1)}}
\ptmeasure\ptTimeWidth{\mbox{23.5 (12.0)}}
\ptmeasure\ptTimeWidth{\mbox{246 (51.6)}}
\ptmeasure\ptTimeWidth{\mbox{253 (55.2)}}
\ptmeasure\ptTimeWidth{\mbox{26.3 (7.28)}}
\ptmeasure\ptTimeWidth{\mbox{28.0 (5.82)}}
\ptmeasure\ptTimeWidth{\mbox{284 (135)}}
\ptmeasure\ptTimeWidth{\mbox{3.46 (1.01)}}
\ptmeasure\ptTimeWidth{\mbox{3.51 (1.00)}}
\ptmeasure\ptTimeWidth{\mbox{3.56 (1.11)}}
\ptmeasure\ptTimeWidth{\mbox{3.66 (1.10)}}
\ptmeasure\ptTimeWidth{\mbox{3.79 (1.24)}}
\ptmeasure\ptTimeWidth{\mbox{3.85 (1.67)}}
\ptmeasure\ptTimeWidth{\mbox{30.3 (4.47)}}
\ptmeasure\ptTimeWidth{\mbox{312 (175)}}
\ptmeasure\ptTimeWidth{\mbox{32.3 (9.06)}}
\ptmeasure\ptTimeWidth{\mbox{32.8 (7.93)}}
\ptmeasure\ptTimeWidth{\mbox{325 (353)}}
\ptmeasure\ptTimeWidth{\mbox{3360 (656)}}
\ptmeasure\ptTimeWidth{\mbox{337 (527)}}
\ptmeasure\ptTimeWidth{\mbox{34.7 (8.42)}}
\ptmeasure\ptTimeWidth{\mbox{3540 (762)}}
\ptmeasure\ptTimeWidth{\mbox{36.4 (3.33)}}
\ptmeasure\ptTimeWidth{\mbox{36.6 (7.04)}}
\ptmeasure\ptTimeWidth{\mbox{360 (277)}}
\ptmeasure\ptTimeWidth{\mbox{37.3 (11.1)}}
\ptmeasure\ptTimeWidth{\mbox{37.3 (21.5)}}
\ptmeasure\ptTimeWidth{\mbox{37.4 (23.2)}}
\ptmeasure\ptTimeWidth{\mbox{376 (471)}}
\ptmeasure\ptTimeWidth{\mbox{38.2 (11.6)}}
\ptmeasure\ptTimeWidth{\mbox{38.2 (6.28)}}
\ptmeasure\ptTimeWidth{\mbox{38.7 (7.16)}}
\ptmeasure\ptTimeWidth{\mbox{39.7 (23.8)}}
\ptmeasure\ptTimeWidth{\mbox{391 (674)}}
\ptmeasure\ptTimeWidth{\mbox{4.68 (1.22)}}
\ptmeasure\ptTimeWidth{\mbox{40.9 (12.7)}}
\ptmeasure\ptTimeWidth{\mbox{402 (238)}}
\ptmeasure\ptTimeWidth{\mbox{408 (818)}}
\ptmeasure\ptTimeWidth{\mbox{41.3 (5.74)}}
\ptmeasure\ptTimeWidth{\mbox{41.6 (6.40)}}
\ptmeasure\ptTimeWidth{\mbox{41.6 (7.89)}}
\ptmeasure\ptTimeWidth{\mbox{43.9 (4.76)}}
\ptmeasure\ptTimeWidth{\mbox{431 (790)}}
\ptmeasure\ptTimeWidth{\mbox{437 (264)}}
\ptmeasure\ptTimeWidth{\mbox{44.6 (24.3)}}
\ptmeasure\ptTimeWidth{\mbox{4450 (1700)}}
\ptmeasure\ptTimeWidth{\mbox{45.4 (23.1)}}
\ptmeasure\ptTimeWidth{\mbox{46.6 (32.3)}}
\ptmeasure\ptTimeWidth{\mbox{47.5 (8.47)}}
\ptmeasure\ptTimeWidth{\mbox{4760 (1100)}}
\ptmeasure\ptTimeWidth{\mbox{477 (793)}}
\ptmeasure\ptTimeWidth{\mbox{49.3 (26.5)}}
\ptmeasure\ptTimeWidth{\mbox{49.4 (42.4)}}
\ptmeasure\ptTimeWidth{\mbox{490 (428)}}
\ptmeasure\ptTimeWidth{\mbox{5.40 (1.41)}}
\ptmeasure\ptTimeWidth{\mbox{50.1 (19.8)}}
\ptmeasure\ptTimeWidth{\mbox{508 (418)}}
\ptmeasure\ptTimeWidth{\mbox{51.2 (9.57)}}
\ptmeasure\ptTimeWidth{\mbox{55.6 (4.94)}}
\ptmeasure\ptTimeWidth{\mbox{559 (1060)}}
\ptmeasure\ptTimeWidth{\mbox{56.9 (7.78)}}
\ptmeasure\ptTimeWidth{\mbox{568 (33.2)}}
\ptmeasure\ptTimeWidth{\mbox{572 (73.0)}}
\ptmeasure\ptTimeWidth{\mbox{606 (47.1)}}
\ptmeasure\ptTimeWidth{\mbox{61.0 (11.8)}}
\ptmeasure\ptTimeWidth{\mbox{61.1 (7.09)}}
\ptmeasure\ptTimeWidth{\mbox{61.9 (10.6)}}
\ptmeasure\ptTimeWidth{\mbox{617 (252)}}
\ptmeasure\ptTimeWidth{\mbox{629 (505)}}
\ptmeasure\ptTimeWidth{\mbox{635 (59.1)}}
\ptmeasure\ptTimeWidth{\mbox{648 (135)}}
\ptmeasure\ptTimeWidth{\mbox{67.0 (11.6)}}
\ptmeasure\ptTimeWidth{\mbox{670 (444)}}
\ptmeasure\ptTimeWidth{\mbox{683 (67.5)}}
\ptmeasure\ptTimeWidth{\mbox{687 (316)}}
\ptmeasure\ptTimeWidth{\mbox{713 (1460)}}
\ptmeasure\ptTimeWidth{\mbox{7960 (1780)}}
\ptmeasure\ptTimeWidth{\mbox{838 (210)}}
\ptmeasure\ptTimeWidth{\mbox{839 (605)}}
\ptmeasure\ptTimeWidth{\mbox{85.5 (10.5)}}
\ptmeasure\ptTimeWidth{\mbox{851 (673)}}
\ptmeasure\ptTimeWidth{\mbox{86.6 (13.2)}}
\ptmeasure\ptTimeWidth{\mbox{8980 (2390)}}
\ptmeasure\ptTimeWidth{\mbox{9.87 (3.27)}}
\ptmeasure\ptTimeWidth{\mbox{91.8 (11.5)}}
\ptmeasure\ptTimeWidth{\mbox{925 (153)}}
\ptmeasure\ptTimeWidth{\mbox{93.9 (12.1)}}
\ptmeasure\ptTimeWidth{\mbox{958 (221)}}
\def\ptUpdateWidth{\edef\ptTableWidth{\the\dimexpr\ptMethodWidth+\ptSrmseWidth+\ptFnWidth+\ptFpWidth+\ptIterWidth+\ptTimeWidth+\ptTimeWidth+\ptTimeWidth+14\tabcolsep\relax}}
\ptUpdateWidth\ifdim\ptTableWidth>\linewidth\setlength{\tabcolsep}{1pt}\ptUpdateWidth\fi
\typeout{PERFORMANCE-WIDTH supp_borehole: table=\ptTableWidth; time=\ptTimeWidth; linewidth=\the\linewidth}
\ifdim\ptTableWidth>\linewidth\PackageError{performance-tables}{Table exceeds linewidth}{Check the table layout.}\fi
\setlength{\LTcapwidth}{\ptTableWidth}
\Needspace{570pt}
\begin{longtable}{@{}>{\raggedright\arraybackslash}p{\ptMethodWidth}>{\raggedleft\arraybackslash}p{\ptSrmseWidth}>{\raggedleft\arraybackslash}p{\ptFnWidth}>{\raggedleft\arraybackslash}p{\ptFpWidth}>{\raggedleft\arraybackslash}p{\ptIterWidth}>{\raggedleft\arraybackslash}p{\ptTimeWidth}>{\raggedleft\arraybackslash}p{\ptTimeWidth}>{\raggedleft\arraybackslash}p{\ptTimeWidth}@{}}
\caption{Borehole function; $d=20,40$, $p=8$, $m=1000$, $B=100$.}\label{tab:s-borehole}\\
\toprule
Method & SRMSE & FN & FP & Iter & \makecell[r]{1-core\\(Sec)} & \makecell[r]{2-core\\(Sec)} & \makecell[r]{4-core\\(Sec)} \\
\midrule\endfirsthead
\caption[]{Borehole function; $d=20,40$, $p=8$, $m=1000$, $B=100$. (continued).}\\\toprule
Method & SRMSE & FN & FP & Iter & \makecell[r]{1-core\\(Sec)} & \makecell[r]{2-core\\(Sec)} & \makecell[r]{4-core\\(Sec)} \\
\midrule\endhead
\midrule
\multicolumn{8}{r@{}}{\footnotesize Continued on next page.}\\\endfoot
\bottomrule
\endlastfoot
\multicolumn{8}{@{}l}{\textbf{$n=200$, $d=20$}}\\*
\texttt{optK}\textsubscript{0.005} & \mbox{0.0657 (0.0149)} & \mbox{2.28 (0.73)} & \mbox{2.49 (1.28)} & \mbox{35.4 (3.6)} & \mbox{49.4 (42.4)} & \mbox{49.3 (26.5)} & \mbox{44.6 (24.3)} \\*
\texttt{optK-pruned} & \mbox{0.0574 (0.0146)} & \mbox{2.72 (0.45)} & \mbox{1.02 (0.91)} & \mbox{35.6 (3.9)} & \mbox{46.6 (32.3)} & \mbox{50.1 (19.8)} & \mbox{45.4 (23.1)} \\*
\texttt{optK}\textsubscript{0.05} & \mbox{0.0787 (0.0317)} & \mbox{3.10 (0.33)} & \mbox{0.12 (0.36)} & \mbox{19.2 (3.3)} & \mbox{4.68 (1.22)} & \mbox{9.87 (3.27)} & \mbox{10.6 (3.42)} \\*
\texttt{laGP} & \mbox{0.0857 (0.00765)} & \mbox{NA} & \mbox{NA} & \mbox{---} & \mbox{3.85 (1.67)} & \mbox{3.46 (1.01)} & \mbox{3.51 (1.00)} \\*
\texttt{mlegp} & \mbox{0.0512 (0.0447)} & \mbox{2.99 (1.53)} & \mbox{2.69 (3.69)} & \mbox{---} & \mbox{38.7 (7.16)} & \mbox{36.4 (3.33)} & \mbox{36.6 (7.04)} \\*
\texttt{MRFA} & \mbox{0.135 (0.0188)} & \mbox{2.12 (0.77)} & \mbox{3.42 (1.40)} & \mbox{---} & \mbox{15.4 (3.59)} & \mbox{30.3 (4.47)} & \mbox{32.8 (7.93)} \\*
\texttt{DART} & \mbox{0.0849 (0.00726)} & \mbox{3.00 (0.00)} & \mbox{0.01 (0.10)} & \mbox{---} & \mbox{116 (9.36)} & \mbox{61.9 (10.6)} & \mbox{55.6 (4.94)} \\*
\texttt{SVGP-ARD} & \mbox{0.128 (0.0121)} & \mbox{NA} & \mbox{NA} & \mbox{---} & \mbox{217 (47.8)} & \mbox{559 (1060)} & \mbox{431 (790)} \\*
\texttt{DGP-2} & \mbox{0.0531 (0.00747)} & \mbox{NA} & \mbox{NA} & \mbox{---} & \mbox{838 (210)} & \mbox{670 (444)} & \mbox{629 (505)} \\*
\texttt{AS-GP} & \mbox{0.00429 (0.00135)} & \mbox{NA} & \mbox{NA} & \mbox{---} & \mbox{3360 (656)} & \mbox{1770 (317)} & \mbox{925 (153)} \\
\midrule
\multicolumn{8}{@{}l}{\textbf{$n=500$, $d=20$}}\\*
\texttt{optK}\textsubscript{0.005} & \mbox{0.0292 (0.00671)} & \mbox{2.69 (0.58)} & \mbox{1.40 (1.31)} & \mbox{44.6 (4.1)} & \mbox{402 (238)} & \mbox{284 (135)} & \mbox{219 (112)} \\*
\texttt{optK-pruned} & \mbox{0.0262 (0.00637)} & \mbox{2.93 (0.29)} & \mbox{0.38 (0.55)} & \mbox{44.8 (3.8)} & \mbox{437 (264)} & \mbox{312 (175)} & \mbox{215 (101)} \\*
\texttt{optK}\textsubscript{0.05} & \mbox{0.0521 (0.0313)} & \mbox{3.11 (0.31)} & \mbox{0.10 (0.36)} & \mbox{18.4 (3.7)} & \mbox{23.5 (12.0)} & \mbox{22.1 (7.86)} & \mbox{20.1 (7.35)} \\*
\texttt{laGP} & \mbox{0.0501 (0.00339)} & \mbox{NA} & \mbox{NA} & \mbox{---} & \mbox{40.9 (12.7)} & \mbox{38.2 (11.6)} & \mbox{37.3 (11.1)} \\*
\texttt{mlegp} & \mbox{0.0461 (0.0539)} & \mbox{3.02 (1.61)} & \mbox{3.29 (3.56)} & \mbox{---} & \mbox{606 (47.1)} & \mbox{568 (33.2)} & \mbox{572 (73.0)} \\*
\texttt{MRFA} & \mbox{0.0455 (0.00665)} & \mbox{2.01 (0.81)} & \mbox{3.40 (1.54)} & \mbox{---} & \mbox{28.0 (5.82)} & \mbox{47.5 (8.47)} & \mbox{51.2 (9.57)} \\*
\texttt{DART} & \mbox{0.0557 (0.00336)} & \mbox{2.86 (0.40)} & \mbox{0.58 (0.71)} & \mbox{---} & \mbox{180 (27.5)} & \mbox{91.8 (11.5)} & \mbox{86.6 (13.2)} \\*
\texttt{SVGP-ARD} & \mbox{0.0899 (0.00633)} & \mbox{NA} & \mbox{NA} & \mbox{---} & \mbox{246 (51.6)} & \mbox{408 (818)} & \mbox{376 (471)} \\*
\texttt{DGP-2} & \mbox{0.0279 (0.00442)} & \mbox{NA} & \mbox{NA} & \mbox{---} & \mbox{1800 (481)} & \mbox{1170 (601)} & \mbox{1130 (810)} \\*
\texttt{AS-GP} & \mbox{0.00160 (0.000169)} & \mbox{NA} & \mbox{NA} & \mbox{---} & \mbox{14700 (2510)} & \mbox{7960 (1780)} & \mbox{4450 (1700)} \\
\midrule
\multicolumn{8}{@{}l}{\textbf{$n=200$, $d=40$}}\\*
\texttt{optK}\textsubscript{0.005} & \mbox{0.0691 (0.0170)} & \mbox{2.71 (0.52)} & \mbox{3.65 (2.02)} & \mbox{46.1 (4.9)} & \mbox{151 (193)} & \mbox{130 (84.1)} & \mbox{106 (74.3)} \\*
\texttt{optK-pruned} & \mbox{0.0611 (0.0169)} & \mbox{2.83 (0.47)} & \mbox{1.13 (1.00)} & \mbox{45.0 (5.0)} & \mbox{152 (206)} & \mbox{143 (192)} & \mbox{103 (72.4)} \\*
\texttt{optK}\textsubscript{0.05} & \mbox{0.0762 (0.0231)} & \mbox{3.06 (0.24)} & \mbox{0.11 (0.35)} & \mbox{18.5 (3.3)} & \mbox{5.40 (1.41)} & \mbox{11.9 (4.49)} & \mbox{11.4 (3.78)} \\*
\texttt{laGP} & \mbox{0.122 (0.0110)} & \mbox{NA} & \mbox{NA} & \mbox{---} & \mbox{3.79 (1.24)} & \mbox{3.66 (1.10)} & \mbox{3.56 (1.11)} \\*
\texttt{mlegp} & \mbox{0.0660 (0.0813)} & \mbox{3.84 (1.30)} & \mbox{2.94 (9.19)} & \mbox{---} & \mbox{43.9 (4.76)} & \mbox{41.6 (6.40)} & \mbox{41.3 (5.74)} \\*
\texttt{MRFA} & \mbox{0.154 (0.0266)} & \mbox{2.30 (0.72)} & \mbox{7.39 (1.92)} & \mbox{---} & \mbox{17.0 (4.25)} & \mbox{38.2 (6.28)} & \mbox{41.6 (7.89)} \\*
\texttt{DART} & \mbox{0.0861 (0.00765)} & \mbox{3.00 (0.00)} & \mbox{0 (0)} & \mbox{---} & \mbox{120 (16.5)} & \mbox{61.1 (7.09)} & \mbox{56.9 (7.78)} \\*
\texttt{SVGP-ARD} & \mbox{0.166 (0.0185)} & \mbox{NA} & \mbox{NA} & \mbox{---} & \mbox{224 (49.1)} & \mbox{337 (527)} & \mbox{477 (793)} \\*
\texttt{DGP-2} & \mbox{0.154 (0.160)} & \mbox{NA} & \mbox{NA} & \mbox{---} & \mbox{1130 (290)} & \mbox{839 (605)} & \mbox{851 (673)} \\*
\texttt{AS-GP} & \mbox{0.0490 (0.0176)} & \mbox{NA} & \mbox{NA} & \mbox{---} & \mbox{3540 (762)} & \mbox{1860 (475)} & \mbox{958 (221)} \\*
\midrule
\multicolumn{8}{@{}l}{\textbf{$n=500$, $d=40$}}\\*
\texttt{optK}\textsubscript{0.005} & \mbox{0.0300 (0.00776)} & \mbox{2.84 (0.37)} & \mbox{1.47 (1.32)} & \mbox{53.0 (5.9)} & \mbox{687 (316)} & \mbox{508 (418)} & \mbox{360 (277)} \\*
\texttt{optK-pruned} & \mbox{0.0306 (0.00805)} & \mbox{2.95 (0.22)} & \mbox{0.52 (0.72)} & \mbox{53.0 (6.1)} & \mbox{617 (252)} & \mbox{490 (428)} & \mbox{325 (353)} \\*
\texttt{optK}\textsubscript{0.05} & \mbox{0.0447 (0.0265)} & \mbox{3.07 (0.26)} & \mbox{0.09 (0.29)} & \mbox{18.6 (3.5)} & \mbox{32.3 (9.06)} & \mbox{26.3 (7.28)} & \mbox{22.4 (6.68)} \\*
\texttt{laGP} & \mbox{0.0754 (0.00617)} & \mbox{NA} & \mbox{NA} & \mbox{---} & \mbox{39.7 (23.8)} & \mbox{37.3 (21.5)} & \mbox{37.4 (23.2)} \\*
\texttt{mlegp} & \mbox{0.0787 (0.161)} & \mbox{4.49 (1.93)} & \mbox{3.96 (10.42)} & \mbox{---} & \mbox{683 (67.5)} & \mbox{635 (59.1)} & \mbox{648 (135)} \\*
\texttt{MRFA} & \mbox{0.0473 (0.00700)} & \mbox{2.17 (0.73)} & \mbox{7.58 (1.93)} & \mbox{---} & \mbox{34.7 (8.42)} & \mbox{61.0 (11.8)} & \mbox{67.0 (11.6)} \\*
\texttt{DART} & \mbox{0.0552 (0.00297)} & \mbox{2.96 (0.20)} & \mbox{0.69 (0.79)} & \mbox{---} & \mbox{182 (27.5)} & \mbox{93.9 (12.1)} & \mbox{85.5 (10.5)} \\*
\texttt{SVGP-ARD} & \mbox{0.123 (0.0106)} & \mbox{NA} & \mbox{NA} & \mbox{---} & \mbox{253 (55.2)} & \mbox{391 (674)} & \mbox{713 (1460)} \\*
\texttt{DGP-2} & \mbox{0.0464 (0.00814)} & \mbox{NA} & \mbox{NA} & \mbox{---} & \mbox{1890 (693)} & \mbox{1180 (536)} & \mbox{1170 (836)} \\*
\texttt{AS-GP} & \mbox{0.00346 (0.00102)} & \mbox{NA} & \mbox{NA} & \mbox{---} & \mbox{16500 (2500)} & \mbox{8980 (2390)} & \mbox{4760 (1100)} \\*
\end{longtable}\endgroup

\begingroup\spacingset{1}
\begin{figure}[p]\centering
\includegraphics[width=\linewidth]{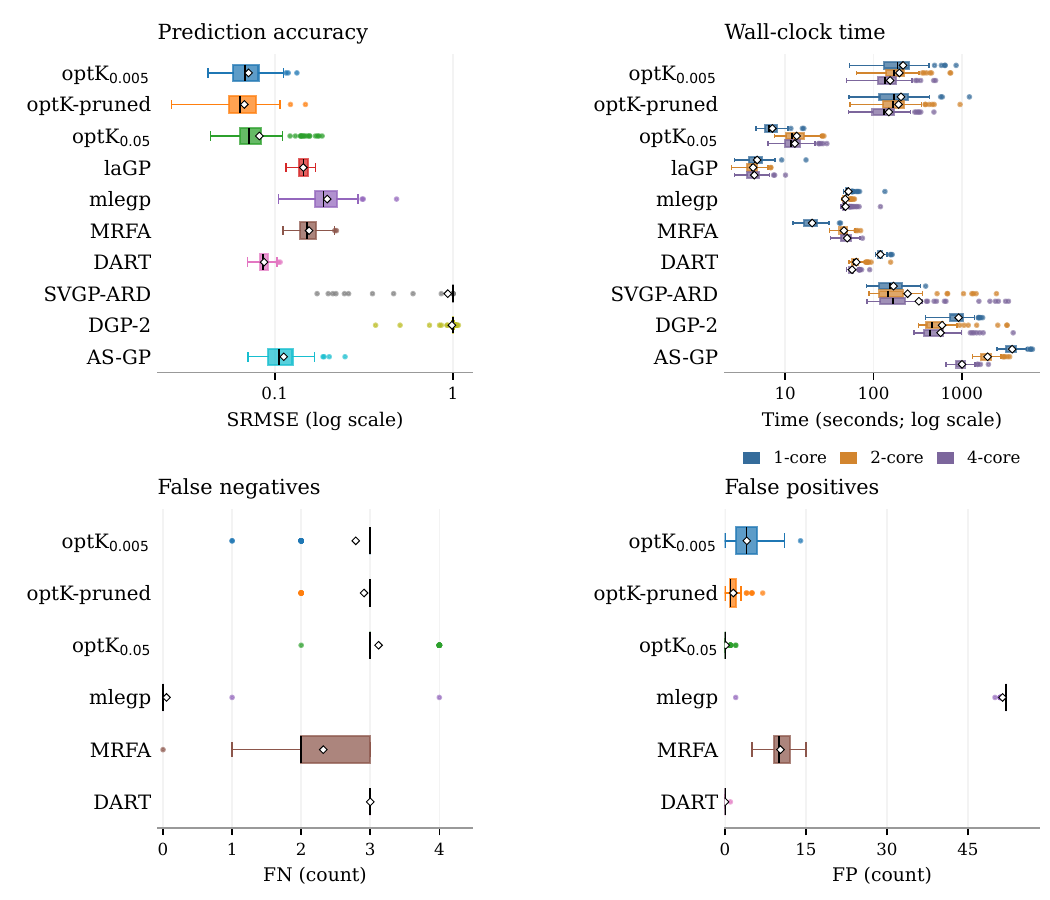}
\caption{Prediction, time and variable selection for the Borehole function; $n=200$, $d=60$, $p=8$.}\label{fig:s-case14}\end{figure}\clearpage
\endgroup

\subsection{Additional Satellite Example}\label{sec:s3-satellite}
Table~\ref{tab:s-satellite} reports results for the three Satellite responses at $n=200$. 
The $n=500$ results and distributions are provided in Table~\ref{tab:satellite} and Figures~\ref{fig:main-satellite-he}--\ref{fig:main-satellite-o2} of the main text. 
The FN/FP reference set is defined in Section~\ref{sec:satellite}.
\par\begingroup\spacingset{1}\footnotesize
\setlength{\tabcolsep}{2pt}\renewcommand{\arraystretch}{1}
\def\ptmeasure#1#2{\settowidth{\dimen0}{#2}\ifdim\dimen0>#1\relax\edef#1{\the\dimen0}\fi}
\def\ptMethodWidth{0pt}
\ptmeasure\ptMethodWidth{Method}
\ptmeasure\ptMethodWidth{\texttt{AS-GP}}
\ptmeasure\ptMethodWidth{\texttt{DART}}
\ptmeasure\ptMethodWidth{\texttt{DGP-2}}
\ptmeasure\ptMethodWidth{\texttt{MRFA}}
\ptmeasure\ptMethodWidth{\texttt{SVGP-ARD}}
\ptmeasure\ptMethodWidth{\texttt{laGP}}
\ptmeasure\ptMethodWidth{\texttt{mlegp}}
\ptmeasure\ptMethodWidth{\texttt{optK-pruned}}
\ptmeasure\ptMethodWidth{\texttt{optK}\textsubscript{0.005}}
\ptmeasure\ptMethodWidth{\texttt{optK}\textsubscript{0.05}}
\def\ptSrmseWidth{0pt}
\ptmeasure\ptSrmseWidth{SRMSE}
\ptmeasure\ptSrmseWidth{\mbox{0.0857 (0.0137)}}
\ptmeasure\ptSrmseWidth{\mbox{0.0863 (0.0165)}}
\ptmeasure\ptSrmseWidth{\mbox{0.0906 (0.0215)}}
\ptmeasure\ptSrmseWidth{\mbox{0.0935 (0.0112)}}
\ptmeasure\ptSrmseWidth{\mbox{0.0979 (0.0118)}}
\ptmeasure\ptSrmseWidth{\mbox{0.102 (0.0175)}}
\ptmeasure\ptSrmseWidth{\mbox{0.118 (0.0138)}}
\ptmeasure\ptSrmseWidth{\mbox{0.119 (0.00957)}}
\ptmeasure\ptSrmseWidth{\mbox{0.120 (0.0292)}}
\ptmeasure\ptSrmseWidth{\mbox{0.121 (0.0105)}}
\ptmeasure\ptSrmseWidth{\mbox{0.122 (0.0396)}}
\ptmeasure\ptSrmseWidth{\mbox{0.126 (0.0113)}}
\ptmeasure\ptSrmseWidth{\mbox{0.127 (0.0206)}}
\ptmeasure\ptSrmseWidth{\mbox{0.127 (0.0331)}}
\ptmeasure\ptSrmseWidth{\mbox{0.130 (0.0124)}}
\ptmeasure\ptSrmseWidth{\mbox{0.138 (0.0139)}}
\ptmeasure\ptSrmseWidth{\mbox{0.142 (0.0191)}}
\ptmeasure\ptSrmseWidth{\mbox{0.146 (0.0144)}}
\ptmeasure\ptSrmseWidth{\mbox{0.149 (0.0144)}}
\ptmeasure\ptSrmseWidth{\mbox{0.149 (0.0230)}}
\ptmeasure\ptSrmseWidth{\mbox{0.151 (0.0136)}}
\ptmeasure\ptSrmseWidth{\mbox{0.161 (0.0285)}}
\ptmeasure\ptSrmseWidth{\mbox{0.164 (0.0145)}}
\ptmeasure\ptSrmseWidth{\mbox{0.187 (0.0140)}}
\ptmeasure\ptSrmseWidth{\mbox{0.191 (0.0176)}}
\ptmeasure\ptSrmseWidth{\mbox{0.207 (0.115)}}
\ptmeasure\ptSrmseWidth{\mbox{0.208 (0.0132)}}
\ptmeasure\ptSrmseWidth{\mbox{0.216 (0.109)}}
\ptmeasure\ptSrmseWidth{\mbox{0.220 (0.105)}}
\ptmeasure\ptSrmseWidth{\mbox{0.222 (0.107)}}
\def\ptFnWidth{0pt}
\ptmeasure\ptFnWidth{FN}
\ptmeasure\ptFnWidth{\mbox{0.59 (0.67)}}
\ptmeasure\ptFnWidth{\mbox{0.59 (0.68)}}
\ptmeasure\ptFnWidth{\mbox{0.61 (0.49)}}
\ptmeasure\ptFnWidth{\mbox{0.70 (0.46)}}
\ptmeasure\ptFnWidth{\mbox{0.71 (0.46)}}
\ptmeasure\ptFnWidth{\mbox{0.71 (0.73)}}
\ptmeasure\ptFnWidth{\mbox{0.76 (0.43)}}
\ptmeasure\ptFnWidth{\mbox{0.77 (0.42)}}
\ptmeasure\ptFnWidth{\mbox{0.86 (0.35)}}
\ptmeasure\ptFnWidth{\mbox{0.89 (0.31)}}
\ptmeasure\ptFnWidth{\mbox{0.93 (0.26)}}
\ptmeasure\ptFnWidth{\mbox{1.00 (0.00)}}
\ptmeasure\ptFnWidth{\mbox{1.01 (0.22)}}
\ptmeasure\ptFnWidth{\mbox{1.10 (0.39)}}
\ptmeasure\ptFnWidth{\mbox{1.14 (0.51)}}
\ptmeasure\ptFnWidth{\mbox{NA}}
\def\ptFpWidth{0pt}
\ptmeasure\ptFpWidth{FP}
\ptmeasure\ptFpWidth{\mbox{0 (0)}}
\ptmeasure\ptFpWidth{\mbox{0.01 (0.10)}}
\ptmeasure\ptFpWidth{\mbox{0.14 (0.40)}}
\ptmeasure\ptFpWidth{\mbox{0.19 (0.39)}}
\ptmeasure\ptFpWidth{\mbox{0.19 (0.42)}}
\ptmeasure\ptFpWidth{\mbox{0.72 (0.81)}}
\ptmeasure\ptFpWidth{\mbox{0.77 (0.71)}}
\ptmeasure\ptFpWidth{\mbox{0.90 (0.76)}}
\ptmeasure\ptFpWidth{\mbox{2.22 (0.95)}}
\ptmeasure\ptFpWidth{\mbox{2.27 (1.20)}}
\ptmeasure\ptFpWidth{\mbox{2.29 (1.32)}}
\ptmeasure\ptFpWidth{\mbox{2.34 (1.11)}}
\ptmeasure\ptFpWidth{\mbox{2.58 (1.28)}}
\ptmeasure\ptFpWidth{\mbox{2.62 (1.07)}}
\ptmeasure\ptFpWidth{\mbox{3.75 (2.02)}}
\ptmeasure\ptFpWidth{\mbox{4.03 (1.90)}}
\ptmeasure\ptFpWidth{\mbox{4.23 (2.02)}}
\ptmeasure\ptFpWidth{\mbox{NA}}
\def\ptIterWidth{0pt}
\ptmeasure\ptIterWidth{Iter}
\ptmeasure\ptIterWidth{\mbox{---}}
\ptmeasure\ptIterWidth{\mbox{15.1 (3.1)}}
\ptmeasure\ptIterWidth{\mbox{15.8 (3.4)}}
\ptmeasure\ptIterWidth{\mbox{19.1 (3.4)}}
\ptmeasure\ptIterWidth{\mbox{28.4 (2.9)}}
\ptmeasure\ptIterWidth{\mbox{28.9 (2.8)}}
\ptmeasure\ptIterWidth{\mbox{29.3 (2.7)}}
\ptmeasure\ptIterWidth{\mbox{29.4 (2.8)}}
\ptmeasure\ptIterWidth{\mbox{31.9 (2.8)}}
\ptmeasure\ptIterWidth{\mbox{32.2 (2.9)}}
\def\ptTimeWidth{0pt}
\ptmeasure\ptTimeWidth{\makecell[r]{1-core\\(Sec)}}
\ptmeasure\ptTimeWidth{\makecell[r]{2-core\\(Sec)}}
\ptmeasure\ptTimeWidth{\makecell[r]{4-core\\(Sec)}}
\ptmeasure\ptTimeWidth{\mbox{10.6 (3.60)}}
\ptmeasure\ptTimeWidth{\mbox{10.8 (4.19)}}
\ptmeasure\ptTimeWidth{\mbox{12.0 (3.33)}}
\ptmeasure\ptTimeWidth{\mbox{12.4 (3.59)}}
\ptmeasure\ptTimeWidth{\mbox{12.4 (5.43)}}
\ptmeasure\ptTimeWidth{\mbox{1780 (380)}}
\ptmeasure\ptTimeWidth{\mbox{1830 (430)}}
\ptmeasure\ptTimeWidth{\mbox{1840 (382)}}
\ptmeasure\ptTimeWidth{\mbox{2.98 (0.920)}}
\ptmeasure\ptTimeWidth{\mbox{217 (58.5)}}
\ptmeasure\ptTimeWidth{\mbox{22.1 (3.57)}}
\ptmeasure\ptTimeWidth{\mbox{22.5 (3.54)}}
\ptmeasure\ptTimeWidth{\mbox{22.6 (4.12)}}
\ptmeasure\ptTimeWidth{\mbox{220 (52.1)}}
\ptmeasure\ptTimeWidth{\mbox{222 (48.2)}}
\ptmeasure\ptTimeWidth{\mbox{23.3 (4.58)}}
\ptmeasure\ptTimeWidth{\mbox{23.8 (4.58)}}
\ptmeasure\ptTimeWidth{\mbox{23.9 (4.43)}}
\ptmeasure\ptTimeWidth{\mbox{24.0 (16.2)}}
\ptmeasure\ptTimeWidth{\mbox{26.5 (15.8)}}
\ptmeasure\ptTimeWidth{\mbox{26.9 (18.3)}}
\ptmeasure\ptTimeWidth{\mbox{27.5 (39.5)}}
\ptmeasure\ptTimeWidth{\mbox{293 (364)}}
\ptmeasure\ptTimeWidth{\mbox{295 (257)}}
\ptmeasure\ptTimeWidth{\mbox{3.06 (1.04)}}
\ptmeasure\ptTimeWidth{\mbox{3.19 (0.986)}}
\ptmeasure\ptTimeWidth{\mbox{3.20 (1.02)}}
\ptmeasure\ptTimeWidth{\mbox{3.27 (1.14)}}
\ptmeasure\ptTimeWidth{\mbox{3.54 (1.19)}}
\ptmeasure\ptTimeWidth{\mbox{3.60 (1.48)}}
\ptmeasure\ptTimeWidth{\mbox{3.70 (1.58)}}
\ptmeasure\ptTimeWidth{\mbox{31.9 (11.7)}}
\ptmeasure\ptTimeWidth{\mbox{32.0 (12.5)}}
\ptmeasure\ptTimeWidth{\mbox{32.4 (14.6)}}
\ptmeasure\ptTimeWidth{\mbox{33.7 (3.06)}}
\ptmeasure\ptTimeWidth{\mbox{33.8 (18.5)}}
\ptmeasure\ptTimeWidth{\mbox{3350 (614)}}
\ptmeasure\ptTimeWidth{\mbox{3370 (595)}}
\ptmeasure\ptTimeWidth{\mbox{34.1 (2.74)}}
\ptmeasure\ptTimeWidth{\mbox{34.1 (7.22)}}
\ptmeasure\ptTimeWidth{\mbox{34.6 (5.63)}}
\ptmeasure\ptTimeWidth{\mbox{34.7 (20.0)}}
\ptmeasure\ptTimeWidth{\mbox{34.7 (6.05)}}
\ptmeasure\ptTimeWidth{\mbox{34.8 (20.4)}}
\ptmeasure\ptTimeWidth{\mbox{3460 (655)}}
\ptmeasure\ptTimeWidth{\mbox{35.0 (8.27)}}
\ptmeasure\ptTimeWidth{\mbox{35.5 (14.8)}}
\ptmeasure\ptTimeWidth{\mbox{35.7 (23.0)}}
\ptmeasure\ptTimeWidth{\mbox{36.3 (3.27)}}
\ptmeasure\ptTimeWidth{\mbox{36.4 (4.45)}}
\ptmeasure\ptTimeWidth{\mbox{36.7 (6.49)}}
\ptmeasure\ptTimeWidth{\mbox{36.8 (5.88)}}
\ptmeasure\ptTimeWidth{\mbox{36.8 (6.51)}}
\ptmeasure\ptTimeWidth{\mbox{37.4 (6.84)}}
\ptmeasure\ptTimeWidth{\mbox{39.4 (6.16)}}
\ptmeasure\ptTimeWidth{\mbox{39.8 (7.63)}}
\ptmeasure\ptTimeWidth{\mbox{4.08 (1.93)}}
\ptmeasure\ptTimeWidth{\mbox{4.38 (1.16)}}
\ptmeasure\ptTimeWidth{\mbox{4.79 (1.64)}}
\ptmeasure\ptTimeWidth{\mbox{40.3 (25.3)}}
\ptmeasure\ptTimeWidth{\mbox{41.0 (7.55)}}
\ptmeasure\ptTimeWidth{\mbox{42.0 (20.0)}}
\ptmeasure\ptTimeWidth{\mbox{44.3 (19.9)}}
\ptmeasure\ptTimeWidth{\mbox{445 (1180)}}
\ptmeasure\ptTimeWidth{\mbox{446 (819)}}
\ptmeasure\ptTimeWidth{\mbox{45.9 (27.5)}}
\ptmeasure\ptTimeWidth{\mbox{455 (804)}}
\ptmeasure\ptTimeWidth{\mbox{46.4 (26.4)}}
\ptmeasure\ptTimeWidth{\mbox{48.1 (19.2)}}
\ptmeasure\ptTimeWidth{\mbox{5.31 (1.52)}}
\ptmeasure\ptTimeWidth{\mbox{571 (1240)}}
\ptmeasure\ptTimeWidth{\mbox{576 (135)}}
\ptmeasure\ptTimeWidth{\mbox{578 (261)}}
\ptmeasure\ptTimeWidth{\mbox{605 (338)}}
\ptmeasure\ptTimeWidth{\mbox{645 (323)}}
\ptmeasure\ptTimeWidth{\mbox{727 (964)}}
\ptmeasure\ptTimeWidth{\mbox{74.5 (9.60)}}
\ptmeasure\ptTimeWidth{\mbox{74.9 (9.95)}}
\ptmeasure\ptTimeWidth{\mbox{75.4 (7.08)}}
\ptmeasure\ptTimeWidth{\mbox{793 (1920)}}
\ptmeasure\ptTimeWidth{\mbox{8.65 (2.70)}}
\ptmeasure\ptTimeWidth{\mbox{8.81 (2.87)}}
\ptmeasure\ptTimeWidth{\mbox{813 (165)}}
\ptmeasure\ptTimeWidth{\mbox{850 (187)}}
\ptmeasure\ptTimeWidth{\mbox{857 (229)}}
\ptmeasure\ptTimeWidth{\mbox{9.13 (3.44)}}
\ptmeasure\ptTimeWidth{\mbox{9.27 (3.29)}}
\ptmeasure\ptTimeWidth{\mbox{917 (163)}}
\ptmeasure\ptTimeWidth{\mbox{922 (174)}}
\ptmeasure\ptTimeWidth{\mbox{963 (238)}}
\def\ptUpdateWidth{\edef\ptTableWidth{\the\dimexpr\ptMethodWidth+\ptSrmseWidth+\ptFnWidth+\ptFpWidth+\ptIterWidth+\ptTimeWidth+\ptTimeWidth+\ptTimeWidth+14\tabcolsep\relax}}
\ptUpdateWidth\ifdim\ptTableWidth>\linewidth\setlength{\tabcolsep}{1pt}\ptUpdateWidth\fi
\typeout{PERFORMANCE-WIDTH supp_satellite: table=\ptTableWidth; time=\ptTimeWidth; linewidth=\the\linewidth}
\ifdim\ptTableWidth>\linewidth\PackageError{performance-tables}{Table exceeds linewidth}{Check the table layout.}\fi
\setlength{\LTcapwidth}{\ptTableWidth}
\Needspace{570pt}
\begin{longtable}{@{}>{\raggedright\arraybackslash}p{\ptMethodWidth}>{\raggedleft\arraybackslash}p{\ptSrmseWidth}>{\raggedleft\arraybackslash}p{\ptFnWidth}>{\raggedleft\arraybackslash}p{\ptFpWidth}>{\raggedleft\arraybackslash}p{\ptIterWidth}>{\raggedleft\arraybackslash}p{\ptTimeWidth}>{\raggedleft\arraybackslash}p{\ptTimeWidth}>{\raggedleft\arraybackslash}p{\ptTimeWidth}@{}}
\caption{Satellite drag, three responses fitted separately; $n=200$, $d=14$, $m=100$, $B=100$. The reference set contains the seven physical inputs.}\label{tab:s-satellite}\\
\toprule
Method & SRMSE & FN & FP & Iter & \makecell[r]{1-core\\(Sec)} & \makecell[r]{2-core\\(Sec)} & \makecell[r]{4-core\\(Sec)} \\
\midrule\endfirsthead
\caption[]{Satellite drag, three responses fitted separately; $n=200$, $d=14$, $m=100$, $B=100$. The reference set contains the seven physical inputs. (continued).}\\\toprule
Method & SRMSE & FN & FP & Iter & \makecell[r]{1-core\\(Sec)} & \makecell[r]{2-core\\(Sec)} & \makecell[r]{4-core\\(Sec)} \\
\midrule\endhead
\midrule
\multicolumn{8}{@{}p{\ptTableWidth}@{}}{\footnotesize Values are reported as mean (sample SD). Prediction, FN/FP and Iter use 1-core. An em dash in Iter denotes inapplicability. NA denotes no approved binary selection rule.}\\
\multicolumn{8}{r@{}}{\footnotesize Continued on next page.}\\\endfoot
\bottomrule
\endlastfoot
\multicolumn{8}{@{}l}{\textbf{Response: $\mathrm{He}$}}\\*
\texttt{optK}\textsubscript{0.005} & \mbox{0.102 (0.0175)} & \mbox{0.71 (0.46)} & \mbox{2.29 (1.32)} & \mbox{32.2 (2.9)} & \mbox{45.9 (27.5)} & \mbox{48.1 (19.2)} & \mbox{44.3 (19.9)} \\*
\texttt{optK-pruned} & \mbox{0.0863 (0.0165)} & \mbox{0.86 (0.35)} & \mbox{0.72 (0.81)} & \mbox{31.9 (2.8)} & \mbox{40.3 (25.3)} & \mbox{46.4 (26.4)} & \mbox{42.0 (20.0)} \\*
\texttt{optK}\textsubscript{0.05} & \mbox{0.120 (0.0292)} & \mbox{1.01 (0.22)} & \mbox{0.19 (0.39)} & \mbox{19.1 (3.4)} & \mbox{5.31 (1.52)} & \mbox{10.6 (3.60)} & \mbox{10.8 (4.19)} \\*
\texttt{laGP} & \mbox{0.149 (0.0144)} & \mbox{NA} & \mbox{NA} & \mbox{---} & \mbox{4.08 (1.93)} & \mbox{3.60 (1.48)} & \mbox{3.70 (1.58)} \\*
\texttt{mlegp} & \mbox{0.220 (0.105)} & \mbox{0.59 (0.67)} & \mbox{3.75 (2.02)} & \mbox{---} & \mbox{36.3 (3.27)} & \mbox{33.7 (3.06)} & \mbox{34.1 (7.22)} \\*
\texttt{MRFA} & \mbox{0.207 (0.115)} & \mbox{0.76 (0.43)} & \mbox{2.27 (1.20)} & \mbox{---} & \mbox{12.4 (5.43)} & \mbox{22.6 (4.12)} & \mbox{23.9 (4.43)} \\*
\texttt{DART} & \mbox{0.138 (0.0139)} & \mbox{1.00 (0.00)} & \mbox{0 (0)} & \mbox{---} & \mbox{75.4 (7.08)} & \mbox{41.0 (7.55)} & \mbox{37.4 (6.84)} \\*
\texttt{SVGP-ARD} & \mbox{0.191 (0.0176)} & \mbox{NA} & \mbox{NA} & \mbox{---} & \mbox{217 (58.5)} & \mbox{446 (819)} & \mbox{571 (1240)} \\*
\texttt{DGP-2} & \mbox{0.118 (0.0138)} & \mbox{NA} & \mbox{NA} & \mbox{---} & \mbox{813 (165)} & \mbox{645 (323)} & \mbox{605 (338)} \\*
\texttt{AS-GP} & \mbox{0.127 (0.0206)} & \mbox{NA} & \mbox{NA} & \mbox{---} & \mbox{3370 (595)} & \mbox{1830 (430)} & \mbox{963 (238)} \\
\midrule
\multicolumn{8}{@{}l}{\textbf{Response: $\mathrm{N}_2$}}\\*
\texttt{optK}\textsubscript{0.005} & \mbox{0.0935 (0.0112)} & \mbox{0.61 (0.49)} & \mbox{2.34 (1.11)} & \mbox{29.3 (2.7)} & \mbox{26.5 (15.8)} & \mbox{35.7 (23.0)} & \mbox{31.9 (11.7)} \\*
\texttt{optK-pruned} & \mbox{0.0857 (0.0137)} & \mbox{0.89 (0.31)} & \mbox{0.77 (0.71)} & \mbox{28.9 (2.8)} & \mbox{26.9 (18.3)} & \mbox{35.5 (14.8)} & \mbox{32.0 (12.5)} \\*
\texttt{optK}\textsubscript{0.05} & \mbox{0.122 (0.0396)} & \mbox{1.14 (0.51)} & \mbox{0.19 (0.42)} & \mbox{15.8 (3.4)} & \mbox{4.79 (1.64)} & \mbox{9.27 (3.29)} & \mbox{9.13 (3.44)} \\*
\texttt{laGP} & \mbox{0.164 (0.0145)} & \mbox{NA} & \mbox{NA} & \mbox{---} & \mbox{3.54 (1.19)} & \mbox{3.20 (1.02)} & \mbox{3.19 (0.986)} \\*
\texttt{mlegp} & \mbox{0.222 (0.107)} & \mbox{0.71 (0.73)} & \mbox{4.23 (2.02)} & \mbox{---} & \mbox{36.7 (6.49)} & \mbox{34.7 (6.05)} & \mbox{35.0 (8.27)} \\*
\texttt{MRFA} & \mbox{0.142 (0.0191)} & \mbox{0.77 (0.42)} & \mbox{2.62 (1.07)} & \mbox{---} & \mbox{12.4 (3.59)} & \mbox{22.5 (3.54)} & \mbox{23.3 (4.58)} \\*
\texttt{DART} & \mbox{0.119 (0.00957)} & \mbox{1.00 (0.00)} & \mbox{0 (0)} & \mbox{---} & \mbox{74.9 (9.95)} & \mbox{39.4 (6.16)} & \mbox{36.8 (6.51)} \\*
\texttt{SVGP-ARD} & \mbox{0.208 (0.0132)} & \mbox{NA} & \mbox{NA} & \mbox{---} & \mbox{220 (52.1)} & \mbox{293 (364)} & \mbox{455 (804)} \\*
\texttt{DGP-2} & \mbox{0.130 (0.0124)} & \mbox{NA} & \mbox{NA} & \mbox{---} & \mbox{850 (187)} & \mbox{793 (1920)} & \mbox{727 (964)} \\*
\texttt{AS-GP} & \mbox{0.149 (0.0230)} & \mbox{NA} & \mbox{NA} & \mbox{---} & \mbox{3460 (655)} & \mbox{1840 (382)} & \mbox{917 (163)} \\
\midrule
\multicolumn{8}{@{}l}{\textbf{Response: $\mathrm{O}_2$}}\\*
\texttt{optK}\textsubscript{0.005} & \mbox{0.0979 (0.0118)} & \mbox{0.71 (0.46)} & \mbox{2.22 (0.95)} & \mbox{29.4 (2.8)} & \mbox{24.0 (16.2)} & \mbox{33.8 (18.5)} & \mbox{34.8 (20.4)} \\*
\texttt{optK-pruned} & \mbox{0.0906 (0.0215)} & \mbox{0.93 (0.26)} & \mbox{0.90 (0.76)} & \mbox{28.4 (2.9)} & \mbox{27.5 (39.5)} & \mbox{34.7 (20.0)} & \mbox{32.4 (14.6)} \\*
\texttt{optK}\textsubscript{0.05} & \mbox{0.127 (0.0331)} & \mbox{1.10 (0.39)} & \mbox{0.14 (0.40)} & \mbox{15.1 (3.1)} & \mbox{4.38 (1.16)} & \mbox{8.65 (2.70)} & \mbox{8.81 (2.87)} \\*
\texttt{laGP} & \mbox{0.146 (0.0144)} & \mbox{NA} & \mbox{NA} & \mbox{---} & \mbox{3.27 (1.14)} & \mbox{2.98 (0.920)} & \mbox{3.06 (1.04)} \\*
\texttt{mlegp} & \mbox{0.216 (0.109)} & \mbox{0.59 (0.68)} & \mbox{4.03 (1.90)} & \mbox{---} & \mbox{36.8 (5.88)} & \mbox{34.1 (2.74)} & \mbox{34.6 (5.63)} \\*
\texttt{MRFA} & \mbox{0.151 (0.0136)} & \mbox{0.70 (0.46)} & \mbox{2.58 (1.28)} & \mbox{---} & \mbox{12.0 (3.33)} & \mbox{22.1 (3.57)} & \mbox{23.8 (4.58)} \\*
\texttt{DART} & \mbox{0.121 (0.0105)} & \mbox{1.00 (0.00)} & \mbox{0.01 (0.10)} & \mbox{---} & \mbox{74.5 (9.60)} & \mbox{39.8 (7.63)} & \mbox{36.4 (4.45)} \\*
\texttt{SVGP-ARD} & \mbox{0.187 (0.0140)} & \mbox{NA} & \mbox{NA} & \mbox{---} & \mbox{222 (48.2)} & \mbox{295 (257)} & \mbox{445 (1180)} \\*
\texttt{DGP-2} & \mbox{0.126 (0.0113)} & \mbox{NA} & \mbox{NA} & \mbox{---} & \mbox{857 (229)} & \mbox{576 (135)} & \mbox{578 (261)} \\*
\texttt{AS-GP} & \mbox{0.161 (0.0285)} & \mbox{NA} & \mbox{NA} & \mbox{---} & \mbox{3350 (614)} & \mbox{1780 (380)} & \mbox{922 (174)} \\
\end{longtable}\endgroup

\end{document}